\documentclass{article} 
\usepackage{iclr2027_conference,times}

\usepackage{amsmath,amsfonts,bm}

\def\eqref#1{equation~\ref{#1}}

\def\1{\bm{1}}

\DeclareMathAlphabet{\mathsfit}{\encodingdefault}{\sfdefault}{m}{sl}
\SetMathAlphabet{\mathsfit}{bold}{\encodingdefault}{\sfdefault}{bx}{n}

\usepackage{hyperref}
\usepackage{url}
\usepackage{mathtools} 
\usepackage{amsthm}

\usepackage{booktabs}   
\usepackage{multirow}   
\usepackage{colortbl}
\usepackage{wrapfig}
\usepackage{enumitem}
\usepackage{makecell}
\usepackage{capt-of}
\usepackage{ragged2e}
\usepackage{pifont}
\usepackage{graphicx}
\usepackage{subcaption}

\newcommand{\cmark}{\ding{51}} 
\newcommand{\xmark}{\ding{55}} 

\title{PoSTCache: From Forecasting to Postdictive Smoothing for Revisable Diffusion Caching}

\author{%
Liwei Lan$^{1}$ \quad Hui Chen$^{2}$
 \quad \textbf{Jungong Han}$^{3}$ \quad \textbf{Guiguang Ding}$^{1}$ \\
{$^1$School of Software, Tsinghua University \quad$^2$BNRist, Tsinghua University} \\ {$^3$Department of Automation, Tsinghua University} \\
{\tt\small llw25@mails.tsinghua.edu.cn, huichen@mail.tsinghua.edu.cn} \\
{\tt\small \{jghan,\ dinggg\}@tsinghua.edu.cn}
}

\iclrfinalcopy 
\begin{document}

\maketitle

\begin{abstract}
Diffusion and flow matching models achieve impressive visual generation quality but remain computationally expensive due to iterative sampling. Caching-based acceleration provides a practical training-free solution by reusing or predicting intermediate features. However, existing methods largely enforce a forward-only information flow: approximations produced for skipped model evaluations are immediately committed to the sampling trajectory and cannot benefit from feedback generated by later full-computation steps. We propose \textbf{PoSTCache} (\emph{Postdictive Smoothing Trajectory Cache}), a training-free and plug-and-play framework that reformulates diffusion caching as revisable trajectory inference. PoSTCache first uses an existing caching strategy to construct forward proposals for skipped steps. At a later full-computation step, the discrepancy between the refreshed prediction and its forward proposal yields a correction signal, which PoSTCache propagates backward to refine and reintegrate the preceding predictions. Viewing full-computation steps as sources of corrective feedback also changes how they should be selected. We therefore introduce the Posterior Variance Reduction (PVR) strategy, which estimates cross-timestep dependencies from offline denoising trajectories and selects anchors according to their expected reduction in surrogate trajectory uncertainty rather than local approximation error alone. Experiments on representative image and video generation models show that PoSTCache consistently improves the quality--efficiency trade-off over representative caching methods. 
\end{abstract} 

\section{Introduction}

Diffusion models~\citep{song2019generative, ho2020denoising, song2020score} and flow matching models~\citep{lipman2023flow, liu2023flow} have become a dominant paradigm for modern visual generation. By progressively transporting noisy latent variables toward structured data samples, they enable impressive image~\citep{rombach2022high, podell2024sdxl} and video synthesis quality~\citep{blattmann2023stable, chen2024videocrafter2}. Despite this progress, sampling still requires tens to hundreds of steps, making user-facing applications latency-bound.

\begin{figure}[t]
    \centering
    \includegraphics[width=\linewidth]{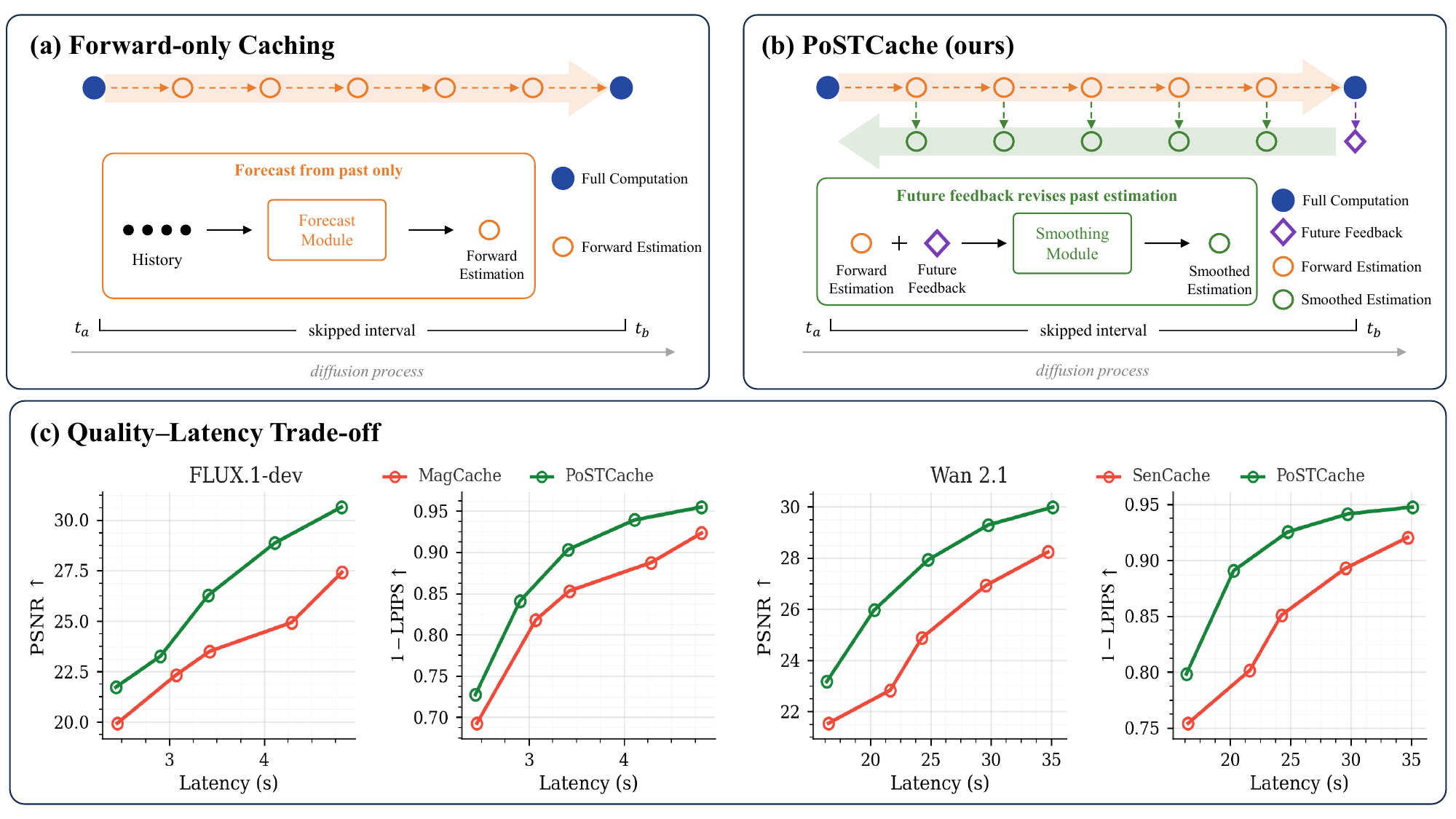}
    \caption{\textbf{From forecasting to postdictive smoothing.}
    (a) Forward-only caching commits skipped predictions using historical information and leaves the resulting trajectory drift unrevised.
    (b) PoSTCache additionally propagates feedback from a later full-computation step backward to retrospectively refine skipped predictions and correct the trajectory.
    (c) PoSTCache consistently improves the quality--latency trade-off over strong caching baselines across image and video models.}
    \label{fig:teaser}
\end{figure}

Traditional acceleration techniques, such as distillation~\citep{song2023consistency} and post-training quantization~\citep{li2023q}, often incur additional training costs, architectural modifications, or non-trivial deployment efforts. In contrast, caching provides a lightweight, training-free alternative by exploiting the temporal redundancy of intermediate representations~\citep{liu2025timestep,bu2025dicache,chung2026seacache}. Existing caching approaches have evolved from static feature reuse~\citep{selvaraju2024fora} to adaptive strategies that estimate inter-step redundancy online~\citep{liu2025timestep,ma2026magcache,haghighi2026sencache}, extrapolate future feature trajectories with polynomial approximations~\citep{liu2025reusing,feng2026hicache}, or exploit shallow-layer dynamics to guide deep-layer feature reuse~\citep{bu2025dicache}. More recently, caching has also been extended beyond the original feature space to spectral representations~\citep{chung2026seacache}.

Despite their diversity, most existing caching methods follow a \emph{strictly forward information flow}, which we refer to as \emph{forecasting}: skipped-step predictions are constructed from historical information and committed to the sampling trajectory without later revision, allowing approximation errors to propagate through subsequent updates, as illustrated in Figure~\ref{fig:teaser}(a). This raises a broader question: does later sampling reveal new information that could help correct these earlier errors? In particular, periodic full-computation steps produce fresh predictions after the trajectory has evolved through preceding approximations, potentially exposing backward-looking information about accumulated errors. For example, the discrepancy between a refreshed prediction and its forward proposal can provide such corrective evidence. However, most existing methods use this newly revealed information only for subsequent sampling, rather than propagating it backward to revise earlier estimates. This motivates a \emph{postdictive smoothing} view of diffusion caching, inspired by Bayesian smoothing, in which later evidence retrospectively revises uncertain intermediate estimates~\citep{rauch1965maximum,rabiner1989tutorial,sarkka2023bayesian}. Accordingly, we treat skipped predictions as provisional rather than final, allowing feedback from later full-computation steps to retrospectively refine these estimates and enable \emph{backward information flow}, as illustrated in Figure~\ref{fig:teaser}(b).

Motivated by this perspective, we propose PoSTCache, short for \emph{Postdictive Smoothing Trajectory Cache}, a training-free and plug-and-play framework that reformulates cache-based acceleration as revisable trajectory inference. Building on an existing caching strategy, PoSTCache propagates feedback from subsequent full-computation anchors backward to refine skipped predictions and reintegrates the corrected trajectory. This shift also changes how anchors should be valued: from their local approximation error to their trajectory-level corrective value. We therefore introduce the Posterior Variance Reduction (PVR) strategy, which estimates cross-timestep dependencies from offline denoising trajectories and greedily selects anchors to reduce surrogate trajectory uncertainty under a fixed computation budget. Figure~\ref{fig:teaser}(c) previews the resulting quality--latency trade-off.

Our contributions are summarized as follows:
\begin{itemize}
\item \textbf{From Forecasting to Postdictive Smoothing.} We propose PoSTCache, a training-free and plug-and-play framework that reformulates diffusion caching from forward-only forecasting into postdictive smoothing by retrospectively revising skipped predictions.
\item \textbf{Posterior Variance Reduction (PVR) Anchor Selection.} We propose a smoothing-aware PVR strategy that profiles offline denoising trajectories and greedily selects anchors to reduce surrogate trajectory uncertainty under a fixed computation budget.
\item  \textbf{Superior Performance.} Comprehensive experiments across representative image and video models show that PoSTCache improves reconstruction fidelity at comparable or higher acceleration while maintaining competitive semantic generation quality.
\end{itemize}
\section{Related Work}

\textbf{Generative Model Acceleration.}
Acceleration methods for diffusion and flow-based models reduce
either the cost of each model evaluation or the number of sampling steps.
The former includes model compression techniques such as
pruning~\citep{fang2023structural},
quantization~\citep{li2023q},
and knowledge distillation~\citep{meng2023distillation},
as well as computation-efficient designs such as
FlashAttention~\citep{dao2022flashattention}
and token reduction~\citep{bolya2022token}.
The latter improves sampling efficiency through deterministic sampling with
DDIM~\citep{song2020denoising},
high-order ODE solvers such as DPM-Solver~\citep{lu2022dpm},
alternative generative dynamics including flow matching~\citep{lipman2023flow}
and rectified flow~\citep{liu2023flow},
or few-step generation through progressive distillation~\citep{salimans2022progressive},
consistency models~\citep{song2023consistency},
Shortcut Models~\citep{frans2025one},
and MeanFlow~\citep{geng2026mean}.
Although effective, these approaches often require additional training,
architectural changes, or modifications to the sampling process.

\textbf{Cache-based Acceleration.}
Caching provides a lightweight and training-free alternative by exploiting the
temporal redundancy of intermediate representations.
Early methods reuse U-Net features with fixed schedules, including
DeepCache~\citep{ma2024deepcache}
and FasterDiffusion~\citep{li2024faster},
while FORA~\citep{selvaraju2024fora}
and $\Delta$-DiT~\citep{chen2024delta}
extend feature or attention reuse to diffusion transformers.
More recent approaches adaptively determine when to refresh cached features
using different error proxies, such as
TeaCache~\citep{liu2025timestep},
MagCache~\citep{ma2026magcache},
and DiCache~\citep{bu2025dicache}.
TaylorSeer~\citep{liu2025reusing}
and HiCache~\citep{feng2026hicache}
instead predict future features from historical states.
Other methods incorporate trajectory priors, including
spectral evolution in SeaCache~\citep{chung2026seacache},
global path planning in DPCache~\citep{cui2026dpcache}, 
average-velocity stability in MeanCache~\citep{gao2026meancache}, and offline residual rectification in ERTACache~\citep{peng2026ertacache}.
Despite these advances, most methods remain online forward-only: skipped states are determined using historical information or offline priors and remain unrevised after
future full-computation states become available.

\textbf{Bayesian Posterior Smoothing.}
Bayesian smoothing estimates an intermediate latent state using both past and
future observations, whereas filtering conditions only on observations
available up to the current time.
For linear Gaussian systems, this principle underlies Kalman
filtering~\citep{kalman1960new}
and the Rauch--Tung--Striebel smoother~\citep{rauch1965maximum}.
A similar forward--backward inference procedure is used in hidden Markov
models~\citep{rabiner1989tutorial}.
Bayesian smoothing has been generalized to nonlinear and non-Gaussian
state-space models~\citep{sarkka2023bayesian}.
These methods share the central idea that future evidence can retrospectively
correct uncertain intermediate estimates, motivating our reformulation of
diffusion caching as revisable trajectory inference.
\section{Method}
\label{sec:method}
\subsection{Preliminaries}
\label{sec:preliminaries}



Diffusion sampling evolves a latent trajectory through a sequence of model
predictions and state updates, resembling sequential Bayesian estimation~\citep{kalman1960new}.
At each denoising step, the model produces a prediction from the information
available so far, which is then used by the numerical solver to advance the
trajectory. When some predictions are replaced by inaccurate approximations, their errors can propagate through subsequent updates and
accumulate along the trajectory.

Bayesian smoothing provides a principled mechanism for revising such forward
estimates once informative future evidence becomes available~\citep{rauch1965maximum,rabiner1989tutorial,sarkka2023bayesian}. Under a linear
Gaussian approximation, conditioning the forward estimate at step $t$ on an
observation at a later step $b$ gives
\begin{equation}
    m_{t\mid b}
    =
    m_t
    +
    P_{tb}(P_{bb}+R_b)^{-1}(y_b-m_b),
    \label{eq:linear_gaussian_mean_update}
\end{equation}
where $y_b-m_b$ is the future innovation, defined as the residual between the future observation $y_b$ and its forward prediction $m_b$. The factor $P_{tb}(P_{bb}+R_b)^{-1}$ is the smoothing gain. The conditional mean, therefore, takes the generic form
\begin{equation}
    \text{smoothed estimate}
    =
    \text{forward estimate}
    +
    \text{smoothing gain}
    \times
    \text{future innovation}.
\end{equation}
This perspective motivates treating skipped diffusion predictions as revisable forward estimates rather than fixed approximations. The sequential-estimation interpretation and the corresponding smoothing derivation are provided in Appendices~\ref{app:trajectory_estimation_view} and~\ref{app:bayesian_smoothing}, respectively.

\begin{figure}[t]
    \centering
    \includegraphics[width=\linewidth]{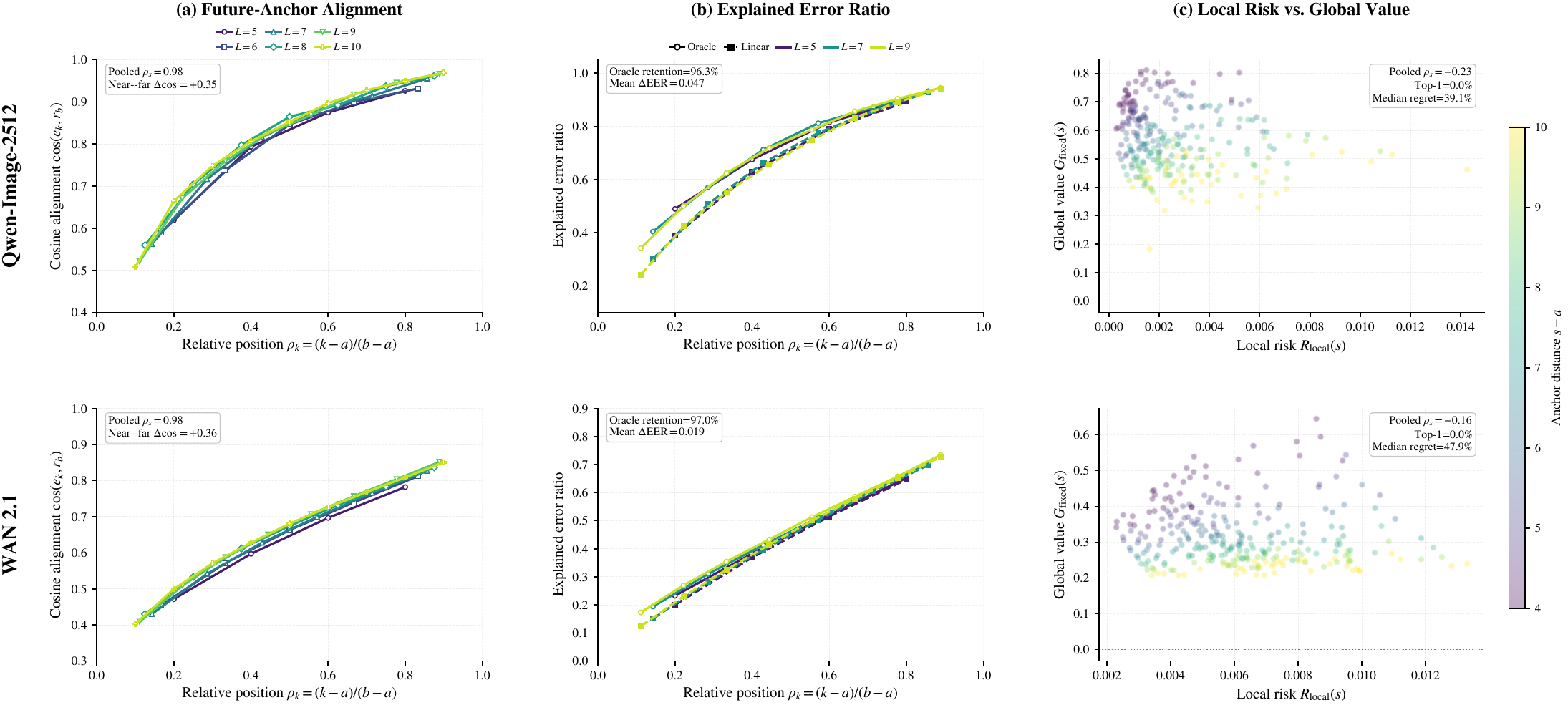}
    \caption{
        \textbf{Empirical motivation for postdictive smoothing.}
        (a) Future-anchor residuals increasingly align with preceding proposal
        errors.
        (b) The relative-position gain retains over 96\% of the explained error achieved by the oracle scalar gain, with a small mean EER gap.
        (c) Local cache risk is weakly correlated with trajectory-level correction
        value.
        }
    \label{fig:trajectory_analysis}
\end{figure}

\subsection{Trajectory Analysis}
\label{sec:trajectory_analysis}

Existing caching methods construct skipped predictions from historical states,
whose approximation errors are integrated into the sampling trajectory and
remain unrevised. We investigate whether a later full-computation step provides
useful evidence for retrospectively correcting these errors.

Consider a cache interval bounded by two full-computation steps $a<b$. Let
$\widetilde{u}_k$ and $\bar{u}_k$ denote the cached and full-computation
predictions at step $k$, respectively. We define
\begin{equation}
    e_k=\bar{u}_k-\widetilde{u}_k,
    \qquad
    r_b=\bar{u}_b-\widetilde{u}_b,
    \quad a<k<b,
\end{equation}
where $e_k$ is the proposal error, and $r_b$ is the future-anchor residual. Following the Bayesian smoothing interpretation
in Sec.~\ref{sec:preliminaries}, we hereafter refer to this residual as
the \emph{anchor innovation}.

\paragraph{Future anchors provide corrective information.}
Figure~\ref{fig:trajectory_analysis}(a) shows that $e_k$ and $r_b$ are consistently positively aligned, with stronger alignment as the intermediate step approaches the future anchor. Across interval lengths from $5$ to $10$, the relative position $\rho_k=(k-a)/(b-a)$ and the cosine alignment $\cos(e_k,r_b)$ exhibit pooled Spearman correlations of $0.98$ on both Qwen-Image-2512 and Wan~2.1. Moreover, the mean alignment at near-anchor positions ($\rho_k\geq0.75$) exceeds that at far-anchor positions ($\rho_k\leq0.25$) by $0.35$ and $0.36$, respectively. These results indicate that the anchor innovation $r_b$ contains increasingly accurate directional information about the errors accumulated within the preceding interval.

Figure~\ref{fig:trajectory_analysis}(b) measures the fraction of proposal
error removed by a scalar correction:
\begin{equation}
    \operatorname{EER}_{k,b}(g)
    =
    1-\frac{\|e_k-g r_b\|_2^2}{\|e_k\|_2^2}.
\end{equation}
We compare the oracle scalar gain
$g_{k,b}^{\mathrm{oracle}}=\arg\min_g\|e_k-g r_b\|_2^2$
with the relative-position gain
$g_{k,b}^{\mathrm{linear}}=\rho_k$. Despite requiring no sample-specific estimation, the linear gain retains
$96.3\%$ and $97.0\%$ of the oracle explained error on Qwen-Image-2512 and
Wan~2.1, with mean EER gaps of only $0.047$ and $0.019$, respectively.
This shows that sampling progress provides a simple approximation to the
oracle scalar correction strength, motivating our time-dependent smoothing gain.

\paragraph{Local error is a poor proxy for anchor informativeness.}
Figure~\ref{fig:trajectory_analysis}(c) compares local cache risk with trajectory-level correction value (Appendix~\ref{app:additional_trajectory_analysis}). The two criteria are poorly aligned: their pooled Spearman correlations are negative, with $0\%$ top-$1$ agreement and median regrets of $39.1\%$ and $47.9\%$ on Qwen-Image-2512 and Wan~2.1, respectively. Therefore, the step with the largest local error is not necessarily the most informative anchor for correcting the preceding trajectory.



\subsection{PoSTCache}
\label{sec:postcache}

\begin{figure}[t]
    \begin{center}
    \includegraphics[width=1.0\linewidth]{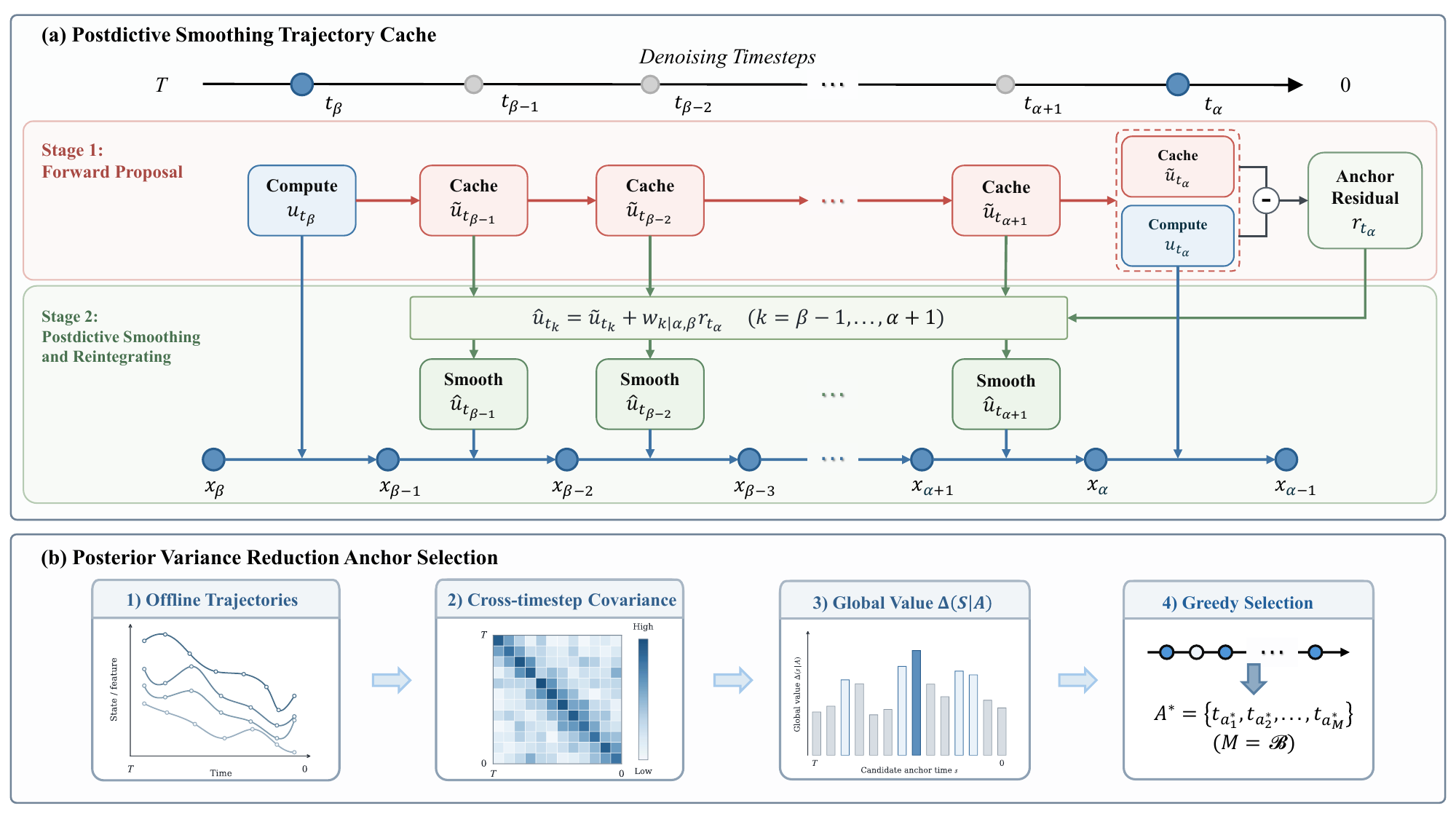}
    \end{center}
    \caption{\textbf{Overview of PoSTCache}. PoSTCache consists of postdictive smoothing
and reintegration together with posterior variance reduction anchor selection.
The former uses future anchor innovations to retrospectively correct skipped
predictions and reintegrate the sampling trajectory. The latter profiles
complete offline denoising trajectories, estimates their cross-timestep
covariance, and greedily selects anchors that increase global
informativeness under a fixed computation budget.}
    \label{fig:pipeline}
\end{figure}

Motivated by the Bayesian smoothing principle in
Sec.~\ref{sec:preliminaries}, which revises a forward estimate using a
gain-weighted anchor innovation, and supported by the empirical observations
in Sec.~\ref{sec:trajectory_analysis}, PoSTCache uses future-anchor innovations
for retrospective trajectory correction and trajectory-level uncertainty for
anchor selection, as shown in Figure~\ref{fig:pipeline}.

\subsubsection{Postdictive Smoothing Trajectory Cache}
\label{sec:postdictive_smoothing}

Let $u_k=f_\theta(x_k,\tau_k)$ denote the model prediction consumed by the
numerical solver at the $k$-th denoising step. Consider two consecutive
full-computation anchors $a<b$. Starting from anchor $a$, an existing caching
method constructs forward proposals for the intermediate predictions:
\begin{equation}
    \widetilde{u}_k
    =
    \mathcal{P}_k(\mathcal{H}_a),
    \qquad
    k=a+1,\ldots,b,
    \label{eq:forward_proposal}
\end{equation}
where $\mathcal{H}_a$ contains the historical states available at anchor $a$,
and $\mathcal{P}_k$ can represent feature reuse, extrapolation, or other
forward caching strategies. In our default implementation,
$\mathcal{P}_k$ simply reuses the latest full-computation prediction.

At the next anchor $b$, a full model evaluation produces $\bar{u}_b$. The resulting anchor residual is
\begin{equation}
    r_b
    =
    \bar{u}_b-\widetilde{u}_b,
    \label{eq:anchor_innovation}
\end{equation}
where $\widetilde{u}_b$ is the corresponding forward proposal.
As established in Sec.~\ref{sec:trajectory_analysis}, this residual is
the anchor innovation used in the following smoothing update.

Under a local linear Gaussian approximation of the proposal errors,
conditioning the intermediate proposal on the future anchor innovation yields
\begin{equation}
    \widehat{u}_k
    =
    \widetilde{u}_k
    +
    G_{k\leftarrow b}
    \left(
        \bar{u}_b-\widetilde{u}_b
    \right),
    \label{eq:bayesian_postdictive_correction}
\end{equation}
where $G_{k\leftarrow b}$ is a covariance-dependent smoothing gain that propagates
the anchor innovation backward to step $k$.
It assigns stronger corrections when the intermediate proposal error is more
closely correlated with the anchor innovation, while accounting for the
uncertainty of the future evidence. This conditioning also reduces the
remaining uncertainty of the intermediate prediction. The detailed forms of
the smoothing gain and posterior covariance are provided in
Appendix~\ref{app:postdictive_derivation}.

Although computing the full high-dimensional gain is impractical, the analysis in Sec.~\ref{sec:trajectory_analysis} shows that a simple relative-position scalar gain captures most of the oracle scalar correction. We therefore adopt the lightweight isotropic approximation $G_{k\leftarrow b}\approx w_{k\mid a,b}I$, giving
\begin{equation}
    \widehat{u}_k
    =
    \widetilde{u}_k
    +
    w_{k\mid a,b}
    \left(
        \bar{u}_b-\widetilde{u}_b
    \right),
    \qquad
    w_{k\mid a,b}
    =
    \frac{\eta_k-\eta_a}{\eta_b-\eta_a},
    \quad a < k\le b,
    \label{eq:linear_postdictive_correction}
\end{equation}
where $\eta_k$ denotes the sampling progress, instantiated by the step index. The resulting gain increases from $0$ to $1$ across the interval, gradually incorporating the future evidence.

Correcting the model predictions alone does not revise the latent states
already produced within the skipped interval. We therefore reintegrate the
corrected predictions from anchor $a$:
\begin{equation}
    \widehat{x}_{k+1}
    =
    \mathcal{S}_k
    \left(
        \widehat{x}_k,
        \widehat{u}_k,
        \tau_k
    \right),
    \qquad
    \widehat{x}_a=x_a,
    \label{eq:trajectory_reintegration}
\end{equation}
where $\mathcal{S}_k$ is the original scheduler or numerical solver.
Reintegration involves only inexpensive solver operations and requires no
additional neural-network evaluations at skipped steps; for efficiency, we
also retain the already-computed future-anchor prediction $\bar{u}_b$ rather
than re-evaluating the model at the reintegrated anchor state $\widehat{x}_b$. We further examine the resulting anchor-state consistency and its compute trade-off in Appendix~\ref{app:anchor_consistency}.

\subsubsection{Posterior Variance Reduction Anchor Selection}
\label{sec:posterior_anchor_selection}

The analysis in Sec.~\ref{sec:trajectory_analysis} shows that local cache risk does not reliably reflect the trajectory-level correction value required for effective postdictive smoothing. Since directly modeling proposal errors requires costly rollouts and depends on the proposal method, cache interval, and caching configuration, we instead profile $N$ complete offline denoising trajectories to construct a proposal-agnostic surrogate. Let $z_k^{(n)}=\bar{u}_k^{(n)}\in\mathbb{R}^{D}$ denote the full-computation model prediction at timestep $k$ for the $n$-th sample. We construct the timestep-level covariance
kernel
\begin{equation}
    \mu_k
    =
    \frac{1}{N}
    \sum_{n=1}^{N} z_k^{(n)},
    \qquad
    K_{ij}
    =
    \frac{1}{ND}
    \sum_{n=1}^{N}
    \left\langle
        z_i^{(n)}-\mu_i,\,
        z_j^{(n)}-\mu_j
    \right\rangle,
    \label{eq:cross_timestep_kernel}
\end{equation}
where $K_{ij}$ summarizes the normalized cross-timestep covariance. A large
magnitude $|K_{ij}|$ indicates a strong dependency between trajectory states
at timesteps $i$ and $j$.

Using $K$ as a covariance model, the posterior variance at step $t$
conditioned on an anchor set $A$ is
\begin{equation}
    V_t(A)
    =
    K_{tt}
    -
    K_{tA}
    K_{AA}^{-1}
    K_{At}.
    \label{eq:posterior_variance}
\end{equation}
Given a computation budget $\mathcal{B}$, we select anchors by minimizing
the weighted trajectory uncertainty:
\begin{equation}
    A^\star
    =
    \arg\min_{\substack{
        A_0\subseteq A\\
        |A|\leq\mathcal{B}
    }}
    \sum_{t=1}^{T}
    \gamma_t V_t(A),
    \label{eq:anchor_selection_objective}
\end{equation}
where $A_0$ contains the dense-prefix anchors and the final denoising step,
and $\gamma_t$ is a scheduler-aware temporal weight. Detailed implementation
settings are provided in
Appendix~\ref{app:postcache_configuration}.

Since exact optimization is combinatorial, we greedily select the candidate
with the largest marginal variance reduction. Given the current anchor set
$A$, we define the conditional cross-covariance
\begin{equation}
    C_{ts\mid A}
    =
    K_{ts}
    -
    K_{tA}
    K_{AA}^{-1}
    K_{As}.
    \label{eq:conditional_cross_covariance}
\end{equation}
Under Gaussian conditioning, adding candidate $s$ reduces the posterior
variance at timestep $t$ by $C_{ts\mid A}^{\,2}/V_s(A)$. Its weighted marginal
gain over the trajectory is therefore
\begin{equation}
    \Delta(s\mid A)
    =
    \sum_{t=1}^{T}
    \gamma_t
    \frac{C_{ts\mid A}^{\,2}}
         {V_s(A)},
    \qquad
    s^\star
    =
    \arg\max_{s\notin A}
    \Delta(s\mid A).
    \label{eq:posterior_variance_gain}
\end{equation}
This criterion favors anchors with high trajectory-level correction value rather than large local variation alone. Appendix~\ref{app:posterior_anchor_derivation} derives it as posterior variance reduction under the linear Gaussian surrogate, while Appendix~\ref{app:pvr_surrogate_validation} validates this surrogate on held-out trajectories, showing strong correlation with measured post-smoothing error ($\rho=0.858/0.963$ on Qwen-Image/Wan).
\begin{table*}[t]
\centering
\caption{Text-to-image efficiency and quality on \textbf{FLUX.1-dev} and \textbf{Qwen-Image-2512}.}
\label{tab:text_to_image}
\setlength\tabcolsep{7pt}
\resizebox{1.0\textwidth}{!}{
\begin{tabular}{l|cccc|ccccc}
\toprule
\multirow{2}{*}{\textbf{Method}} &
\multicolumn{4}{c|}{\textbf{Efficiency}} &
\multicolumn{5}{c}{\textbf{Visual Quality}} \\
\cmidrule(lr){2-5}\cmidrule(lr){6-10}

& \textbf{Latency(s) $\downarrow$}
& \textbf{Speed $\uparrow$}
& \textbf{FLOPs(T) $\downarrow$}
& \textbf{Mem(GB) $\downarrow$}
& \textbf{LPIPS $\downarrow$}
& \textbf{SSIM $\uparrow$}
& \textbf{PSNR $\uparrow$}
& \textbf{Image Reward$\uparrow$}
& \textbf{CLIP Score$\uparrow$}\\

\midrule
\midrule
\multicolumn{10}{c}{\textbf{FLUX.1-dev $1024{\times}1024$}} \\
\midrule

\textbf{Original: 50 steps}
& 9.33 & 1.00$\times$ & 2990.80 & 33.85
& -- & -- & -- & 1.0347 & 31.666\\

60\% steps
& 5.91 & 1.58$\times$ & 1800.39 & 33.85
& 0.2447 & 0.7737 & 19.177 & 1.0269 & 31.607 \\

30\% steps
& 3.30 & 2.83$\times$ & 907.59 & 33.85
& 0.4192 & 0.6472 & 15.322 & 0.9477 & 31.540 \\

\midrule

TeaCache
& 3.49 & 2.67$\times$ & 908.16 & 33.85
& 0.3413 & 0.7032 & 17.217 & 1.0249 & 31.647 \\

TaylorSeer
& 3.63 & 2.57$\times$ & 730.24 & 37.86
& 0.3515 & 0.6868 & 16.666 & 1.0033 & 31.698\\

HiCache
& 3.73 & 2.51$\times$ & 729.99 & 37.86
& 0.3500 & 0.6877 & 16.694 & 0.9996 & 31.695 \\

SeaCache
& 3.42 & 2.73$\times$ & 908.16 & 33.85
& 0.1727 & 0.8331 & 22.740 & 1.0135 & 31.674 \\

DiCache-slow
& 3.53 & 2.65$\times$ & 944.70 & 33.85
& 0.1813 & 0.8338 & 23.803 & 1.0034 & 31.702 \\

MagCache-slow
& 3.43 & 2.72$\times$ & 967.66 & 33.85
& 0.1479 & 0.8488 & 23.648 & 1.0188 & 31.712 \\

\rowcolor{gray!15}
\textbf{PoSTCache-slow}
& \bf 3.41
& \bf 2.74$\times$ & 967.11 & 33.85
& \textbf{0.0966}
& \textbf{0.8961}
& \textbf{26.625} 
& \bf 1.0322
& \bf 31.728\\

\midrule

DiCache-fast
& 3.05 & 3.06$\times$ & 769.32 & 33.85
& 0.2335 & 0.7913 & 21.713 & 0.9715 & 31.655 \\

MagCache-fast
& 3.08 & 3.03$\times$ & 848.65 & 33.85
& 0.1838 & 0.8230 & 22.448 & 1.0235 & 31.729 \\

\rowcolor{gray!15}
\textbf{PoSTCache-fast}
& \textbf{2.90}
& \textbf{3.21}$\times$ & 788.55 & 33.85
& \bf 0.1592
& \bf 0.8411
& \bf 23.410 
& \bf 1.0317
& \bf 31.758 \\
\midrule
\midrule

\multicolumn{10}{c}{
    \textbf{Qwen-Image-2512 (CFG enabled)}  $1328{\times}1328$
} \\
\midrule

\textbf{Original: 50 steps}
& 31.29
& 1.00$\times$ & 9407.07 & 60.78
& -- & -- & -- & 1.0961 & 31.544 \\

60\% steps
& 19.22
& 1.63$\times$ & 5645.49  & 60.78
& 0.1681 & 0.8278 & 20.481 & 1.0850 & 31.515 \\

30\% steps
& 10.23
& 3.06$\times$ & 2824.31 & 60.78
& 0.3239 & 0.6956 & 15.824 & 1.1055 & 31.586 \\

\midrule

TaylorSeer
& 13.56 & 2.31$\times$ & 2263.84 & 89.37
& 0.3858 & 0.6218 & 13.860 
& 1.0648
& 31.392 \\

HiCache
& 14.15 & 2.21$\times$ & 2262.15 & 89.37
& 0.3600 & 0.6235 & 14.604 
& 1.0730
& 31.385 \\

MagCache-slow
& 11.75 & 2.66$\times$ & 3200.87 & 60.78
& 0.1624 & 0.8321 & 24.397 
& 1.0631
& 31.409 \\

DiCache-slow
& 11.62 & 2.69$\times$ & 3119.38 & 60.78
& 0.1278 & 0.8670 & 24.632 & 1.0674 & 31.432 \\

\rowcolor{gray!15}
\textbf{PoSTCache-slow}
& \bf 10.78
& \bf 2.90$\times$ & 3012.39 & 60.78
& \textbf{0.0608}
& \textbf{0.9302}
& \textbf{27.588}
& \bf 1.0901
& \bf 31.631\\

\midrule

MagCache-fast
& 10.94 & 2.86$\times$ & 2918.52 & 60.78
& 0.1791 & 0.8156 & 22.623 
& 1.0706
& 31.449 \\

DiCache-fast
& 9.88 & 3.17$\times$ & 2564.58 & 60.78
& 0.1428 & 0.8504 & 23.046 & 1.0611 & 31.428 \\

\rowcolor{gray!15}
\textbf{PoSTCache-fast}
& \textbf{9.56}
& \textbf{3.27}$\times$ & 2636.23 & 60.78
& \bf 0.0899
& \bf 0.8979
& \bf 24.712
& \bf 1.0883
& \bf 31.637\\

\bottomrule
\end{tabular}}
\end{table*}

\begin{figure}[t]
    \centering
    \includegraphics[
    width=1.0\linewidth,
    trim={3.3cm 0.7cm 3.75cm 0.7cm},
    clip
    ]{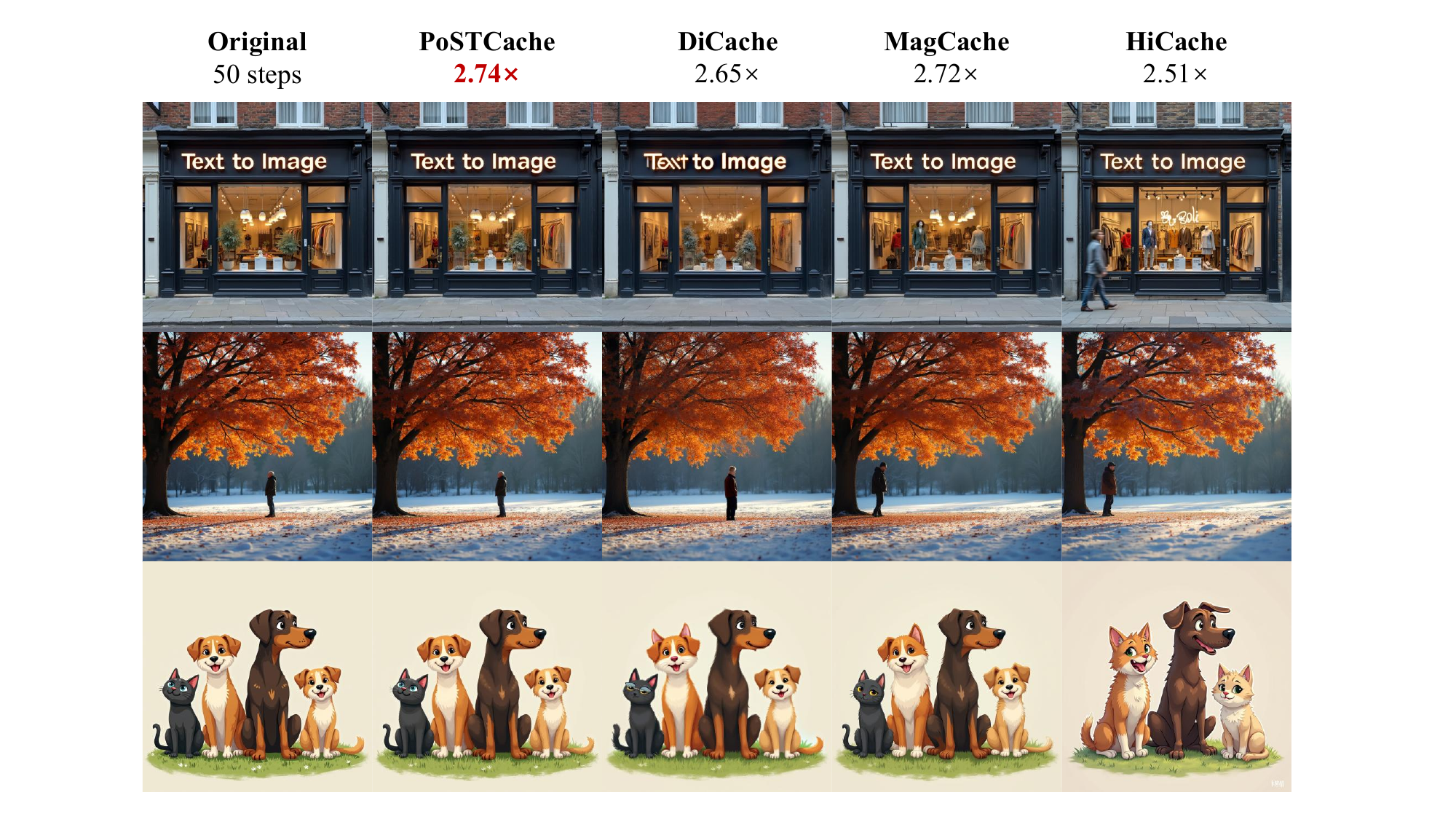}
    \caption{Qualitative comparison with previous acceleration methods on FLUX.1-dev.}
    \label{fig:image_compare}
\end{figure}

\section{Experiments}
\label{experiments}
\subsection{Settings}
\label{sec:experimental_settings}
\textbf{Base Models and Compared Methods.}
We evaluate PoSTCache on three video generation models,
Open-Sora~1.2~\citep{zheng2024open},
CogVideoX~\citep{yang2025cogvideox}, and
Wan~2.1~\citep{wang2025wan}, and two image generation models,
FLUX.1-dev~\citep{flux2024} and
Qwen-Image-2512~\citep{wu2025qwen}.
We compare against representative diffusion caching methods, including
TeaCache~\citep{liu2025timestep},
TaylorSeer~\citep{liu2025reusing},
MagCache~\citep{ma2026magcache},
DiCache~\citep{bu2025dicache},
HiCache~\citep{feng2026hicache},
SeaCache~\citep{chung2026seacache}, and
SenCache~\citep{haghighi2026sencache}. We enable classifier-free guidance (CFG)~\citep{ho2022classifier} for Qwen-Image-2512 and
Wan 2.1, using identical guidance settings for full inference and
all accelerated methods.

\textbf{Evaluation Metrics.}
We evaluate 200 DrawBench prompts~\citep{saharia2022photorealistic} with
three random seeds for text-to-image generation and 944 VBench
prompts~\citep{huang2024vbench} with five random seeds for text-to-video
generation. Following prior work~\citep{bu2025dicache,liu2025timestep}, we
report latency, speedup, FLOPs, and peak allocated GPU memory. ImageReward
and CLIP Score assess image quality, while VBench evaluates video quality.
LPIPS~\citep{zhang2018unreasonable}, SSIM~\citep{wang2004image}, and PSNR
measure reconstruction fidelity against full inference. All results are
averaged over random seeds.

\textbf{Implementation Details.}
All experiments are conducted on a single NVIDIA H200 GPU. We use 10 offline
calibration trajectories to estimate the cross-timestep covariance statistics.
Detailed model configurations, profiling protocols, and baseline
hyperparameters are provided in
Appendix~\ref{app:experimental_settings}.

\begin{table*}[t]
\centering
\caption{Text-to-video efficiency and quality on video generation models.}
\label{tab:text_to_video}
\setlength\tabcolsep{7pt}
\resizebox{1.0\textwidth}{!}{
\begin{tabular}{l|cccc|cccc}
\toprule
\multirow{2}{*}{\textbf{Method}} &
\multicolumn{4}{c|}{\textbf{Acceleration}} &
\multicolumn{4}{c}{\textbf{Visual Quality}} \\
\cmidrule(lr){2-5}\cmidrule(lr){6-9}

& \textbf{Latency(s) $\downarrow$}
& \textbf{Speed $\uparrow$}
& \textbf{FLOPs(T) $\downarrow$}
& \textbf{Mem(GB) $\downarrow$}
& \textbf{LPIPS $\downarrow$}
& \textbf{SSIM $\uparrow$}
& \textbf{PSNR $\uparrow$}
& \textbf{VBench $\uparrow$}\\

\midrule
\midrule
\multicolumn{9}{c}{
    \textbf{Wan 2.1 1.3B (CFG enabled)} (81 frames, $832{\times}480$)
} \\
\midrule

\textbf{Original: 50 steps}
& 69.40 & 1.00$\times$ & 8243.74 & 17.92
& -- & -- & -- & 79.63\% \\

60\% steps
& 42.61 & 1.63$\times$ & 4959.36 & 17.92
& 0.3489 & 0.6268 & 16.759 & 79.08\% \\

30\% steps
& 22.71 & 3.06$\times$ & 2496.07 & 17.92
& 0.4822 & 0.5385 & 14.795 & 76.93\% \\

\midrule


MagCache
& 30.79 & 2.25$\times$ & 3482.28 & 17.92
& 0.2178 & 0.7568 & 21.478 & 76.97\% \\

DiCache-slow
& 30.40 & 2.28$\times$ & 3244.21 & 17.92
& 0.2161 & 0.7480 & 21.195 & 77.30\% \\

SenCache-slow
& 29.91 & 2.32$\times$ & 3318.10 & 17.92
& 0.1034 & 0.8685 & 26.867 & 78.86\% \\

\rowcolor{gray!15}
\textbf{PoSTCache-slow}
& \bf 29.31
& \bf 2.37$\times$
& 3317.17 
& 17.92
& \textbf{0.0597}
& \textbf{0.9067}
& \textbf{29.232}
& \bf 79.44\% \\

\midrule

DiCache-fast
& 25.22 & 2.75$\times$ & 2688.71 & 17.92
& 0.2307 & 0.7293 & 20.754 & 76.50\% \\

SenCache-fast
& 24.40 & 2.84$\times$ & 2661.34 & 17.92
& 0.1449 & 0.8280 & 24.811 & 78.35\% \\

\rowcolor{gray!15}
\textbf{PoSTCache-fast}
& \textbf{24.17}
& \textbf{2.87}$\times$
& 2660.29
& 17.92
& \bf 0.0757
& \bf 0.8890
& \bf 27.829
& \bf 79.19\%\\

\midrule
\midrule
\multicolumn{9}{c}{\textbf{CogVideoX 2B} (49 frames, $720{\times}480$)} \\
\midrule

\textbf{Original: 50 steps}
& 34.92 & 1.00$\times$ & 4751.86 & 29.39
& -- & -- & -- & 77.11\% \\

60\% steps
& 22.10 & 1.58$\times$ & 2863.84 & 29.39
& 0.3329 & 0.6737 & 17.381 & 75.15\% \\

30\% steps
& 12.69 & 2.75$\times$ & 1447.83 & 29.39
& 0.4663 & 0.5922 & 14.743 & 70.13\% \\

\midrule

TeaCache-slow
& 13.29 & 2.63$\times$ & 1543.07 & 29.39
& 0.1730 & 0.8041 & 23.126 & \bf 75.90\% \\

SenCache-slow
& 13.55 & 2.58$\times$ & 1354.32 & 29.39
& 0.1197 & 0.8608 & 26.066 & 75.57\% \\

\rowcolor{gray!15}
\textbf{PoSTCache-slow}
& \bf 13.25
& \bf 2.64$\times$
& 1542.24 & 29.39
& \textbf{0.0996}
& \textbf{0.8741}
& \textbf{27.694}
& 75.67\%\\

\midrule

TeaCache-fast
& 11.31 & 3.09$\times$ & 1259.94 & 29.39
& 0.2271 & 0.7565 & 21.158 & 75.30\% \\

SenCache-fast
& 11.79 & 2.96$\times$ & 1165.57 & 29.39
& 0.2036 & 0.7831 & 22.837 & 73.66\% \\

\rowcolor{gray!15}
\textbf{PoSTCache-fast}
& \textbf{11.23}
& \textbf{3.11}$\times$
& 1259.03 & 29.39
& \bf 0.1412
& \bf 0.8271
& \bf 24.898
& \bf 75.44\%\\
\midrule
\midrule

\multicolumn{9}{c}{\textbf{OpenSora 1.2} (51 frames, $854{\times}480$)} \\
\midrule

\textbf{Original: 30 steps}
& 30.13
& 1.00$\times$ & 3481.73 & 21.70
& -- & -- & -- & 79.35\% \\

60\% steps
& 19.02
& 1.58$\times$ & 2168.73 & 21.70
& 0.2891 & 0.7181 & 17.263 & 78.01\% \\

30\% steps
& 10.65
& 2.83$\times$ & 1183.98 & 21.70
& 0.4327 & 0.6254 & 13.920 & 74.37\% \\

\midrule

TeaCache-slow
& 14.37 & 2.10$\times$ & 1512.43 & 22.11
& 0.2439 & 0.7475 & 19.460 & \bf 78.12\% \\

MagCache-slow
& 13.53 & 2.23$\times$ & 1512.43 & 21.91
& 0.1526 & 0.8262 & 23.444 & 77.54\% \\

\rowcolor{gray!15}
\textbf{PoSTCache-slow}
& \bf 13.51
& \bf 2.23$\times$
& 1512.23
& 21.73
& \textbf{0.1180}
& \textbf{0.8545}
& \textbf{25.767}
&  78.00\%\\

\midrule

TeaCache-fast
& 11.80 & 2.55$\times$ & 1293.62 & 22.11
& 0.2840 & 0.7216 & 18.605 & 77.03\% \\

MagCache-fast
& 11.72 & 2.57$\times$ & 1293.62 & 21.91
& 0.2068 & 0.7840 & 21.453 & 77.28\% \\

\rowcolor{gray!15}
\textbf{PoSTCache-fast}
& \textbf{11.70}
& \textbf{2.58}$\times$
& 1293.40
& 21.73
& \bf 0.1756
& \bf 0.8054
& \bf 23.197
& \bf 77.57\%\\

\bottomrule
\end{tabular}}
\end{table*}

\subsection{Main Results}
\label{sec:main_results}

\paragraph{Text-to-image generation.}
Table~\ref{tab:text_to_image} compares PoSTCache with representative caching
methods and direct step-reduction baselines on FLUX.1-dev and
Qwen-Image-2512. Across both models and acceleration regimes, PoSTCache
consistently improves reconstruction fidelity at comparable or higher
speedups while largely preserving semantic quality, yielding the strongest
efficiency--fidelity trade-off overall. As illustrated in
Figure~\ref{fig:image_compare}, these quantitative gains are accompanied by
better preservation of object identity, geometric structure, and
fine-grained appearance than competing caching methods. Qualitative comparison on Qwen-Image-2512 is provided in Appendix~\ref{app:additional_qualitative}.

\paragraph{Text-to-video generation.}
Table~\ref{tab:text_to_video} reports the results on Wan~2.1, CogVideoX, and
OpenSora~1.2. PoSTCache likewise achieves consistently stronger
reconstruction fidelity under comparable computation budgets while
maintaining competitive overall VBench quality. Its advantage persists
across both moderate and aggressive acceleration settings, as well as across
different architectures and resolutions, demonstrating
the effectiveness of PoSTCache in limiting accumulated
spatiotemporal trajectory errors. Additional qualitative comparisons are
provided in Appendix~\ref{app:additional_qualitative}.

\subsection{Ablation Studies}
\label{sec:ablation_studies}

\begin{table*}[t]
    \centering
    \caption{
        Ablation studies on FLUX.1-dev.
        (a) Contributions of postdictive smoothing and anchor selection
        under $B=16$.
        (b) Generality across different forward caching strategies.
    }
    \label{tab:ablations}

    \scriptsize
    \setlength{\tabcolsep}{3.2pt}
    \renewcommand{\arraystretch}{1.08}

    \begin{tabular}{
        @{} l c c c c
        @{\hspace{18pt}}
        l c c c c @{}
    }
        \toprule

        \multicolumn{5}{c}{\textbf{(a) Component Analysis}}
        &
        \multicolumn{5}{c}{\textbf{(b) Forward-Strategy Generality}}
        \\

        \cmidrule(lr){1-5}
        \cmidrule(lr){6-10}

        \textbf{Selector}
        & \textbf{Smooth.}
        & \textbf{LPIPS}$\downarrow$
        & \textbf{SSIM}$\uparrow$
        & \textbf{PSNR}$\uparrow$
        &
        \textbf{Variant}
        & \textbf{Speed}$\uparrow$
        & \textbf{LPIPS}$\downarrow$
        & \textbf{SSIM}$\uparrow$
        & \textbf{PSNR}$\uparrow$
        \\

        \midrule

        Fixed Interval
        & \xmark
        & 0.2639
        & 0.7498
        & 19.539
        &
        MagCache
        & 2.71$\times$
        & 0.1468
        & 0.8477
        & 23.507
        \\

        Fixed Interval
        & \cmark
        & 0.2316
        & 0.7776
        & 20.017
        &
        + Smoothing
        & 2.68$\times$
        & 0.1259
        & 0.8635
        & 24.319
        \\

        MagCache
        & \xmark
        & 0.1466
        & 0.8488
        & 23.589
        &
        \textbf{+ Smoothing + PVR}
        & \textbf{2.71$\times$}
        & \textbf{0.0967}
        & \textbf{0.8943}
        & \textbf{26.267}
        \\

        MagCache
        & \cmark
        & 0.1256
        & 0.8635
        & 24.352
        &
        DiCache
        & \bf 2.65$\times$
        & 0.1804
        & 0.8326
        & 23.612
        \\

        PVR
        & \xmark
        & 0.1293
        & 0.8712
        & 25.287
        &
        + Smoothing
        & 2.59$\times$
        & 0.1696
        & 0.8341
        & 23.662
        \\

        \textbf{PVR}
        & \cmark
        & \textbf{0.0968}
        & \textbf{0.8944}
        & \textbf{26.284}
        &
        \textbf{+ Smoothing + PVR}
        & 2.62$\times$
        & \textbf{0.0741}
        & \textbf{0.9044}
        & \textbf{27.157}
        \\

        \bottomrule
    \end{tabular}
\end{table*}

    
    

\paragraph{Component Analysis.}
Table~\ref{tab:ablations}(a) evaluates the contributions of postdictive
smoothing and anchor selection under the same computation budget.
Postdictive smoothing consistently improves all three selection strategies,
confirming that anchor innovations provide effective corrective
information for skipped predictions. The PVR selector also outperforms fixed
intervals and the MagCache selector, and combining PVR selection with
postdictive smoothing yields the best performance.

\paragraph{Generality across Forward Caching Strategies.}
Table~\ref{tab:ablations}(b) evaluates PoSTCache with MagCache and DiCache
as forward caching strategies. Postdictive smoothing improves reconstruction
fidelity with only marginal runtime overhead, while combining it with PVR
selection provides substantial additional gains. These results demonstrate
that PoSTCache is compatible with different forward proposal mechanisms.

\paragraph{Additional Analysis.}

Appendices~\ref{app:calibration_analysis}--\ref{app:efficiency_analysis} further examine calibration-size robustness, validate the PVR objective and schedule transferability, compare alternative global scheduling objectives, and analyze temporal weighting, dense-prefix anchors, and anchor-state
consistency. These results support our default design choices, quantify the modest one-time offline calibration cost, and confirm negligible online overhead and favorable quality--efficiency trade-offs.

    
    


    
    

\section{Conclusion and Future Work}

In this work, we presented PoSTCache, a training-free and plug-and-play framework that reformulates diffusion caching as a revisable trajectory inference problem. By using feedback from subsequent full-computation steps, PoSTCache retrospectively refines skipped predictions and reintegrates the corrected sampling trajectory. Viewing anchors as sources of corrective feedback, we further introduced a posterior variance reduction criterion to select informative anchors according to their expected reduction in surrogate trajectory uncertainty. Extensive experiments on representative image and video generation models demonstrate that PoSTCache consistently improves the efficiency--fidelity trade-off and can be readily integrated with different forward caching strategies. Currently, PoSTCache estimates a shared anchor configuration from offline trajectories and applies it uniformly to all samples. Although this design is simple and effective, it may not fully capture sample-specific trajectory dynamics. A promising direction for future work is to combine the offline uncertainty structure with lightweight online observations, enabling the anchor schedule to adapt to each sampling trajectory without introducing substantial inference overhead.
\bibliography{iclr2027_conference}

@article{song2019generative,
  title={Generative modeling by estimating gradients of the data distribution},
  author={Song, Yang and Ermon, Stefano},
  journal={Advances in neural information processing systems},
  volume={32},
  year={2019}
}

@article{song2020score,
  title={Score-based generative modeling through stochastic differential equations},
  author={Song, Yang and Sohl-Dickstein, Jascha and Kingma, Diederik P and Kumar, Abhishek and Ermon, Stefano and Poole, Ben},
  journal={arXiv preprint arXiv:2011.13456},
  year={2020}
}

@article{ho2020denoising,
  title={Denoising diffusion probabilistic models},
  author={Ho, Jonathan and Jain, Ajay and Abbeel, Pieter},
  journal={Advances in neural information processing systems},
  volume={33},
  pages={6840--6851},
  year={2020}
}

@inproceedings{lipman2023flow,
  title={Flow Matching for Generative Modeling},
  author={Lipman, Yaron and Chen, Ricky TQ and Ben-Hamu, Heli and Nickel, Maximilian and Le, Matt},
  booktitle={The Eleventh International Conference on Learning Representations (ICLR)},
  year={2023}
}

@inproceedings{liu2023flow,
  title={Flow Straight and Fast: Learning to Generate and Transfer Data with Rectified Flow},
  author={Liu, Xingchao and Gong, Chengyue and Liu, Qiang},
  booktitle={The Eleventh International Conference on Learning Representations (ICLR)},
  year={2023}
}

@inproceedings{rombach2022high,
  title={High-resolution image synthesis with latent diffusion models},
  author={Rombach, Robin and Blattmann, Andreas and Lorenz, Dominik and Esser, Patrick and Ommer, Bj{\"o}rn},
  booktitle={Proceedings of the IEEE/CVF conference on computer vision and pattern recognition},
  pages={10684--10695},
  year={2022}
}

@inproceedings{podell2024sdxl,
  title={Sdxl: Improving latent diffusion models for high-resolution image synthesis},
  author={Podell, Dustin and English, Zion and Lacey, Kyle and Blattmann, Andreas and Dockhorn, Tim and M{\"u}ller, Jonas and Penna, Joe and Rombach, Robin},
  booktitle={International Conference on Learning Representations},
  volume={2024},
  pages={1862--1874},
  year={2024}
}

@article{blattmann2023stable,
  title={Stable video diffusion: Scaling latent video diffusion models to large datasets},
  author={Blattmann, Andreas and Dockhorn, Tim and Kulal, Sumith and Mendelevitch, Daniel and Kilian, Maciej and Lorenz, Dominik and Levi, Yam and English, Zion and Voleti, Vikram and Letts, Adam and others},
  journal={arXiv preprint arXiv:2311.15127},
  year={2023}
}

@inproceedings{chen2024videocrafter2,
  title={Videocrafter2: Overcoming data limitations for high-quality video diffusion models},
  author={Chen, Haoxin and Zhang, Yong and Cun, Xiaodong and Xia, Menghan and Wang, Xintao and Weng, Chao and Shan, Ying},
  booktitle={Proceedings of the IEEE/CVF conference on computer vision and pattern recognition},
  pages={7310--7320},
  year={2024}
}

@inproceedings{song2023consistency,
  title={Consistency models},
  author={Song, Yang and Dhariwal, Prafulla and Chen, Mark and Sutskever, Ilya},
  booktitle={Proceedings of the 40th International Conference on Machine Learning},
  pages={32211--32252},
  year={2023}
}

@article{li2023q,
  title={Q-dm: An efficient low-bit quantized diffusion model},
  author={Li, Yanjing and Xu, Sheng and Cao, Xianbin and Sun, Xiao and Zhang, Baochang},
  journal={Advances in neural information processing systems},
  volume={36},
  pages={76680--76691},
  year={2023}
}

@article{selvaraju2024fora,
  title={Fora: Fast-forward caching in diffusion transformer acceleration},
  author={Selvaraju, Pratheba and Ding, Tianyu and Chen, Tianyi and Zharkov, Ilya and Liang, Luming},
  journal={arXiv preprint arXiv:2407.01425},
  year={2024}
}

@misc{fang2023structural,
      title={Structural Pruning for Diffusion Models}, 
      author={Gongfan Fang and Xinyin Ma and Xinchao Wang},
      year={2023},
      eprint={2305.10924},
      archivePrefix={arXiv},
      primaryClass={cs.LG},
      url={https://arxiv.org/abs/2305.10924}, 
}

@inproceedings{meng2023distillation,
  title={On distillation of guided diffusion models},
  author={Meng, Chenlin and Rombach, Robin and Gao, Ruiqi and Kingma, Diederik and Ermon, Stefano and Ho, Jonathan and Salimans, Tim},
  booktitle={Proceedings of the IEEE/CVF conference on computer vision and pattern recognition},
  pages={14297--14306},
  year={2023}
}

@article{dao2022flashattention,
  title={Flashattention: Fast and memory-efficient exact attention with io-awareness},
  author={Dao, Tri and Fu, Dan and Ermon, Stefano and Rudra, Atri and R{\'e}, Christopher},
  journal={Advances in neural information processing systems},
  volume={35},
  pages={16344--16359},
  year={2022}
}

@article{bolya2022token,
  title={Token merging: Your vit but faster},
  author={Bolya, Daniel and Fu, Cheng-Yang and Dai, Xiaoliang and Zhang, Peizhao and Feichtenhofer, Christoph and Hoffman, Judy},
  journal={arXiv preprint arXiv:2210.09461},
  year={2022}
}

@article{song2020denoising,
  title={Denoising diffusion implicit models},
  author={Song, Jiaming and Meng, Chenlin and Ermon, Stefano},
  journal={arXiv preprint arXiv:2010.02502},
  year={2020}
}

@article{lu2022dpm,
  title={Dpm-solver: A fast ode solver for diffusion probabilistic model sampling in around 10 steps},
  author={Lu, Cheng and Zhou, Yuhao and Bao, Fan and Chen, Jianfei and Li, Chongxuan and Zhu, Jun},
  journal={Advances in neural information processing systems},
  volume={35},
  pages={5775--5787},
  year={2022}
}

@article{salimans2022progressive,
  title={Progressive distillation for fast sampling of diffusion models},
  author={Salimans, Tim and Ho, Jonathan},
  journal={arXiv preprint arXiv:2202.00512},
  year={2022}
}

@inproceedings{frans2025one,
  title={One step diffusion via shortcut models},
  author={Frans, Kevin and Hafner, Danijar and Levine, Sergey and Abbeel, Pieter},
  booktitle={International Conference on Learning Representations},
  volume={2025},
  pages={34668--34684},
  year={2025}
}

@article{geng2026mean,
  title={Mean flows for one-step generative modeling},
  author={Geng, Zhengyang and Deng, Mingyang and Bai, Xingjian and Kolter, Zico and He, Kaiming},
  journal={Advances in Neural Information Processing Systems},
  volume={38},
  pages={75460--75482},
  year={2026}
}

@inproceedings{ma2024deepcache,
  title={Deepcache: Accelerating diffusion models for free},
  author={Ma, Xinyin and Fang, Gongfan and Wang, Xinchao},
  booktitle={Proceedings of the IEEE/CVF conference on computer vision and pattern recognition},
  pages={15762--15772},
  year={2024}
}

@article{li2024faster,
  title={Faster diffusion: Rethinking the role of the encoder for diffusion model inference},
  author={Li, Senmao and Hu, Taihang and van de Weijer, Joost and Khan, Fahad S and Liu, Tao and Li, Linxuan and Yang, Shiqi and Wang, Yaxing and Cheng, Ming-Ming and Yang, Jian},
  journal={Advances in Neural Information Processing Systems},
  volume={37},
  pages={85203--85240},
  year={2024}
}

@article{chen2024delta,
  title={{$\Delta$-DiT}: A Training-Free Acceleration Method Tailored for Diffusion Transformers},
  author={Chen, Pengtao and Shen, Mingzhu and Ye, Peng and Cao, Jianjian and Tu, Chongjun and Bouganis, Christos-Savvas and Zhao, Yiren and Chen, Tao},
  journal={arXiv preprint arXiv:2406.01125},
  year={2024}
}

@inproceedings{cui2026dpcache,
  title={Denoising as Path Planning: Training-Free Acceleration of Diffusion Models with DPCache},
  author={Cui, Bowen and Wang, Yuanbin and Xu, Huajiang and Chen, Biaolong and Zhang, Aixi and Jiang, Hao and Jin, Zhengzheng and Liu, Xu and Huang, Pipei},
  booktitle={Proceedings of the IEEE/CVF Conference on Computer Vision and Pattern Recognition},
  pages={43632--43642},
  year={2026}
}

@article{gao2026meancache,
  title={MeanCache: From Instantaneous to Average Velocity for Accelerating Flow Matching Inference},
  author={Gao, Huanlin and Chen, Ping and Shi, Fuyuan and Wu, Ruijia and YanTao, Li and Hui, Qiang and You, Yuren and Lu, Ting and Tan, Chao and Zhao, Shaoan and others},
  journal={arXiv preprint arXiv:2601.19961},
  year={2026}
}

@article{kalman1960new,
  title={A New Approach to Linear Filtering and Prediction Problems},
  author={Kalman, RE},
  journal={Journal of Basic Engineering},
  volume={82},
  number={1},
  pages={35--45},
  year={1960},
  publisher={ASME International}
}

@article{rauch1965maximum,
  title={Maximum likelihood estimates of linear dynamic systems},
  author={Rauch, Herbert E and Tung, Fong and Striebel, Charlotte T},
  journal={AIAA journal},
  volume={3},
  number={8},
  pages={1445--1450},
  year={1965}
}

@article{rabiner1989tutorial,
  title={A tutorial on hidden Markov models and selected applications in speech recognition},
  author={Rabiner, Lawrence R},
  journal={Proceedings of the IEEE},
  volume={77},
  number={2},
  pages={257--286},
  year={1989},
  publisher={Ieee}
}

@book{sarkka2023bayesian,
  title={Bayesian filtering and smoothing},
  author={S{\"a}rkk{\"a}, Simo and Svensson, Lennart},
  volume={17},
  year={2023},
  publisher={Cambridge university press}
}

@article{saharia2022photorealistic,
  title={Photorealistic text-to-image diffusion models with deep language understanding},
  author={Saharia, Chitwan and Chan, William and Saxena, Saurabh and Li, Lala and Whang, Jay and Denton, Emily L and Ghasemipour, Kamyar and Gontijo Lopes, Raphael and Karagol Ayan, Burcu and Salimans, Tim and others},
  journal={Advances in neural information processing systems},
  volume={35},
  pages={36479--36494},
  year={2022}
}

@inproceedings{huang2024vbench,
  title={Vbench: Comprehensive benchmark suite for video generative models},
  author={Huang, Ziqi and He, Yinan and Yu, Jiashuo and Zhang, Fan and Si, Chenyang and Jiang, Yuming and Zhang, Yuanhan and Wu, Tianxing and Jin, Qingyang and Chanpaisit, Nattapol and others},
  booktitle={Proceedings of the IEEE/CVF Conference on Computer Vision and Pattern Recognition},
  pages={21807--21818},
  year={2024}
}

@article{bu2025dicache,
  title={Dicache: Let diffusion model determine its own cache},
  author={Bu, Jiazi and Ling, Pengyang and Zhou, Yujie and Wang, Yibin and Zang, Yuhang and Lin, Dahua and Wang, Jiaqi},
  journal={arXiv preprint arXiv:2508.17356},
  year={2025}
}

@inproceedings{zhang2018unreasonable,
  title={The unreasonable effectiveness of deep features as a perceptual metric},
  author={Zhang, Richard and Isola, Phillip and Efros, Alexei A and Shechtman, Eli and Wang, Oliver},
  booktitle={Proceedings of the IEEE conference on computer vision and pattern recognition},
  pages={586--595},
  year={2018}
}

@article{wang2004image,
  title={Image quality assessment: from error visibility to structural similarity},
  author={Wang, Zhou and Bovik, Alan C and Sheikh, Hamid R and Simoncelli, Eero P},
  journal={IEEE transactions on image processing},
  volume={13},
  number={4},
  pages={600--612},
  year={2004},
  publisher={IEEE}
}

@inproceedings{liu2025timestep,
  title={Timestep Embedding Tells: It's Time to Cache for Video Diffusion Model},
  author={Liu, Feng and Zhang, Shiwei and Wang, Xiaofeng and Wei, Yujie and Qiu, Haonan and Zhao, Yuzhong and Zhang, Yingya and Ye, Qixiang and Wan, Fang},
  booktitle={Proceedings of the Computer Vision and Pattern Recognition Conference},
  pages={7353--7363},
  year={2025}
}

@inproceedings{yang2025cogvideox,
  title={Cogvideox: Text-to-video diffusion models with an expert transformer},
  author={Yang, Zhuoyi and Teng, Jiayan and Zheng, Wendi and Ding, Ming and Huang, Shiyu and Xu, Jiazheng and Yang, Yuanming and Hong, Wenyi and Zhang, Xiaohan and Feng, Guanyu and others},
  booktitle={International Conference on Learning Representations},
  volume={2025},
  pages={83048--83077},
  year={2025}
}

@article{wang2025wan,
  title={Wan: Open and advanced large-scale video generative models},
  author={Wan, Team and Wang, Ang and Ai, Baole and Wen, Bin and Mao, Chaojie and Xie, Chen-Wei and Chen, Di and Yu, Feiwu and Zhao, Haiming and Yang, Jianxiao and others},
  journal={arXiv preprint arXiv:2503.20314},
  year={2025}
}

@article{zheng2024open,
  title={Open-sora: Democratizing efficient video production for all},
  author={Zheng, Zangwei and Peng, Xiangyu and Yang, Tianji and Shen, Chenhui and Li, Shenggui and Liu, Hongxin and Zhou, Yukun and Li, Tianyi and You, Yang},
  journal={arXiv preprint arXiv:2412.20404},
  year={2024}
}

@article{wu2025qwen,
  title={Qwen-image technical report},
  author={Wu, Chenfei and Li, Jiahao and Zhou, Jingren and Lin, Junyang and Gao, Kaiyuan and Yan, Kun and Yin, Sheng-ming and Bai, Shuai and Xu, Xiao and Chen, Yilei and others},
  journal={arXiv preprint arXiv:2508.02324},
  year={2025}
}

@misc{flux2024,
    author={Black Forest Labs},
    title={FLUX},
    year={2024},
    howpublished={\url{https://github.com/black-forest-labs/flux}},
}

@article{ma2026magcache,
  title={Magcache: Fast video generation with magnitude-aware cache},
  author={Ma, Zehong and Wei, Longhui and Wang, Feng and Zhang, Shiliang and Tian, Qi},
  journal={Advances in Neural Information Processing Systems},
  volume={38},
  pages={34348--34380},
  year={2026}
}

@inproceedings{liu2025reusing,
  title={From reusing to forecasting: Accelerating diffusion models with taylorseers},
  author={Liu, Jiacheng and Zou, Chang and Lyu, Yuanhuiyi and Chen, Junjie and Zhang, Linfeng},
  booktitle={Proceedings of the IEEE/CVF International Conference on Computer Vision},
  pages={15853--15863},
  year={2025}
}

@inproceedings{feng2026hicache,
  title={HiCache: A Plug-in Scaled-Hermite Upgrade for Taylor-Style Cache-then-Forecast Diffusion Acceleration},
  author={Feng, Liang and Zheng, Shikang and Liu, Jiacheng and Lin, Yuqi and Zhou, Qinming and Cai, Peiliang and Wang, Xinyu and Chen, Junjie and Zou, Chang and Ma, Yue and Zhang, Linfeng},
  booktitle={International Conference on Learning Representations (ICLR)},
  year={2026}
}

@article{haghighi2026sencache,
  title={SenCache: Accelerating Diffusion Model Inference via Sensitivity-Aware Caching},
  author={Haghighi, Yasaman and Alahi, Alexandre},
  journal={arXiv preprint arXiv:2602.24208},
  year={2026}
}

@article{chung2026seacache,
  title={SeaCache: Spectral-Evolution-Aware Cache for Accelerating Diffusion Models},
  author={Chung, Jiwoo and Hyun, Sangeek and Lee, MinKyu and Han, Byeongju and Cha, Geonho and Wee, Dongyoon and Hong, Youngjun and Heo, Jae-Pil},
  journal={arXiv preprint arXiv:2602.18993},
  year={2026}
}

@inproceedings{hessel2021clipscore,
  title={Clipscore: A reference-free evaluation metric for image captioning},
  author={Hessel, Jack and Holtzman, Ari and Forbes, Maxwell and Le Bras, Ronan and Choi, Yejin},
  booktitle={Proceedings of the 2021 conference on empirical methods in natural language processing},
  pages={7514--7528},
  year={2021}
}

@article{xu2023imagereward,
  title={Imagereward: Learning and evaluating human preferences for text-to-image generation},
  author={Xu, Jiazheng and Liu, Xiao and Wu, Yuchen and Tong, Yuxuan and Li, Qinkai and Ding, Ming and Tang, Jie and Dong, Yuxiao},
  journal={Advances in Neural Information Processing Systems},
  volume={36},
  pages={15903--15935},
  year={2023}
}

@article{ho2022classifier,
  title={Classifier-free diffusion guidance},
  author={Ho, Jonathan and Salimans, Tim},
  journal={arXiv preprint arXiv:2207.12598},
  year={2022}
}

@article{yu2022scaling,
  title={Scaling Autoregressive Models for Content-Rich Text-to-Image Generation},
  author={Yu, Jiahui and Xu, Yuanzhong and Koh, Jing Yu and Luong, Thang and Baid, Gunjan and Wang, Zirui and Vasudevan, Vijay and Ku, Alexander and Yang, Yinfei and Ayan, Burcu Karagol and others},
  journal={arXiv preprint arXiv:2206.10789},
  year={2022}
}

@inproceedings{peng2026ertacache,
  title={Ertacache: Error rectification and timesteps adjustment for efficient diffusion},
  author={Peng, Xurui and Yan, Chenqian and Liu, Hong and Ma, Rui and Chen, Fangmin and Wang, Xing and Wu, Zhihua and Liu, Songwei and Lin, Mingbao},
  booktitle={International Conference on Learning Representations},
  volume={2026},
  pages={82104--82120},
  year={2026}
}
\bibliographystyle{iclr2027_conference}

\appendix
\section*{Appendix Overview}

We organize the appendix as follows.

\paragraph{Theoretical Background and Derivations.}
\begin{itemize}
    \item Section~\ref{app:trajectory_estimation_view} introduces a sequential trajectory-estimation view of diffusion sampling.
    \item Section~\ref{app:bayesian_smoothing} reviews linear Gaussian filtering
    and smoothing.
    \item Section~\ref{app:postdictive_derivation} derives the postdictive
    smoothing formulation.
    \item Section~\ref{app:posterior_anchor_derivation} derives posterior
    variance reduction anchor selection.
\end{itemize}

\paragraph{Experimental Settings.}
\begin{itemize}
    \item Section~\ref{app:models_datasets} describes the evaluated models,
    datasets, and generation configurations.
    \item Section~\ref{app:postcache_configuration} reports the implementation
    and hyperparameter settings of PoSTCache.
    \item Section~\ref{app:baseline_settings} provides the configurations of
    the compared methods.
    \item Section~\ref{app:evaluation_protocol} describes the evaluation and
    runtime measurement protocols.
    \item Section~\ref{app:metrics} defines the evaluation metrics.
\end{itemize}

\paragraph{Additional Experimental Results and Analysis.}
\begin{itemize}
    \item Section~\ref{app:additional_trajectory_analysis} provides additional trajectory analysis and corresponding evaluation details.
    \item Section~\ref{app:calibration_analysis} studies calibration size and selector robustness.
    \item Section~\ref{app:pvr_surrogate_validation} directly validates PVR as a schedule-level surrogate on held-out trajectories.
    \item Section~\ref{app:pvr_transfer} evaluates the transferability of offline PVR schedules across prompt distributions, guidance scales, and spatial configurations.
    \item Section~\ref{app:additional_ablations} provides additional ablations.
    \item Section~\ref{app:anchor_consistency} analyzes anchor-state consistency after retrospective reintegration and the compute trade-off of explicit anchor refresh.
    \item Section~\ref{app:efficiency_analysis} analyzes offline calibration cost, online runtime overhead, and the efficiency--quality trade-off.
    \item Section~\ref{app:additional_qualitative} presents additional
    qualitative comparisons.
    \item Section~\ref{app:limitations} discusses limitations and boundary conditions.
\end{itemize}

\section{Theoretical Background and Derivations}
\label{app:sec_theoretical_derivations}
\subsection{Sequential Trajectory Estimation View of Diffusion and Flow Sampling}
\label{app:trajectory_estimation_view}

A broad class of diffusion, flow-matching, and rectified-flow samplers can be
expressed as a sequence of model predictions and numerical state updates:
\begin{equation}
    u_k
    =
    f_\theta(x_k,\tau_k),
    \qquad
    x_{k+1}
    =
    \mathcal{S}_k(x_k,u_k,\tau_k),
    \label{eq:generic_sampling_update}
\end{equation}
where $x_k$ is the latent state at the $k$-th sampling step, $\tau_k$ is the
corresponding noise or time level, $u_k$ is the model prediction consumed by
the sampler, and $\mathcal{S}_k$ denotes the scheduler or numerical solver.
Depending on the model parameterization, $u_k$ may represent noise, a clean
sample estimate, a velocity, or a flow vector.

Equation~(\ref{eq:generic_sampling_update}) defines a recursive trajectory
estimation process. At each step, the model predicts the local direction of
the trajectory from the current latent state, and the solver propagates this
prediction to the next state. This structure resembles sequential Bayesian
estimation, in which an estimate is repeatedly updated using the information
available at the current step.

Under cache-based acceleration, some full model predictions are replaced by
forward approximations. Let
\begin{equation}
    \widetilde{u}_k
    =
    u_k-e_k
    \label{eq:cached_prediction_error}
\end{equation}
denote the cached prediction, where $e_k$ is the corresponding approximation
error. The approximate sampling trajectory then evolves according to
\begin{equation}
    \widetilde{x}_{k+1}
    =
    \mathcal{S}_k
    \left(
        \widetilde{x}_k,
        \widetilde{u}_k,
        \tau_k
    \right).
    \label{eq:cached_sampling_update}
\end{equation}
Because each state is used as the input to subsequent model evaluations and
solver updates, an error introduced at one step can affect all later states.

To make this propagation explicit, define the latent-state deviation
$\delta x_k=\widetilde{x}_k-x_k$. A first-order expansion of the solver around
the full-computation trajectory gives
\begin{equation}
    \delta x_{k+1}
    \approx
    A_k\delta x_k
    -
    B_k e_k,
    \qquad
    A_k
    =
    \frac{\partial\mathcal{S}_k}{\partial x_k},
    \qquad
    B_k
    =
    \frac{\partial\mathcal{S}_k}{\partial u_k}.
    \label{eq:cache_error_propagation}
\end{equation}
Thus, the deviation at step $k+1$ depends not only on the current prediction
error but also on the deviation accumulated from previous steps. Repeatedly
applying Eq.~(\ref{eq:cache_error_propagation}) over a skipped interval
$a<k<b$ yields
\begin{equation}
    \delta x_b
    \approx
    A_{b-1}\cdots A_a\delta x_a
    -
    \sum_{j=a}^{b-1}
    \left(
        A_{b-1}\cdots A_{j+1}
    \right)
    B_j e_j,
    \label{eq:accumulated_cache_error}
\end{equation}
where an empty matrix product is understood as the identity. Consequently,
even individually small approximation errors may accumulate and alter the
subsequent sampling trajectory.

Suppose that step $b$ is selected as a full-computation anchor. A complete
model evaluation at the current latent state produces
$\bar{u}_b=f_\theta(\widetilde{x}_b,\tau_b)$, while the forward caching rule
also provides a proposal $\widetilde{u}_b$. Their discrepancy,
\begin{equation}
    r_b
    =
    \bar{u}_b-\widetilde{u}_b,
    \label{eq:appendix_anchor_innovation}
\end{equation}
is the \emph{anchor innovation}. Although $r_b$ does not directly
observe each preceding error $e_k$, it reveals the discrepancy between the
forward approximation and the newly available full computation. Since the
preceding cache errors jointly influence the latent state and the forward
proposal at step $b$, the innovation can contain information about the errors
accumulated over the skipped interval.

More formally, let the centered intermediate error $e_k$ and future
innovation $r_b$ be locally approximated by a joint Gaussian distribution.
Conditioning on the observed innovation gives
\begin{equation}
    \mathbb{E}
    \left[
        e_k \mid r_b
    \right]
    =
    \Sigma_{e_k r_b}
    \left(
        \Sigma_{r_b r_b}+R_b
    \right)^{-1}
    r_b,
    \label{eq:error_conditioned_on_innovation}
\end{equation}
where $\Sigma_{e_k r_b}$ is the cross-covariance between the intermediate
proposal error and the future innovation, and $R_b$ accounts for residual
uncertainty in the future evidence. The corresponding corrected prediction
therefore has the generic form
\begin{equation}
    \widehat{u}_k
    =
    \widetilde{u}_k
    +
    G_{k\leftarrow b}r_b,
    \qquad
    G_{k\leftarrow b}
    =
    \Sigma_{e_k r_b}
    \left(
        \Sigma_{r_b r_b}+R_b
    \right)^{-1}.
    \label{eq:trajectory_smoothing_view}
\end{equation}
That is, a forward estimate is revised by propagating a future innovation
backward through a covariance-dependent smoothing gain. This conditional
estimation view motivates the postdictive correction used by PoSTCache. The
standard linear Gaussian filtering and smoothing identities underlying
Eq.~(\ref{eq:trajectory_smoothing_view}) are reviewed in
Appendix~\ref{app:bayesian_smoothing}.

\subsection{Linear Gaussian Filtering and Smoothing}
\label{app:bayesian_smoothing}

Consider a linear Gaussian state-space model
\begin{equation}
\label{app:eq:linear_system}
    z_{t+1}=A_tz_t+w_t,
    \qquad
    y_t=H_tz_t+v_t,
\end{equation}
where
$w_t\sim\mathcal{N}(0,Q_t)$ and
$v_t\sim\mathcal{N}(0,R_t)$.

Given the filtering distribution
\begin{equation}
    p(z_t\mid y_{1:t})
    =
    \mathcal{N}(m_{t\mid t},P_{t\mid t}),
\end{equation}
the one-step predictive distribution is
\begin{equation}
    p(z_{t+1}\mid y_{1:t})
    =
    \mathcal{N}(m_{t+1\mid t},P_{t+1\mid t}),
\end{equation}
where
\begin{align}
    m_{t+1\mid t}
    &=
    A_tm_{t\mid t},\\
    P_{t+1\mid t}
    &=
    A_tP_{t\mid t}A_t^\top+Q_t.
\end{align}

The Rauch--Tung--Striebel smoother~\citep{rauch1965maximum}
performs a backward recursion after the forward filtering pass:
\begin{align}
\label{app:eq:rts_mean}
    m_{t\mid T}
    &=
    m_{t\mid t}
    +
    G_t
    \left(
        m_{t+1\mid T}-m_{t+1\mid t}
    \right),\\
\label{app:eq:rts_cov}
    P_{t\mid T}
    &=
    P_{t\mid t}
    +
    G_t
    \left(
        P_{t+1\mid T}-P_{t+1\mid t}
    \right)
    G_t^\top,
\end{align}
with smoothing gain
\begin{equation}
\label{app:eq:rts_gain}
    G_t
    =
    P_{t\mid t}A_t^\top P_{t+1\mid t}^{-1}.
\end{equation}
The innovation
$m_{t+1\mid T}-m_{t+1\mid t}$
measures the discrepancy between the future smoothed estimate and its forward
prediction, which is propagated backward to revise the current estimate.

More generally, suppose an intermediate state $z_t$ and future evidence $y_b$
follow a joint Gaussian distribution:
\begin{equation}
\label{app:eq:joint_gaussian}
    \begin{bmatrix}
        z_t\\
        y_b
    \end{bmatrix}
    \sim
    \mathcal{N}
    \left(
        \begin{bmatrix}
            m_t\\
            m_b
        \end{bmatrix},
        \begin{bmatrix}
            P_{tt} & P_{tb}\\
            P_{bt} & P_{bb}+R_b
        \end{bmatrix}
    \right).
\end{equation}
Applying the Gaussian conditioning identity gives
\begin{equation}
\label{app:eq:conditional_distribution}
    p(z_t\mid y_b)
    =
    \mathcal{N}
    \left(
        m_{t\mid b},
        P_{t\mid b}
    \right),
\end{equation}
where
\begin{align}
\label{app:eq:conditional_mean}
    m_{t\mid b}
    &=
    m_t+
    P_{tb}(P_{bb}+R_b)^{-1}(y_b-m_b),\\
\label{app:eq:conditional_covariance}
    P_{t\mid b}
    &=
    P_{tt}
    -
    P_{tb}(P_{bb}+R_b)^{-1}P_{bt}.
\end{align}

The conditional mean has the generic form
\begin{equation}
    \text{smoothed estimate}
    =
    \text{forward estimate}
    +
    \text{smoothing gain}
    \times
    \text{future innovation}.
\end{equation}
Meanwhile, the posterior covariance subtracts a positive-semidefinite term
from the prior covariance. Therefore, informative future evidence both
corrects the forward estimate and reduces its remaining uncertainty. This
linear Gaussian result motivates the postdictive correction and posterior
variance reduction formulations used in our method.

\subsection{Derivation of Postdictive Smoothing}
\label{app:postdictive_derivation}

Consider a cache interval bounded by two consecutive full-computation anchors
$a<b$. For an intermediate step $k$, define the proposal error as
\begin{equation}
    e_k
    =
    \bar{u}_k-\widetilde{u}_k,
    \qquad
    a<k<b,
    \label{app:eq:proposal_error}
\end{equation}
where $\bar{u}_k$ denotes the hypothetical full-computation prediction
evaluated at the same forward state.

We locally approximate the temporal evolution of the proposal errors by a
first-order stochastic system:
\begin{equation}
    e_{j+1}
    =
    F_j e_j+\xi_j,
    \qquad
    \xi_j\sim\mathcal{N}(0,Q_j),
    \label{app:eq:first_order_error_dynamics}
\end{equation}
where $F_j$ is the Jacobian of the local error transition around the forward
trajectory. Propagating this system from step $k$ to anchor $b$ gives
\begin{equation}
    e_b
    =
    \Phi_{b\leftarrow k}e_k+\xi_{k:b},
    \qquad
    \Phi_{b\leftarrow k}
    =
    F_{b-1}\cdots F_k,
    \label{app:eq:multistep_error_dynamics}
\end{equation}
where $\xi_{k:b}$ aggregates the process noise over the interval.

Assume that the intermediate proposal error follows a Gaussian prior
\begin{equation}
    e_k\sim\mathcal{N}(0,C_{kk}).
\end{equation}
At the future full-computation anchor, the innovation
\begin{equation}
    r_b
    =
    \bar{u}_b-\widetilde{u}_b
    =
    e_b
    \label{app:eq:anchor_innovation}
\end{equation}
directly reveals the terminal proposal error.

Equation~(\ref{app:eq:multistep_error_dynamics}) induces the joint Gaussian
distribution
\begin{equation}
    \begin{bmatrix}
        e_k\\
        r_b
    \end{bmatrix}
    \sim
    \mathcal{N}
    \left(
        0,
        \begin{bmatrix}
            C_{kk} & C_{kb}\\
            C_{bk} & C_{bb}
        \end{bmatrix}
    \right),
    \label{app:eq:joint_error_anchor}
\end{equation}
where
\begin{align}
    C_{kb}
    &=
    C_{kk}\Phi_{b\leftarrow k}^{\top},\\
    C_{bb}
    &=
    \Phi_{b\leftarrow k}
    C_{kk}
    \Phi_{b\leftarrow k}^{\top}
    +
    Q_{k:b},
\end{align}
and $Q_{k:b}$ denotes the covariance of the accumulated process noise.

Applying the Gaussian conditioning identity to
Eq.~(\ref{app:eq:joint_error_anchor}) yields
\begin{equation}
    p(e_k\mid r_b)
    =
    \mathcal{N}
    \left(
        G_{k\leftarrow b}r_b,
        C_{kk\mid b}
    \right),
    \label{app:eq:error_posterior}
\end{equation}
with smoothing gain
\begin{equation}
    G_{k\leftarrow b}
    =
    C_{kb}C_{bb}^{-1},
    \label{app:eq:smoothing_gain}
\end{equation}
and posterior covariance
\begin{equation}
    C_{kk\mid b}
    =
    C_{kk}
    -
    C_{kb}C_{bb}^{-1}C_{bk}.
    \label{app:eq:smoothed_covariance}
\end{equation}

The conditional mean is the minimum-mean-square-error estimate of $e_k$:
\begin{equation}
    \widehat{e}_k
    =
    \mathbb{E}[e_k\mid r_b]
    =
    G_{k\leftarrow b}r_b.
\end{equation}
Since
$\bar{u}_k=\widetilde{u}_k+e_k$, the corresponding estimate of the
full-computation prediction is
\begin{equation}
    \widehat{u}_k
    =
    \widetilde{u}_k
    +
    G_{k\leftarrow b}
    \left(
        \bar{u}_b-\widetilde{u}_b
    \right).
    \label{app:eq:bayesian_postdictive_correction}
\end{equation}

The covariance reduction induced by the future anchor is
\begin{equation}
    C_{kk}-C_{kk\mid b}
    =
    C_{kb}C_{bb}^{-1}C_{bk}
    \succeq 0.
    \label{app:eq:covariance_reduction}
\end{equation}
Thus, under the assumed linear Gaussian model, conditioning on an informative future anchor cannot increase the uncertainty of the intermediate proposal error.

Direct estimation of the full matrix gain in
Eq.~(\ref{app:eq:smoothing_gain}) is impractical for high-dimensional model
predictions. We therefore use the isotropic approximation
\begin{equation}
    G_{k\leftarrow b}
    \approx
    w_{k\mid a,b}I.
\end{equation}
Motivated by the empirical observation in Sec.~\ref{sec:trajectory_analysis} that the alignment between intermediate errors and the anchor innovation increases toward the future anchor, we set
\begin{equation}
    w_{k\mid a,b}
    =
    \frac{\eta_k-\eta_a}
         {\eta_b-\eta_a}.
\end{equation}
This yields the lightweight postdictive correction used in the main text:
\begin{equation}
    \widehat{u}_k
    =
    \widetilde{u}_k
    +
    w_{k\mid a,b}
    \left(
        \bar{u}_b-\widetilde{u}_b
    \right).
\end{equation}

\subsection{Derivation of Posterior Variance Reduction Selection}
\label{app:posterior_anchor_derivation}

\paragraph{Cross-timestep uncertainty model.}
For each offline sample $n$, we store the complete sequence of
full-computation model predictions
\begin{equation}
    z_k^{(n)}
    =
    \bar{u}_k^{(n)}
    \in \mathbb{R}^{D},
    \qquad
    k=1,\ldots,T,
    \label{app:eq:full_trajectory}
\end{equation}
where $\bar{u}_k^{(n)}$ is the prediction consumed by the numerical solver
after applying the same guidance and output-parameterization transformations
as in inference. All non-temporal dimensions are flattened without
projection. We define
the empirical mean and block covariance as
\begin{align}
    \mu_k
    &=
    \frac{1}{N}
    \sum_{n=1}^{N}
    z_k^{(n)},\\
    \Sigma_{ij}
    &=
    \frac{1}{N}
    \sum_{n=1}^{N}
    \left(z_i^{(n)}-\mu_i\right)
    \left(z_j^{(n)}-\mu_j\right)^\top.
    \label{app:eq:block_covariance}
\end{align}
Explicitly storing every block
$\Sigma_{ij}\in\mathbb{R}^{D\times D}$ is prohibitively expensive. We
therefore summarize each block by its normalized trace:
\begin{align}
    K_{ij}
    &=
    \frac{1}{D}
    \operatorname{tr}(\Sigma_{ij})\\
    &=
    \frac{1}{ND}
    \sum_{n=1}^{N}
    \left\langle
        z_i^{(n)}-\mu_i,\,
        z_j^{(n)}-\mu_j
    \right\rangle.
    \label{app:eq:cross_timestep_kernel}
\end{align}
Thus, $K\in\mathbb{R}^{T\times T}$ provides a lightweight timestep-level
summary of the covariance structure of the complete trajectories. The
resulting empirical kernel is positive semidefinite. In implementation, we
add a small fixed diagonal jitter before matrix solving for numerical
stability and use $K$ to denote the stabilized kernel below.

Let $\delta z_k=z_k-\mu_k$. We adopt the isotropic covariance surrogate
\begin{equation}
    \delta z_{1:T}
    \sim
    \mathcal{N}
    \left(
        0,\,
        K\otimes I_D
    \right).
    \label{app:eq:trajectory_prior}
\end{equation}
For an anchor set $A$, the corresponding centered trajectory states are
observed as
\begin{equation}
    y_A=\delta z_A.
    \label{app:eq:anchor_observation}
\end{equation}

\paragraph{Posterior mean and variance.}
The joint Gaussian distribution of the centered state at timestep $t$ and
the observed anchor states is
\begin{equation}
    \begin{bmatrix}
        \delta z_t\\
        y_A
    \end{bmatrix}
    \sim
    \mathcal{N}
    \left(
        0,\,
        \begin{bmatrix}
            K_{tt}I_D
            &
            K_{tA}\otimes I_D\\
            K_{At}\otimes I_D
            &
            K_{AA}\otimes I_D
        \end{bmatrix}
    \right).
\end{equation}
Gaussian conditioning gives
\begin{equation}
    p(\delta z_t\mid y_A)
    =
    \mathcal{N}
    \left(
        \widehat{\delta z}_t^{A},\,
        V_t(A)I_D
    \right),
\end{equation}
where
\begin{align}
    \widehat{\delta z}_t^{A}
    &=
    \left[
        K_{tA}
        K_{AA}^{-1}
        \otimes I_D
    \right]
    y_A,
    \label{app:eq:kernel_posterior_mean}\\
    V_t(A)
    &=
    K_{tt}
    -
    K_{tA}
    K_{AA}^{-1}
    K_{At}.
    \label{app:eq:kernel_posterior_variance}
\end{align}
For a single anchor $A=\{b\}$, the posterior mean reduces to
\begin{equation}
    \widehat{\delta z}_t^{\{b\}}
    =
    \frac{K_{tb}}{K_{bb}}
    y_b.
\end{equation}
The posterior variance measures the trajectory uncertainty remaining after
conditioning on the selected anchors. Under the isotropic surrogate, the
corresponding conditional mean-squared error is
\begin{equation}
    \mathbb{E}
    \left[
        \left\|
            \delta z_t-\widehat{\delta z}_t^{A}
        \right\|_2^2
        \,\middle|\,
        y_A
    \right]
    =
    D\,V_t(A).
    \label{app:eq:posterior_mse}
\end{equation}

\paragraph{Connection to postdictive correction.}
Consider a candidate anchor $s\notin A$. We define the residual error at
timestep $t$ and the innovation introduced by candidate $s$ as
\begin{equation}
    \epsilon_t^{A}
    =
    \delta z_t-\widehat{\delta z}_t^{A},
    \qquad
    \nu_s^{A}
    =
    y_s-\widehat{\delta z}_s^{A}.
    \label{app:eq:conditional_residual_innovation}
\end{equation}
Conditioned on the current anchor observations $y_A$, these quantities follow
the joint Gaussian distribution
\begin{equation}
    (\begin{bmatrix}
        \epsilon_t^{A}\\
        \nu_s^{A}
    \end{bmatrix}
    \,\mid\,
    y_A)
    \sim
    \mathcal{N}
    \left(
        0,\,
        \begin{bmatrix}
            V_t(A)
            &
            C_{ts\mid A}\\
            C_{ts\mid A}
            &
            V_s(A)
        \end{bmatrix}
        \otimes I_D
    \right),
    \label{app:eq:residual_innovation_joint}
\end{equation}
where
\begin{equation}
    C_{ts\mid A}
    =
    K_{ts}
    -
    K_{tA}
    K_{AA}^{-1}
    K_{As}.
    \label{app:eq:conditional_cross_covariance}
\end{equation}
Therefore, $C_{ts\mid A}I_D$ is precisely the conditional covariance between
the residual error at timestep $t$ and the innovation of candidate anchor
$s$ under the trajectory surrogate.

Conditioning additionally on $\nu_s^{A}$ gives the posterior-mean update
\begin{equation}
    \widehat{\delta z}_t^{A\cup\{s\}}
    =
    \widehat{\delta z}_t^{A}
    +
    \frac{C_{ts\mid A}}{V_s(A)}
    \nu_s^{A}.
    \label{app:eq:conditional_postdictive_update}
\end{equation}
This update has the same form as the linear Gaussian postdictive correction
in Eq.~(\ref{eq:bayesian_postdictive_correction}), with
$C_{ts\mid A}/V_s(A)$ acting as an isotropic smoothing gain.

The corresponding posterior variance satisfies the rank-one update
\begin{equation}
    V_t(A\cup\{s\})
    =
    V_t(A)
    -
    \frac{
        C_{ts\mid A}^{\,2}
    }{
        V_s(A)
    }.
    \label{app:eq:rank_one_update}
\end{equation}
Consequently, observing candidate anchor $s$ reduces the conditional
mean-squared residual error at timestep $t$ by
\begin{equation}
    D
    \left[
        V_t(A)-V_t(A\cup\{s\})
    \right]
    =
    D
    \frac{
        C_{ts\mid A}^{\,2}
    }{
        V_s(A)
    }.
    \label{app:eq:residual_mse_reduction}
\end{equation}
Thus, $C_{ts\mid A}^{\,2}/V_s(A)$ directly measures the expected residual
variance reduction provided by candidate anchor $s$.

The conditional mean above is a statistical surrogate induced by the
complete-trajectory covariance and need not coincide with the specific
forward proposal used by a practical caching method. Accordingly, the
resulting criterion should be interpreted as a proposal-agnostic measure of
intrinsic anchor informativeness.

\paragraph{Budgeted anchor selection.}
Given a full-computation budget $\mathcal{B}$, we minimize the weighted
uncertainty of the complete trajectory:
\begin{equation}
    A^\star
    =
    \arg\min_{\substack{
        A_0\subseteq A\\
        |A|\leq\mathcal{B}
    }}
    \sum_{t=1}^{T}
    \gamma_t V_t(A),
    \label{app:eq:anchor_selection_objective}
\end{equation}
where $A_0$ contains the dense-prefix anchors and the final denoising step,
and $\gamma_t$ is a scheduler-aware temporal weight.

By Eq.~(\ref{app:eq:rank_one_update}), the marginal reduction in weighted
trajectory uncertainty obtained by adding candidate $s$ is
\begin{equation}
    \Delta(s\mid A)
    =
    \sum_{t=1}^{T}
    \gamma_t
    \frac{
        C_{ts\mid A}^{\,2}
    }{
        V_s(A)
    }.
    \label{app:eq:posterior_variance_gain}
\end{equation}
The common factor $D$ in the corresponding mean-squared error reduction does
not affect candidate ranking and is therefore omitted. At each iteration, we
select
\begin{equation}
    s^\star
    =
    \arg\max_{s\notin A}
    \Delta(s\mid A),
\end{equation}
update the posterior covariance, and repeat until the computation budget is
exhausted.

\section{Experimental Settings}
\label{app:experimental_settings}

\subsection{Models, Datasets, and Generation Configurations}
\label{app:models_datasets}

We evaluate PoSTCache on five representative image and video generation
models. For text-to-image generation, we use FLUX.1-dev~\citep{flux2024}
and Qwen-Image-2512~\citep{wu2025qwen}, evaluated on
DrawBench~\citep{saharia2022photorealistic}. For text-to-video generation,
we use Wan~2.1~\citep{wang2025wan}, CogVideoX~\citep{yang2025cogvideox},
and OpenSora~1.2~\citep{zheng2024open}, evaluated on
VBench~\citep{huang2024vbench} prompts.

Table~\ref{tab:app_model_configurations} summarizes the generation
configuration of each model. CogVideoX is evaluated in FP16 following its officially recommended inference
configuration, while the other models use BF16. Unless otherwise specified,
all remaining parameters follow the default configurations of the
corresponding official implementations.

\begin{table*}[t]
    \centering
    \caption{Generation configurations of the evaluated models.}
    \label{tab:app_model_configurations}
    \setlength{\tabcolsep}{5pt}
    \resizebox{\textwidth}{!}{
    \begin{tabular}{lcccccccc}
        \toprule
        \textbf{Model}
        & \textbf{Task}
        & \textbf{Resolution}
        & \textbf{Frames}
        & \textbf{Full steps}
        & \textbf{CFG}
        & \textbf{Scheduler}
        & \textbf{Precision}
        & \textbf{Prompts} \\
        \midrule
        FLUX.1-dev
        & Image
        & $1024{\times}1024$
        & --
        & 50
        & --
        & \texttt{FlowMatch Euler}
        & \texttt{BF16}
        & \texttt{DrawBench} \\

        Qwen-Image-2512
        & Image
        & $1328{\times}1328$
        & --
        & 50
        & 4.0
        & \texttt{FlowMatch Euler}
        & \texttt{BF16}
        & \texttt{DrawBench} \\

        Wan~2.1 1.3B
        & Video
        & $832{\times}480$
        & 81
        & 50
        & 5.0
        & \texttt{UniPC (2nd order)}
        & \texttt{BF16}
        & \texttt{VBench} \\

        CogVideoX 2B
        & Video
        & $720{\times}480$
        & 49
        & 50
        & --
        & \texttt{DDIM (v-prediction)}
        & \texttt{FP16}
        & \texttt{VBench} \\

        OpenSora~1.2
        & Video
        & $854{\times}480$
        & 51
        & 30
        & --
        & \texttt{Rectified Flow (RFlow)}
        & \texttt{BF16}
        & \texttt{VBench} \\
        \bottomrule
    \end{tabular}}
\end{table*}

\subsection{PoSTCache Configuration}
\label{app:postcache_configuration}

PoSTCache consists of a forward proposal strategy, postdictive smoothing and reintegration, and posterior variance reduction anchor selection. Unless otherwise specified, the latest full-computation prediction is reused as the forward proposal. Within each cache interval $[a,b]$, the postdictive gain is set to the relative sampling progress $w_{k\mid a,b}=(\eta_k-\eta_a)/(\eta_b-\eta_a)$.

For offline anchor selection, we collect $N=10$ complete denoising
trajectories for each model. All non-temporal dimensions are flattened
without projection when constructing the cross-timestep covariance kernel.
The empirical kernel is normalized by both the number of calibration samples
and the trajectory dimension. 

\paragraph{Scheduler-aware temporal weighting.}
Many diffusion schedulers use non-uniform discretizations with relatively small
solver intervals in the early denoising stage and progressively larger
intervals later. Such dense sampling reflects the need to track the trajectory
at a finer temporal resolution. We therefore use the local solver spacing as a
scheduler-aware proxy for timestep importance, assigning larger weights to
more densely sampled regions rather than treating all denoising steps
uniformly. Let $\delta_k$ denote the local scheduler spacing at timestep
$k$. When the sampler exposes a noise-level sequence
$\{\sigma_k\}_{k=0}^{T}$, we use the exact transition spacing
\begin{equation}
    \delta_k
    =
    \left|
        \sigma_{k+1}-\sigma_k
    \right|,
    \qquad
    k=0,\ldots,T-1.
    \label{app:eq:sigma_spacing}
\end{equation}
Otherwise, given scheduler timesteps $\{\tau_k\}_{k=1}^{T}$, we approximate
the local spacing by
\begin{equation}
    \delta_k
    =
    \begin{cases}
        |\tau_2-\tau_1|,
        & k=1,\\[2pt]
        \frac{1}{2}
        \left(
            |\tau_k-\tau_{k-1}|
            +
            |\tau_{k+1}-\tau_k|
        \right),
        & 1<k<T,\\[2pt]
        |\tau_T-\tau_{T-1}|,
        & k=T.
    \end{cases}
    \label{app:eq:timestep_spacing}
\end{equation}
The unnormalized temporal weight is
\begin{equation}
    \widetilde{\gamma}_k
    =
    \left(
        \frac{
            \operatorname{median}_j \delta_j
        }{
            \delta_k+\epsilon
        }
    \right)^{p},
    \label{app:eq:raw_temporal_weight}
\end{equation}
where $p$ controls the weighting strength. We first normalize the weights to
unit mean, clip their range, and renormalize:
\begin{equation}
    \bar{\gamma}_k
    =
    \frac{
        \widetilde{\gamma}_k
    }{
        \frac{1}{T}\sum_{j=1}^{T}\widetilde{\gamma}_j
    },
    \qquad
    \gamma_k
    =
    \frac{
        \operatorname{clip}
        \left(
            \bar{\gamma}_k,\gamma_{\min},\gamma_{\max}
        \right)
    }{
        \frac{1}{T}
        \sum_{j=1}^{T}
        \operatorname{clip}
        \left(
            \bar{\gamma}_j,\gamma_{\min},\gamma_{\max}
        \right)
    }.
    \label{app:eq:temporal_weight}
\end{equation}
Unless otherwise specified, we use
$p=0.5$ and
$(\gamma_{\min},\gamma_{\max})=(0.5,2.0)$.

Detailed model-specific settings are reported in
Table~\ref{tab:app_postcache_configurations}.



For all models and variants, we use $N=10$ calibration trajectories and the same scheduler-aware temporal weighting. The final denoising step is always included as a mandatory anchor. The Fast and Slow variants differ only in the full-computation budget $\mathcal{B}$ and dense-prefix length.

\begin{table}[t]
    \centering
    \caption{Model-specific PoSTCache configurations. Dense-prefix lengths follow the empirical rule in Eq.~(\ref{app:eq:dense_prefix_rule}).}
    \label{tab:app_postcache_configurations}
    \setlength{\tabcolsep}{6pt}
    \begin{tabular}{lcccc}
        \toprule
        \multirow{2}{*}{\textbf{Model}}
        & \multicolumn{2}{c}{\textbf{Fast}}
        & \multicolumn{2}{c}{\textbf{Slow}} \\
        \cmidrule(lr){2-3}
        \cmidrule(lr){4-5}
        & \textbf{Budget $\mathcal{B}$}
        & \textbf{Dense prefix}
        & \textbf{Budget $\mathcal{B}$}
        & \textbf{Dense prefix} \\
        \midrule
        FLUX.1-dev
        & {13}
        & {6}
        & {16}
        & {7} \\

        Qwen-Image-2512
        & {14}
        & {7}
        & {16}
        & {7} \\

        Wan~2.1
        & {16}
        & {8}
        & {20}
        & {10} \\

        CogVideoX
        & {13}
        & {6}
        & {16}
        & {7} \\

        OpenSora~1.2
        & {10}
        & {5}
        & {12}
        & {6} \\
        \bottomrule
    \end{tabular}
\end{table}

\subsection{Baseline Generation Settings}
\label{app:baseline_settings}

We compare PoSTCache with representative training-free diffusion caching
methods, including TeaCache~\citep{liu2025timestep},
TaylorSeer~\citep{liu2025reusing},
MagCache~\citep{ma2026magcache},
DiCache~\citep{bu2025dicache},
HiCache~\citep{feng2026hicache},
SeaCache~\citep{chung2026seacache}, and
SenCache~\citep{haghighi2026sencache}.
We use the official implementations whenever available. For each baseline, we perform a lightweight runtime calibration on a small profiling set (typically two prompts), varying only its acceleration-control parameters to identify configurations with latency closest to the corresponding Fast and Slow regimes of PoSTCache. This calibration uses latency only and does not access any generation-quality metric. The selected configurations are then fixed and evaluated on the complete benchmark using the same prompt–seed pairs as the other methods. Tables~\ref{tab:app_image_baselines}
and~\ref{tab:app_video_baselines} report the complete configurations.

\begin{table*}[t]
    \centering
    \caption{Baseline configurations for text-to-image generation.}
    \label{tab:app_image_baselines}
    \setlength{\tabcolsep}{6pt}
    \renewcommand{\arraystretch}{1.12}
    \begin{tabular}{lll p{6.8cm}}
        \toprule
        \textbf{Model}
        & \textbf{Method}
        & \textbf{Regime}
        & \textbf{Configuration} \\
        \midrule

        \multirow{8}{*}{FLUX.1-dev}
        & \multirow{1}{*}{TeaCache}
        & Slow
        & $\delta=0.58$ \\
        \cmidrule(lr){2-4}

        & \multirow{1}{*}{TaylorSeer}
        & Slow
        & $\texttt{prefix}=3$, $N=5$, $O=1$ \\

        \cmidrule(lr){2-4}

        & \multirow{1}{*}{HiCache}
        & Slow
        & $\texttt{prefix}=3$, $N=5$, $O=1$, $\texttt{scale}=0.7$ \\

        \cmidrule(lr){2-4}

        & \multirow{1}{*}{SeaCache}
        & Slow
        & $\delta=0.5$ \\

        \cmidrule(lr){2-4}

        & \multirow{2}{*}{DiCache}
        & Fast
        & $\texttt{prefix\_ratio}=0.15$, $\delta=0.4$ \\
        &
        & Slow
        & $\texttt{prefix\_ratio}=0.2$, $\delta=0.4$ \\

        \cmidrule(lr){2-4}

        & \multirow{2}{*}{MagCache}
        & Fast
        & $\texttt{prefix\_ratio}=0.1$, $K=6$, $\delta=0.4$ \\
        &
        & Slow
        & $\texttt{prefix\_ratio}=0.1$, $K=4$, $\delta=0.2$ \\
        \midrule

        \multirow{6}{*}{Qwen-Image-2512}

        & \multirow{1}{*}{TaylorSeer}
        & Slow
        & $\texttt{prefix}=3$, $N=5$, $O=2$ \\

        \cmidrule(lr){2-4}

        & \multirow{1}{*}{HiCache}
        & Slow
        & $\texttt{prefix}=3$, $N=5$, $O=2$, $\texttt{scale}=0.5$ \\

        \cmidrule(lr){2-4}

        & \multirow{2}{*}{DiCache}
        & Fast
        & $\texttt{prefix\_ratio}=0.1$, $\delta=0.5$ \\
        &
        & Slow
        & $\texttt{prefix\_ratio}=0.2$, $\delta=0.5$ \\

        \cmidrule(lr){2-4}
        
        & \multirow{2}{*}{MagCache}
        & Fast
        & $\texttt{prefix\_ratio}=0.15$, $K=10$, $\delta=0.8$ \\
        &
        & Slow
        & $\texttt{prefix\_ratio}=0.2$, $K=10$, $\delta=0.8$ \\
        \bottomrule
    \end{tabular}
\end{table*}

\begin{table*}[t]
    \centering
    \caption{Baseline configurations for text-to-video generation.}
    \label{tab:app_video_baselines}
    \setlength{\tabcolsep}{6pt}
    \renewcommand{\arraystretch}{1.12}
    \begin{tabular}{lll p{6.8cm}}
        \toprule
        \textbf{Model}
        & \textbf{Method}
        & \textbf{Regime}
        & \textbf{Configuration} \\
        \midrule

        \multirow{5}{*}{Wan~2.1}

        & \multirow{1}{*}{MagCache}
        & Slow
        & $\texttt{prefix\_ratio}=0.2$,
          $K=4$,
          $\delta=0.12$ \\

        \cmidrule(lr){2-4}
        
        & \multirow{2}{*}{DiCache}
        & Fast
        & $\texttt{prefix\_ratio}=0.2$,
          $\delta=0.25$ \\
        &
        & Slow
        & $\texttt{prefix\_ratio}=0.2$,
          $\delta=0.15$ \\

        \cmidrule(lr){2-4}

        & \multirow{2}{*}{SenCache}
        & Fast
        & $\texttt{prefix\_ratio}=0.16$, $\delta=0.55$ \\
        &
        & Slow
        & $\texttt{prefix\_ratio}=0.2$, $\delta=0.4$ \\

        \midrule

        \multirow{4}{*}{CogVideoX}
        & \multirow{2}{*}{TeaCache}
        & Fast
        & $\delta=0.3$ \\
        &
        & Slow
        & $\delta=0.22$ \\

        \cmidrule(lr){2-4}

        & \multirow{2}{*}{SenCache}
        & Fast
        & $\texttt{prefix\_ratio}=0.15$, $\delta=2.5$ \\
        &
        & Slow
        & $\texttt{prefix\_ratio}=0.2$, $\delta=1.4$ \\

        \midrule

        \multirow{4}{*}{OpenSora~1.2}
        & \multirow{2}{*}{TeaCache}
        & Fast
        & $\delta=0.28$ \\
        &
        & Slow
        & $\delta=0.2$ \\

        \cmidrule(lr){2-4}

        & \multirow{2}{*}{MagCache}
        & Fast
        & $\texttt{prefix\_ratio}=0.18$,
          $K=4$,
          $\delta=0.12$ \\
        &
        & Slow
        & $\texttt{prefix\_ratio}=0.2$,
          $K=3$,
          $\delta=0.12$ \\
        \bottomrule
    \end{tabular}
\end{table*}

\subsection{Evaluation and Implementation Details}
\label{app:evaluation_protocol}

\paragraph{Hardware and software.}
All experiments are conducted on a single NVIDIA H200 GPU in a Linux
environment with Python~3.12.13, CUDA Toolkit~13.0.2, PyTorch~2.11.0,
Diffusers~0.37.1, and Transformers~5.4.0. The complete environment
configuration and dependency list will be released with the code repository.

\paragraph{Random seeds.}
We evaluate DrawBench and VBench using three and five different base seeds,
respectively, with one sample generated per prompt in each run. Specifically, the image experiments use base seeds $\{0,2026,2027\}$, while the video experiments use base seeds $\{100,200,300,400,500\}$. For a prompt with global index $p$, its seed is deterministically assigned as
\begin{equation}
    s_p = s_{\mathrm{base}} + p.
\end{equation}
Full inference and all accelerated methods use identical prompt--seed pairs, independent of batch size or data-parallel partitioning. Results are averaged over the corresponding base-seed runs.

\paragraph{Latency measurement.}
Inference latency is measured during the actual generation of the DrawBench
and VBench evaluation samples rather than using isolated synthetic inputs.
For each prompt--seed pair, wall-clock time is recorded using
\texttt{time.perf\_counter()} with CUDA synchronization before each timestamp
to ensure that asynchronous GPU operations are completed. The reported
latency is the arithmetic mean over all evaluated prompts and random seeds for
each model and method under the same hardware and software environment.
Output serialization and metric computation are excluded from the timing.

The speedup ratio is computed as
\begin{equation}
    \mathrm{Speedup}
    =
    \frac{
        \mathrm{Latency}_{\mathrm{full}}
    }{
        \mathrm{Latency}_{\mathrm{method}}
    }.
    \label{app:eq:speedup}
\end{equation}

\paragraph{FLOPs and memory.}
FLOPs and GPU memory are profiled over the generation loop. FLOPs are
estimated using \texttt{torch.profiler} with FLOP counting enabled, and the
estimates reported by all supported ATen operators are summed. The profiler
primarily provides FLOP metadata for matrix multiplication and convolution;
therefore, the reported FLOPs do not necessarily include every element-wise
or framework-level operation. Before each profiled generation, we synchronize
the GPU and reset the CUDA peak-memory statistics. Peak memory is reported as
the maximum allocated CUDA memory obtained from
\texttt{torch.cuda.max\_memory\_allocated()}, expressed in GiB. We report
allocated rather than reserved memory to reflect the peak memory actively used
by tensors during inference.

\paragraph{Offline profiling cost.}
Offline trajectory collection and anchor selection are performed once for
each model. Their costs are excluded from per-sample inference latency because
the resulting anchor schedule is shared across all evaluation samples.


\subsection{Evaluation Metrics}
\label{app:metrics}

We evaluate both inference efficiency and generation quality. Efficiency is
measured by latency, speedup, FLOPs, and peak GPU memory. Generation quality
is evaluated using semantic-quality metrics and paired reconstruction metrics.

\paragraph{ImageReward.}
ImageReward~\citep{xu2023imagereward} is a learned human-preference metric
for text-to-image generation. It evaluates the consistency and perceptual
quality of a generated image with respect to its text prompt. A higher score
indicates stronger alignment with human preference.

\paragraph{CLIP Score.}
CLIP Score~\citep{hessel2021clipscore} measures text--image alignment using the cosine similarity between pretrained CLIP text and image embeddings. We use the \texttt{CLIP-ViT-Base-Patch32} model for feature extraction. Higher values indicate stronger semantic consistency with the input prompt.

\paragraph{VBench.}
VBench~\citep{huang2024vbench} evaluates video generation along multiple spatial, temporal, semantic, and perceptual dimensions. We use the official evaluation implementation and report its aggregated overall score. Higher scores indicate better video quality.

\paragraph{LPIPS.}
Learned Perceptual Image Patch Similarity
(LPIPS)~\citep{zhang2018unreasonable} measures perceptual discrepancy using
deep visual features. Let $\hat{x}$ and $x^{\star}$ denote the accelerated and
full-inference outputs, respectively. LPIPS is computed as
\begin{equation}
    \operatorname{LPIPS}(\hat{x},x^{\star})
    =
    \sum_{l}
    \frac{1}{H_lW_l}
    \sum_{h,w}
    \left\|
        \mathbf{w}_l
        \odot
        \left(
            \widetilde{\phi}_l(\hat{x})_{h,w}
            -
            \widetilde{\phi}_l(x^{\star})_{h,w}
        \right)
    \right\|_2^2,
\end{equation}
where $\widetilde{\phi}_l(\cdot)$ denotes the channel-normalized feature at
layer $l$, and $\mathbf{w}_l$ contains the learned channel-wise weights. We
use AlexNet as the feature-extraction backbone. Lower values indicate that
the accelerated output is perceptually closer to the corresponding
full-inference output.

\paragraph{SSIM.}
Structural Similarity Index Measure
(SSIM)~\citep{wang2004image} compares the luminance, contrast, and structural
information of the accelerated and full-inference outputs. For two local image
windows $\hat{x}$ and $x^{\star}$, it is defined as
\begin{equation}
    \operatorname{SSIM}(\hat{x},x^{\star})
    =
    \frac{
        (2\mu_{\hat{x}}\mu_{x^{\star}}+C_1)
        (2\sigma_{\hat{x}x^{\star}}+C_2)
    }{
        (\mu_{\hat{x}}^2+\mu_{x^{\star}}^2+C_1)
        (\sigma_{\hat{x}}^2+\sigma_{x^{\star}}^2+C_2)
    },
\end{equation}
where $\mu_{\hat{x}}$ and $\mu_{x^{\star}}$ are the local means,
$\sigma_{\hat{x}}^2$ and $\sigma_{x^{\star}}^2$ are the local variances,
$\sigma_{\hat{x}x^{\star}}$ is their covariance, and $C_1$ and $C_2$ are
stabilizing constants. The final score is averaged over all spatial windows.
Higher values indicate stronger structural similarity.

\paragraph{PSNR.}
Peak Signal-to-Noise Ratio measures pixel-level reconstruction fidelity:
\begin{equation}
    \mathrm{PSNR}
    =
    10\log_{10}
    \left(
        \frac{R^2}{\mathrm{MSE}}
    \right),
    \label{app:eq:psnr}
\end{equation}
where $R$ is the maximum representable pixel value. Higher PSNR indicates
smaller pixel-level reconstruction error.

For video reconstruction, LPIPS, SSIM, and PSNR are computed between
temporally corresponding frames and then averaged over frames and evaluation
prompts.

\section{Additional Experimental Results and Analysis}
\label{app:additional_results}

\subsection{Additional Trajectory Analysis}
\label{app:additional_trajectory_analysis}

\begin{figure*}[t]
    \centering

    \begin{subfigure}{1.0\textwidth}
        \centering
        \includegraphics[width=\textwidth]{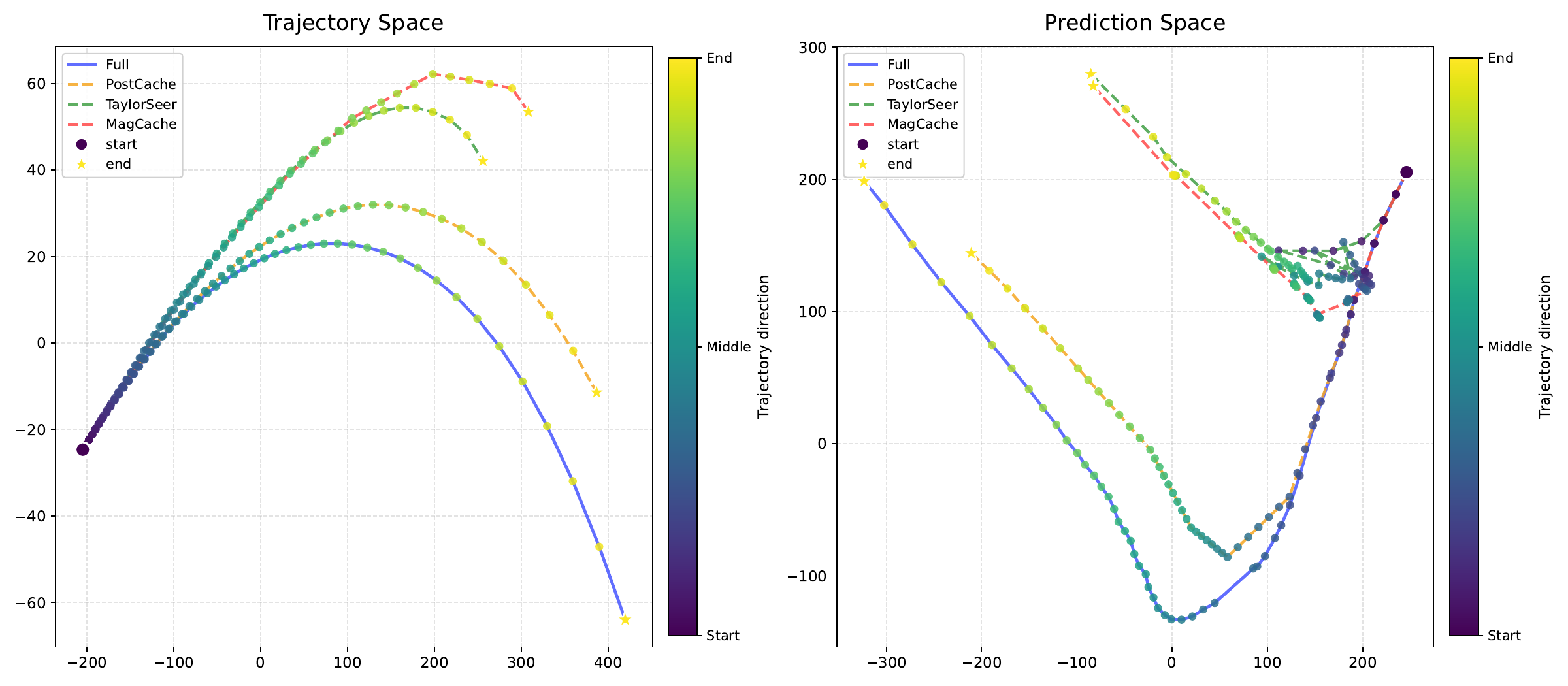}
        \caption{Trajectory visualization on FLUX.1-dev.}
        \label{fig:traj_flux}
    \end{subfigure}

    \vspace{0.6em}

    \begin{subfigure}{1.0\textwidth}
        \centering
        \includegraphics[width=\textwidth]{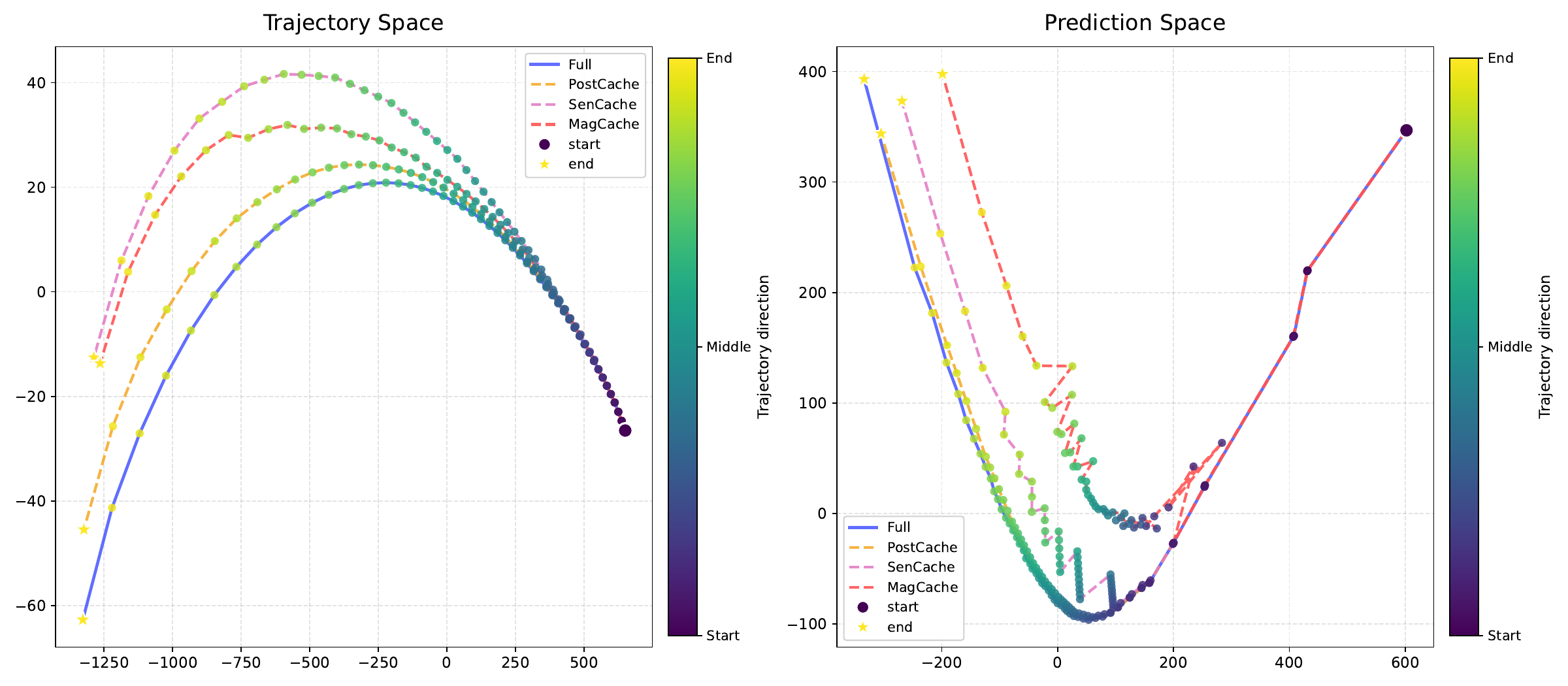}
        \caption{Trajectory visualization on Wan 2.1.}
        \label{fig:traj_wan}
    \end{subfigure}

    \caption{
    Additional trajectory visualizations. For each model, the left panel shows the denoising trajectories and the right panel shows the model predictions in the projected space.
    Compared with TaylorSeer, MagCache and SenCache, PoSTCache stays closer to the full-inference trajectory, illustrating the corrective effect of postdictive smoothing.
    }
    \label{fig:trajectory_visualization_appendix}
\end{figure*}

\paragraph{Additional trajectory visualization.} 
To complement the quantitative trajectory analysis in Sec.~\ref{sec:trajectory_analysis}, we visualize the denoising trajectories of full inference, forward-only caching, and PoSTCache in a low-dimensional space. As shown in Figure~\ref{fig:trajectory_visualization_appendix}, forward-only caching progressively deviates from the full-inference trajectory, whereas postdictive smoothing consistently pulls the trajectory toward the full-inference reference after future anchors become available. This behavior is observed across both image and video generation models.

\paragraph{Evaluation details for Figure~\ref{fig:trajectory_analysis}(c).} 
For a candidate future anchor $s$ following the most recent full-computation
anchor $a$, we quantify its local cache risk under nearest reuse as
\begin{equation}
R_{\mathrm{local}}(s\mid a)
=
\frac{1}{ND}
\sum_{n=1}^{N}
\sum_{k=a+1}^{s}
\gamma_k
\left\|
\bar{u}_k^{(n)}-\bar{u}_a^{(n)}
\right\|_2^2
\end{equation}
To measure its trajectory-level corrective value, we use the same
relative-position gain as PoSTCache,
$w_{k\mid a,s}=(\eta_k-\eta_a)/(\eta_s-\eta_a)$, and define
\begin{equation}
G_{\mathrm{fixed}}(s\mid a)
=
\frac{1}{ND}
\sum_{n=1}^{N}
\sum_{k=a+1}^{s-1}
\gamma_k
\left[
\|e_k^{(n,a)}\|_2^2
-
\|e_k^{(n,a)}-w_{k\mid a,s}r_s^{(n,a)}\|_2^2
\right]
\end{equation}
where
$e_k^{(n,a)}=\bar{u}_k^{(n)}-\bar{u}_a^{(n)}$
and
$r_s^{(n,a)}=\bar{u}_s^{(n)}-\bar{u}_a^{(n)}$.

For each candidate set $\mathcal{C}$, top-1 agreement compares the maximizers
of $R_{\mathrm{local}}$ and $G_{\mathrm{fixed}}$, while normalized regret is
$1-G_{\mathrm{fixed}}(s_{\mathrm{local}})/
\max_{s\in\mathcal{C}}G_{\mathrm{fixed}}(s)$.



\subsection{Calibration Size and Selector Robustness}
\label{app:calibration_analysis}

PoSTCache estimates the cross-timestep covariance kernel from a small set of
offline denoising trajectories. Since the kernel is computed from deviations
around the empirical mean, a single trajectory yields a degenerate zero
covariance and cannot provide a meaningful uncertainty estimate. We therefore
vary the calibration size from $N=2$ to $N=20$ and examine its effect on
anchor selection and reconstruction fidelity.

\begin{table}[t]
    \centering
    \caption{Effect of the number of offline calibration trajectories on
    Qwen-Image-2512 with $\texttt{budget\_steps}=16$ and
    $\texttt{dense\_prefix}=7$. The default setting is highlighted in bold.}
    \label{tab:app_calibration_size}
    \setlength{\tabcolsep}{5pt}
    \begin{tabular}{cccc}
        \toprule
        $N$ & LPIPS $\downarrow$ & SSIM $\uparrow$
        & PSNR $\uparrow$ \\
        \midrule
        2  & 0.0641 & 0.9267 & 27.266 \\
        5  & 0.0607 & 0.9302 & 27.597 \\
        \textbf{10} & \textbf{0.0607} & \textbf{0.9302}
        & \textbf{27.597} \\
        20 & 0.0613 & 0.9285 & 27.223 \\
        \bottomrule
    \end{tabular}
\end{table}

As shown in Table~\ref{tab:app_calibration_size}, using only two trajectories
leads to a less reliable covariance estimate and slightly lower reconstruction
quality. Performance stabilizes from $N=5$: using $N=5$ and $N=10$ produces
identical schedules and reconstruction results, while increasing the calibration set to $N=20$ provides no
further improvement. These results indicate that the selected anchors are
robust to the calibration size once a small number of trajectories is
available. We use $N=10$ throughout the main experiments as a conservative
default that balances estimation reliability and offline profiling cost.

\subsection{Schedule-Level Validation of the PVR Objective}
\label{app:pvr_surrogate_validation}

The derivation in Sec.~\ref{app:posterior_anchor_derivation} motivates PVR as
a schedule-level surrogate: for an anchor set $A$, the weighted posterior
uncertainty
\begin{equation}
    J_{\mathrm{PVR}}(A)
    :=
    \sum_{t=1}^{T}\gamma_t V_t(A)
    \label{app:eq:pvr_validation_objective}
\end{equation}
should be small when the selected anchors provide informative future evidence
for the remaining trajectory. We directly test this interpretation by asking
whether $J_{\mathrm{PVR}}(A)$ predicts the correction quality of complete
anchor schedules on held-out trajectories. Our default selector greedily
optimizes the global PVR objective by selecting the candidate with the largest marginal variance reduction over the full trajectory; we denote it as
\textbf{PVR (ours)} below.

\paragraph{Protocol.}
We evaluate Qwen-Image-2512 and Wan~2.1 using the Slow configurations in
Tables~\ref{tab:app_model_configurations}
and~\ref{tab:app_postcache_configurations}. The proposal, scheduler-aware
temporal weights, and postdictive correction follow
Sec.~\ref{app:postcache_configuration}. For each model, the covariance kernel
is constructed from the same $10$ calibration trajectories used by the
selector and evaluated on $16$ held-out prompt--seed pairs. Their prompts,
seeds, and exact prompt--seed combinations do not overlap with calibration.

Under a fixed computation budget, we evaluate $135$ unique schedules:
$100$ randomly generated schedules, $30$ local perturbations of
PVR (ours), and five named selectors---PVR (ours), preceding-interval PVR,
local proposal risk, fixed interval, and distance only. Random schedules keep
the dense prefix and final anchor fixed and sample the remaining anchor
indices without replacement. Each local perturbation moves one
non-mandatory PVR anchor by one to three timestep indices; collisions and
out-of-range moves are rejected. The primary schedule pool uses seed
\texttt{20260727}.

For schedule $A$, let $\widetilde{u}_k^{(n,A)}$ and
$\widehat{u}_k^{(n,A)}$ denote the nearest-reuse proposal and its corrected
prediction for held-out trajectory $n$. Using the full-computation prediction
$\bar{u}_k^{(n)}$ from Eq.~(\ref{app:eq:full_trajectory}), we measure
\begin{align}
    E_{\mathrm{pre}}(A)
    &=
    \frac{1}{N_{\mathrm{val}}D}
    \sum_{n=1}^{N_{\mathrm{val}}}
    \sum_{k\in\mathcal{K}_{\mathrm{eval}}}
    \gamma_k
    \left\|
        \bar{u}_k^{(n)}
        -
        \widetilde{u}_k^{(n,A)}
    \right\|_2^2,
    \\
    E_{\mathrm{post}}(A)
    &=
    \frac{1}{N_{\mathrm{val}}D}
    \sum_{n=1}^{N_{\mathrm{val}}}
    \sum_{k\in\mathcal{K}_{\mathrm{eval}}}
    \gamma_k
    \left\|
        \bar{u}_k^{(n)}
        -
        \widehat{u}_k^{(n,A)}
    \right\|_2^2,
    \label{app:eq:pvr_validation_errors}
\end{align}
where $\mathcal{K}_{\mathrm{eval}}$ contains all solver steps except the
initial mandatory anchor, and $D$ is the product of all non-batch and
non-temporal prediction dimensions. Right anchors are included in
$E_{\mathrm{pre}}$ and have zero residual in $E_{\mathrm{post}}$. We further
define
\begin{equation}
    Q_{\mathrm{PVR}}(A)
    =
    J_{\mathrm{PVR}}(A_0)-J_{\mathrm{PVR}}(A),
    \qquad
    R_{\mathrm{corr}}(A)
    =
    \frac{
        E_{\mathrm{pre}}(A)-E_{\mathrm{post}}(A)
    }{
        E_{\mathrm{pre}}(A)
    },
    \label{app:eq:pvr_validation_correction}
\end{equation}
where $A_0$ contains the mandatory anchors.

\begin{figure*}[t]
    \centering
    \includegraphics[
        width=\textwidth
    ]{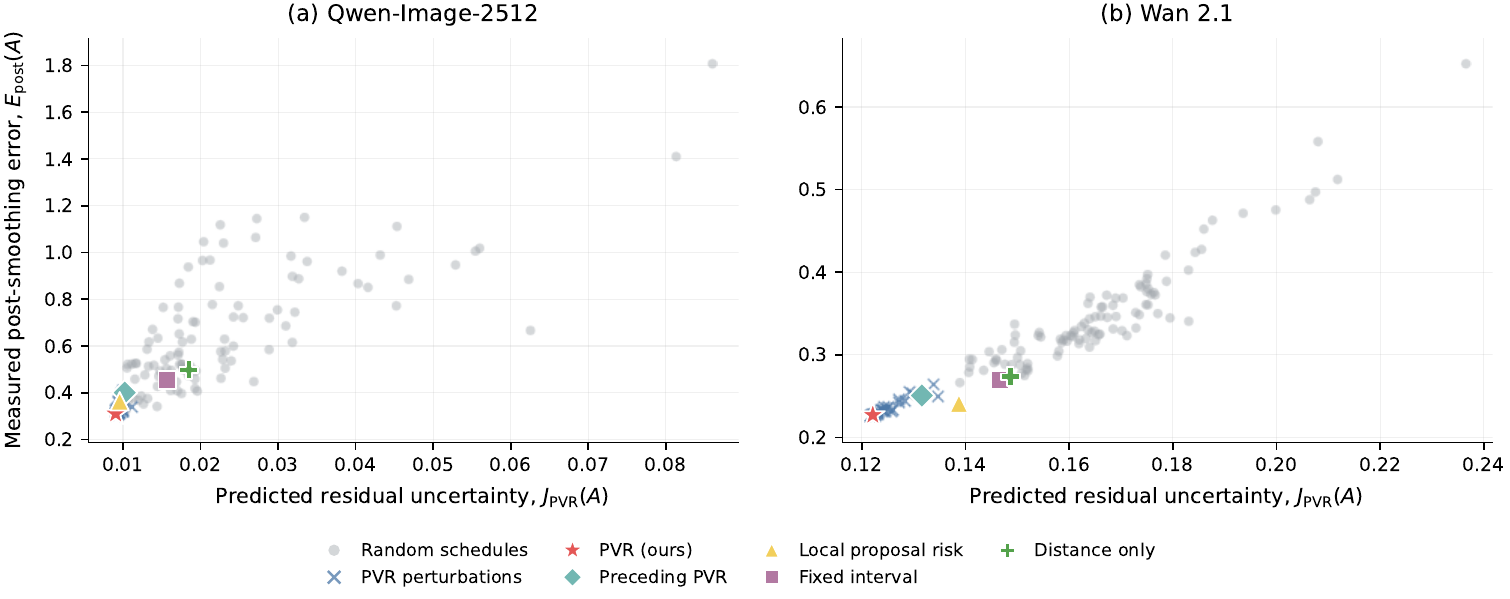}
    \caption{
        Predicted PVR uncertainty versus measured post-smoothing error on
        Qwen-Image-2512 and Wan~2.1. Each point denotes a complete schedule
        evaluated under the same computation budget. Random schedules, local
        perturbations, and named selectors are distinguished by the legend;
        lower-left is better. PVR (ours) denotes the default greedy selector
        that optimizes the global PVR objective.
    }
    \label{fig:app_pvr_schedule_scatter}
\end{figure*}

\begin{table*}[t]
    \centering
    \caption{
        Schedule-level validation of PVR.
        Brackets report $95\%$ paired bootstrap intervals over the $16$
        held-out trajectories. The schedule pool is fixed during bootstrap
        resampling. The PVR percentile is based on $E_{\mathrm{post}}$ over
        the $135$ sampled schedules.
    }
    \label{tab:app_pvr_surrogate_validation}
    \setlength{\tabcolsep}{4.5pt}
    \resizebox{\textwidth}{!}{
    \begin{tabular}{lccccc}
        \toprule
        \textbf{Model}
        & $\boldsymbol{\rho(J_{\mathrm{PVR}},E_{\mathrm{post}})}$
        & $\boldsymbol{\rho(Q_{\mathrm{PVR}},R_{\mathrm{corr}})}$
        & \textbf{Pairwise accuracy}
        & \textbf{PVR percentile}
        & $\boldsymbol{\rho_{\mathrm{local}}}$ \\
        \midrule
        Qwen-Image-2512
        & $0.858\,[0.836,\,0.885]$
        & $0.897\,[0.877,\,0.918]$
        & $83.7\%\,[82.2\%,\,85.4\%]$
        & $95.6\%\,[94.1\%,\,95.6\%]$
        & $0.644\,[0.606,\,0.693]$ \\
        Wan~2.1
        & $0.963\,[0.931,\,0.971]$
        & $0.967\,[0.946,\,0.969]$
        & $92.3\%\,[89.6\%,\,93.3\%]$
        & $97.8\%\,[97.8\%,\,97.8\%]$
        & $0.933\,[0.914,\,0.954]$ \\
        \bottomrule
    \end{tabular}}
\end{table*}

\paragraph{Schedule-level correspondence.}
Figure~\ref{fig:app_pvr_schedule_scatter} shows a clear monotonic relationship
between predicted residual uncertainty and measured post-smoothing error:
schedules with smaller $J_{\mathrm{PVR}}(A)$ are generally concentrated in
the lower-error region. The trend is particularly tight on Wan, whereas
Qwen-Image exhibits greater dispersion among random schedules, indicating
that its realized correction error contains additional variation not fully
captured by the timestep-level covariance surrogate. Nevertheless, the
overall ranking remains strong on both models.

The named selectors provide a complementary view. PVR (ours) lies in the
most favorable lower-left region of both panels and achieves the lowest
$E_{\mathrm{post}}$ among the named schedules. The local perturbations remain
near the selected solution but generally move coherently upward and to the
right, showing that the surrogate remains informative even among schedules
with similar global structure.

These visual trends agree with
Table~\ref{tab:app_pvr_surrogate_validation}. The Spearman correlation between
$J_{\mathrm{PVR}}$ and $E_{\mathrm{post}}$ reaches $0.858$ on Qwen-Image and
$0.963$ on Wan. Pairwise ranking accuracy, computed over all $9045$ unordered
schedule pairs, is $83.7\%$ and $92.3\%$, respectively; no ties occur in the
primary pools. PVR (ours) outperforms $95.6\%$ and $97.8\%$ of the sampled
schedules in terms of $E_{\mathrm{post}}$, and outperforms all $100$ randomly
generated schedules on both models.

\paragraph{Controlling for interval geometry.}
To test whether the observed relationship is explained only by spacing, we
define a schedule as gap matched when its maximum gap differs from
PVR (ours) by at most $2$ steps and the population standard deviation of its
gaps differs by at most $0.75$. This yields $66$ schedules on Qwen-Image and
$75$ on Wan. We also compute partial rank correlations by regressing the
average ranks of both variables on the average ranks of maximum gap and gap
standard deviation.

\begin{table*}[t]
    \centering
    \caption{
        Geometry-controlled schedule-level results.
        Partial correlations control for maximum gap and gap standard
        deviation. The final column counts schedule--trajectory pairs for
        which correction does not increase prediction error.
    }
    \label{tab:app_pvr_geometry_control}
    \setlength{\tabcolsep}{4pt}
    \resizebox{\textwidth}{!}{
    \begin{tabular}{lcccccc}
        \toprule
        \textbf{Model}
        & $\boldsymbol{\rho(\mathrm{maxgap},E_{\mathrm{post}})}$
        & $\boldsymbol{\rho(\mathrm{stdgap},E_{\mathrm{post}})}$
        & $\boldsymbol{\rho(J,E_{\mathrm{post}}\mid\mathrm{geometry})}$
        & $\boldsymbol{\rho(Q,R_{\mathrm{corr}}\mid\mathrm{geometry})}$
        & $\boldsymbol{E_{\mathrm{post}}\leq E_{\mathrm{pre}}}$ \\
        \midrule
        Qwen-Image-2512
        & $0.504$
        & $0.623$
        & $0.764$
        & $0.829$
        & $2160/2160$ \\
        Wan~2.1
        & $0.591$
        & $0.643$
        & $0.938$
        & $0.949$
        & $2160/2160$ \\
        \bottomrule
    \end{tabular}}
\end{table*}

The geometry-only correlations are notably weaker than the corresponding PVR
correlations. The relationship remains strong within the gap-matched pools
and after removing the rank-level effects of maximum gap and gap dispersion.
Moreover, correction reduces the prediction-space error for all
$2160=135\times16$ schedule--trajectory pairs on each model. These results
indicate that PVR captures schedule information beyond simple interval
geometry.

\paragraph{Stability and scope.}
Repeating schedule generation with seeds \texttt{20260728},
\texttt{20260729}, and \texttt{20260730} gives
$\rho(J_{\mathrm{PVR}},E_{\mathrm{post}})$ ranges of $0.860$--$0.890$ on
Qwen-Image and $0.962$--$0.981$ on Wan, with PVR (ours) percentiles of
$94.8\%$--$97.0\%$ and $97.0\%$--$98.5\%$, respectively. Thus, the main
conclusions are stable to schedule-pool construction.

The confidence intervals in Table~\ref{tab:app_pvr_surrogate_validation} are obtained from $1000$ paired bootstrap
replicates using seed \texttt{20260728}, resampling held-out trajectories
while keeping the schedule pool fixed. They therefore quantify uncertainty
across validation trajectories rather than schedule-pool sampling. Finally,
this experiment is an offline prediction-space analysis on stored
full-computation trajectories and does not model the closed-loop effect of
reintegration on subsequent predictions. End-to-end behavior is evaluated
separately by the online component ablations in Sec.~\ref{sec:ablation_studies}.

\subsection{Transferability of Offline PVR Schedules}
\label{app:pvr_transfer}

PoSTCache estimates its cross-timestep dependency profile from a small set
of offline denoising trajectories and uses the resulting PVR schedule for
subsequent inference. We examine whether a schedule profiled in one setting
can be reused after changing the prompt distribution, guidance scale, or
spatial configuration, without collecting a new target calibration set.

For each target setting, we compare two schedules.
\emph{Transferred} directly reuses the schedule selected in the source
setting, whereas \emph{Reprofiled} recomputes the dependency profile from
ten target calibration trajectories and reruns the same greedy PVR selector.
Both schedules use the same full-computation budget, nearest-reuse proposal,
and postdictive smoothing procedure. Each target setting is evaluated
against its own full-inference reference.

We report the post-smoothing prediction error $E_{\mathrm{post}}(A)$ defined in Eq.~\ref{app:eq:pvr_validation_errors}, evaluated on target trajectories disjoint
from calibration. This metric measures the discrepancy between corrected
model predictions and their full-computation counterparts before accounting
for the recursive effect of trajectory reintegration. We additionally report
LPIPS, SSIM, and PSNR against target full inference to measure the final
closed-loop reconstruction quality.

\paragraph{Prompt-distribution transfer.}
We first study FLUX.1-dev with 50 sampling steps and a full-computation
budget of $B=16$. The source schedule is profiled using the original
calibration prompts and then transferred to PartiPrompts-Complex~\citep{yu2022scaling}, whose
prompts contain more challenging compositional relationships. Target
reprofiling uses ten PartiPrompts-Complex calibration trajectories, while
prediction-space evaluation uses ten disjoint held-out trajectories.
End-to-end reconstruction is evaluated on 93 target prompts.

\begin{table}[t]
    \centering
    \caption{
        Prompt-distribution transfer on PartiPrompts-Complex.
        Bold values indicate the better result between the transferred and
        target-reprofiled schedules.
    }
    \label{tab:pvr_transfer_prompt}
    \small
    \setlength{\tabcolsep}{4.5pt}
    \begin{tabular}{lcccc}
        \toprule
        \textbf{Schedule}
        & $\mathbf{E_{\mathrm{post}}}\downarrow$
        & \textbf{LPIPS}$\downarrow$
        & \textbf{SSIM}$\uparrow$
        & \textbf{PSNR}$\uparrow$ \\
        \midrule
        Transferred
        & 0.2128
        & \textbf{0.0898}
        & \textbf{0.8979}
        & \textbf{25.8494} \\
        Reprofiled
        & \textbf{0.2019}
        & 0.0969
        & 0.8898
        & 25.1856 \\
        \bottomrule
    \end{tabular}
\end{table}

As shown in Table~\ref{tab:pvr_transfer_prompt}, target-specific
reprofiling reduces $E_{\mathrm{post}}$, indicating that its selected
anchors better match the held-out target trajectories in prediction space.
However, the transferred schedule is better on all three final
reconstruction metrics. Paired bootstrap intervals over evaluation prompts
exclude zero for LPIPS, SSIM, and PSNR, showing that the final reconstruction
advantage is not caused by a small number of prompts.

The source and target dependency profiles nevertheless remain highly
similar, with a profile-similarity score of $0.989$. This suggests that the
temporal structure exploited by PVR is largely preserved under the prompt
shift. The disagreement between $E_{\mathrm{post}}$ and the final metrics
also highlights the scope of Eq.~\ref{app:eq:pvr_validation_errors}: it measures
prediction correction on stored target trajectories but does not fully
capture how the corrected states alter subsequent denoising predictions
after reintegration.

\paragraph{Guidance-scale transfer.}
We next evaluate Qwen-Image-2512 by profiling the source schedule at
CFG $=4$ and transferring it to CFG $=2$ and CFG $=6$. All configurations
use 50 sampling steps, $B=16$, the same negative prompt, and identical
prompt--seed splits. For each target CFG, ten trajectories are used for
target reprofiling, ten disjoint trajectories are used to evaluate
$E_{\mathrm{post}}$, and 50 prompts are used for end-to-end reconstruction.

\begin{table}[t]
    \centering
    \caption{
        Guidance-scale transfer from the schedule profiled at CFG $=4$.
        Bold values indicate the better result within each target CFG.
    }
    \label{tab:pvr_transfer_cfg}
    \begin{tabular}{clcccc}
        \toprule
        \textbf{Target CFG}
        & \textbf{Schedule}
        & $\mathbf{E_{\mathrm{post}}}\downarrow$
        & \textbf{LPIPS}$\downarrow$
        & \textbf{SSIM}$\uparrow$
        & \textbf{PSNR}$\uparrow$ \\
        \midrule
        2
        & Transferred
        & 0.0855
        & 0.0319
        & 0.9694
        & \textbf{33.0123} \\
        2
        & Reprofiled
        & \textbf{0.0666}
        & \textbf{0.0310}
        & \textbf{0.9710}
        & 32.4347 \\
        \midrule
        6
        & Transferred
        & \textbf{0.6637}
        & \textbf{0.0781}
        & \textbf{0.9139}
        & \textbf{26.1554} \\
        6
        & Reprofiled
        & 0.6981
        & 0.0839
        & 0.9091
        & 25.7871 \\
        \bottomrule
    \end{tabular}
\end{table}

Table~\ref{tab:pvr_transfer_cfg} reveals different behaviors at the two
target guidance scales. At CFG $=2$, reprofiling substantially reduces
$E_{\mathrm{post}}$. The original target-reprofiled schedule also performs
well relative to schedules obtained by resampling the target calibration
set, indicating that the prediction-space improvement is reproducible
rather than an accidental schedule choice. Nevertheless, paired-bootstrap
intervals for LPIPS, SSIM, and PSNR all include zero. Thus, adapting the
profile improves the prediction-space criterion but does not produce a
statistically distinguishable improvement in final reconstruction. This is
the clearest case in which $E_{\mathrm{post}}$ is informative but incomplete
as a proxy for the closed-loop trajectory.

At CFG $=6$, the transferred schedule instead obtains lower
$E_{\mathrm{post}}$ and better mean reconstruction metrics. The LPIPS
improvement is supported by a paired-bootstrap interval excluding zero,
whereas the SSIM and PSNR differences are less conclusive. Additional
cross-evaluation shows that the target-reprofiled schedule is favored by
the profile constructed from its calibration trajectories but becomes worse
when scored using independent target trajectories.

\paragraph{Spatial-configuration transfer.}
Finally, we profile a Qwen-Image-2512 schedule at $1328\times1328$ and reuse it at $1024\times1024$ and $1664\times928$. The first target changes the resolution while preserving the square aspect ratio. The second additionally changes the aspect ratio and spatial token-grid geometry. The guidance scale, sampling steps, full-computation budget, prompts, negative prompt, and profiling procedure are otherwise kept unchanged.

\begin{table}[t]
    \centering
    \caption{
        Spatial-configuration transfer from a schedule profiled at $1328\times1328$. Bold values indicate the better result within         each target configuration.
    }
    \label{tab:pvr_transfer_spatial}
    \begin{tabular}{clcccc}
        \toprule
        \textbf{Target size}
        & \textbf{Schedule}
        & $\mathbf{E_{\mathrm{post}}}\downarrow$
        & \textbf{LPIPS}$\downarrow$
        & \textbf{SSIM}$\uparrow$
        & \textbf{PSNR}$\uparrow$ \\
        \midrule
        $1024\times1024$
        & Transferred
        & 0.3154
        & 0.0506
        & \textbf{0.9397}
        & \textbf{28.4701} \\
        $1024\times1024$
        & Reprofiled
        & \textbf{0.3003}
        & \textbf{0.0497}
        & 0.9380
        & 28.0984 \\
        \midrule
        $1664\times928$
        & Transferred
        & \textbf{0.2487}
        & \textbf{0.0535}
        & \textbf{0.9397}
        & \textbf{28.6091} \\
        $1664\times928$
        & Reprofiled
        & 0.3252
        & 0.0798
        & 0.9184
        & 27.8478 \\
        \bottomrule
    \end{tabular}
\end{table}

For the resolution-only shift to $1024\times1024$, target reprofiling
slightly improves $E_{\mathrm{post}}$ and LPIPS, whereas the transferred
schedule obtains higher SSIM and PSNR. The corresponding held-out PVR
scores are nearly identical, showing that the two schedules are practically comparable.

For the wider $1664\times928$ configuration, the transferred schedule is clearly better in both prediction and reconstruction space. Meanwhile, the source--target profile similarity remains as high as $0.996$, indicating that the underlying temporal dependency structures differ only modestly. Despite this limited profile shift, the two resulting schedules produce substantially different $E_{\mathrm{post}}$ values and final reconstruction quality. This disproportionate effect suggests a limitation of the discrete greedy selector: small perturbations in the estimated profile can change early anchor choices, which in turn alter subsequent conditional gains and lead to markedly different complete schedules. The result therefore highlights the path dependence and finite-sample instability of greedy anchor selection, rather than a large structural difference between the source and target profiles.

\paragraph{Results analysis.}
The experiments above show that target-specific reprofiling may improve,
match, or underperform direct transfer. To identify where these differences
arise, we use three complementary diagnostics.

First, \emph{profile similarity} measures whether the source and target
settings exhibit similar relative cross-timestep dependency patterns.
Let $K^{(s)}$ and $K^{(t)}$ denote the covariance kernels estimated from
the source and target calibration trajectories, respectively, using
Eq.~\ref{app:eq:streaming_covariance_kernel}. Since their marginal
prediction variances may differ across settings, we first convert each
covariance kernel into a correlation profile:
\begin{equation}
    C^{(x)}_{ij}
    =
    \frac{K^{(x)}_{ij}}
    {
        \sqrt{
            \max\!\left(K^{(x)}_{ii},\epsilon\right)
            \max\!\left(K^{(x)}_{jj},\epsilon\right)
        }
    },
    \qquad
    x\in\{s,t\},
    \label{app:eq:normalized_profile}
\end{equation}
where $\epsilon=10^{-12}$ is used for numerical stability.

Let
\begin{equation}
    \mathbf{c}^{(x)}
    =
    \operatorname{vec}_{i<j}
    \left(C^{(x)}_{ij}\right)
\end{equation}
denote the vector containing the strictly upper-triangular entries of
$C^{(x)}$. We define profile similarity as the Pearson correlation between
the source and target profile vectors:
\begin{equation}
    \operatorname{Sim}_{\mathrm{profile}}
    =
    \frac{
        \sum_k
        \left(c_k^{(s)}-\bar{c}^{(s)}\right)
        \left(c_k^{(t)}-\bar{c}^{(t)}\right)
    }{
        \sqrt{
            \sum_k\left(c_k^{(s)}-\bar{c}^{(s)}\right)^2
        }
        \sqrt{
            \sum_k\left(c_k^{(t)}-\bar{c}^{(t)}\right)^2
        }
    }.
    \label{app:eq:profile_similarity}
\end{equation}
The diagonal is excluded because it is one after correlation
normalization, and only the upper triangle is retained to avoid counting
each symmetric timestep pair twice. A value close to one indicates that
the source and target profiles exhibit similar relative cross-timestep
dependency patterns, even when their absolute covariance magnitudes or
marginal variances differ.

Second, we cross-evaluate the two fixed schedules using an independent target profile
and report which one obtains the lower schedule-level uncertainty
$J_{\mathrm{PVR}}(A)$ from
Eq.~\ref{app:eq:pvr_validation_objective}. This checks whether an advantage on
the target calibration profile persists beyond the trajectories used to
construct it.

Third, we assess sensitivity to the particular ten target calibration
trajectories. We repeatedly resample these trajectories with replacement,
recompute the target profile, and rerun greedy schedule selection. Each
bootstrap schedule is then evaluated on the same target held-out
trajectories using $E_{\mathrm{post}}$ from
Eq.~\ref{app:eq:pvr_validation_errors}. The \emph{target-schedule percentile} is the
fraction of these bootstrap schedules outperformed by the original
target-reprofiled schedule. A high percentile means that the reported target
schedule is typical or relatively strong among schedules obtained from the
same target setting, whereas a low percentile indicates an unfavorable
calibration or greedy-selection realization.

\begin{table}[t]
    \centering
    \caption{
        Diagnostics for offline schedule transfer.
        ``Lower held-out $J_{\mathrm{PVR}}$'' identifies the fixed schedule
        with lower predicted residual uncertainty on an independent target
        profile. The target percentile is the fraction of target-bootstrap
        schedules with higher $E_{\mathrm{post}}$ than the original
        reprofiled schedule.
    }
    \label{tab:pvr_transfer_diagnostics}
    \resizebox{\columnwidth}{!}{
    \begin{tabular}{lccc}
        \toprule
        \textbf{Shift}
        & \textbf{Profile similarity}
        & \textbf{Lower held-out $J_{\mathrm{PVR}}$}
        & \textbf{Target percentile} \\
        \midrule
        Prompt OOD
        & 0.989
        & Reprofiled
        & 61.6\% \\
        CFG $4\rightarrow2$
        & 0.850
        & Reprofiled
        & 88.1\% \\
        CFG $4\rightarrow6$
        & 0.943
        & Transferred
        & 4.8\% \\
        $1328\times1328\rightarrow1024\times1024$
        & 0.997
        & Comparable
        & 92.5\% \\
        $1328\times1328\rightarrow1664\times928$
        & 0.996
        & Transferred
        & 14.0\% \\
        \bottomrule
    \end{tabular}
    }
\end{table}

Table~\ref{tab:pvr_transfer_diagnostics} separates three qualitatively
different cases. For prompt OOD, the two dependency profiles are nearly
identical and the original target schedule lies near the middle of its
bootstrap distribution. The reprofiled schedule has lower held-out
$J_{\mathrm{PVR}}$ and $E_{\mathrm{post}}$, but worse final reconstruction.
The result is therefore best interpreted as strong profile overlap together
with a fixed-schedule mismatch between prediction-space ranking and
closed-loop output quality.

At CFG $=2$, the target schedule has lower held-out
$J_{\mathrm{PVR}}$, lower $E_{\mathrm{post}}$, and a high bootstrap
percentile. Target adaptation is therefore consistently beneficial for the
prediction-space objective, even though the final images remain
statistically indistinguishable. In contrast, the target schedules at
CFG $=6$ and $1664\times928$ have low bootstrap percentiles and are also
worse under the independent target profile. Their poor final results do not indicate that source profiles are intrinsically
superior. Instead, estimating a profile from a finite calibration split and
applying discrete greedy selection can produce an unfavorable target
schedule. The $1024\times1024$ case lies between these regimes: target
reprofiling slightly improves the prediction-space metrics, while the two
schedules remain effectively comparable in final reconstruction.

Overall, the source PVR schedule remains a robust practical default across
the evaluated prompt, guidance, and spatial shifts. These results do not
imply that the dependency profile is invariant or that target adaptation is
generally harmful. Rather, target reprofiling can provide genuine
prediction-space improvements, but these improvements need not translate
into better reintegrated outputs. This discrepancy reflects the current
offline--online gap: the offline PVR objective and $E_{\mathrm{post}}$ are
evaluated on stored full-computation trajectories, whereas online
reintegration changes the latent states and recursively affects all
subsequent model predictions. In addition, finite-sample profile estimation with greedy selection can introduce further variability into discrete anchor selection.

\subsection{Additional Ablation Studies}
\label{app:additional_ablations}
\paragraph{Temporal weighting.}
We ablate the power $p$ in the scheduler-aware temporal weight
\begin{equation}
    \widetilde{\gamma}_t
    =
    \left(
        \frac{\operatorname{median}_j \delta_j}
        {\delta_t+\epsilon}
    \right)^p.
\end{equation}
Setting $p=0$ yields uniform weighting, while larger values place greater
emphasis on densely sampled scheduler regions. We conduct this ablation on
Qwen-Image-2512 with $\texttt{budget\_steps}=16$ and
$\texttt{dense\_prefix}=7$, keeping the clipping range and all other
configurations fixed.

As shown in Table~\ref{tab:app_temporal_weighting}, the default setting
$p=0.5$ achieves the best overall reconstruction fidelity, reducing LPIPS
from 0.0632 to 0.0607 and improving SSIM from 0.9262 to 0.9302 and PSNR from
26.781 to 27.597 compared with uniform weighting. Larger powers provide no
further improvement, suggesting that excessively emphasizing densely sampled
regions is not beneficial. We therefore use $p=0.5$ in the main experiments.

\begin{table}[t]
    \centering
    \caption{Effect of the temporal-weighting power $p$ on Qwen-Image-2512 with $\texttt{budget\_steps}=16$ and $\texttt{dense\_prefix}=7$. The default setting is highlighted in bold.}
    \label{tab:app_temporal_weighting}
    \setlength{\tabcolsep}{6pt}
    \begin{tabular}{cccc}
        \toprule
        \textbf{Power $p$}
        & \textbf{LPIPS $\downarrow$}
        & \textbf{SSIM $\uparrow$}
        & \textbf{PSNR $\uparrow$} \\
        \midrule
        $0$
        & {0.0632}
        & {0.9262}
        & {26.781} \\

        $0.25$
        & {0.0625}
        & {0.9268}
        & {26.957} \\

        $\mathbf{0.5}$
        & \textbf{0.0607}
        & \textbf{0.9302}
        & \textbf{27.597} \\

        $0.75$
        & {0.0639}
        & {0.9271}
        & {27.585} \\

        $1.0$
        & {0.0639}
        & {0.9271}
        & {27.585} \\
        \bottomrule
    \end{tabular}
\end{table}

\paragraph{Mandatory prefix anchors.}
We study the effect of the dense-prefix length while fixing the total
full-computation budget. Specifically, we evaluate FLUX.1-dev with
$\texttt{budget\_steps}=16$ and vary the number of consecutive
full-computation steps at the beginning of the sampling trajectory. Since the
total budget is fixed, increasing the dense-prefix length allocates more
anchors to the early denoising stage while leaving fewer anchors for
PVR-based selection over the remaining trajectory.

\begin{table}[t]
    \centering
    \caption{Effect of the dense-prefix length on FLUX.1-dev with
    $\texttt{budget\_steps}=16$. In addition to the dense-prefix and PVR-selected
    anchors reported in the table, the final denoising step is always retained as
    a mandatory anchor, i.e.,
    $L_{\mathrm{prefix}} + N_{\mathrm{PVR}} + 1 = 16$.
    The default setting is highlighted in bold.}
    \label{tab:app_dense_prefix}
    \setlength{\tabcolsep}{5pt}
    \begin{tabular}{ccccc}
        \toprule
        \textbf{Dense prefix}
        & \textbf{PVR-selected anchors}
        & \textbf{LPIPS $\downarrow$}
        & \textbf{SSIM $\uparrow$}
        & \textbf{PSNR $\uparrow$} \\
        \midrule
        5
        & 10
        & 0.1147
        & 0.8756
        & 24.761 \\

        6
        & 9
        & 0.1132
        & 0.8779
        & 24.933 \\

        \textbf{7}
        & \textbf{8}
        & \textbf{0.0968}
        & \textbf{0.8944}
        & \textbf{26.284} \\

        8
        & 7
        & {0.1029}
        & {0.8889}
        & {25.846} \\

        9
        & 6
        & 0.1233
        & 0.8701
        & 24.616 \\

        \bottomrule
    \end{tabular}
\end{table}

As shown in Table~\ref{tab:app_dense_prefix}, reconstruction quality first
improves and then degrades as the dense-prefix length increases. A short
prefix leaves the rapidly varying early denoising stage insufficiently
covered, whereas an overly long prefix consumes too much of the fixed budget
and limits adaptive anchor placement over the remaining trajectory. The
default setting of $7$ achieves the best performance across all three
metrics, reducing LPIPS to $0.0968$ while increasing SSIM and PSNR to
$0.8944$ and $26.284$, respectively.

The optimal dense-prefix length depends on the model, scheduler, and total
budget, and does not admit a simple closed-form prescription. We therefore
adopt the empirical rule
\begin{equation}
    \label{app:eq:dense_prefix_rule}
    L_{\mathrm{prefix}}
    \in
    \left\{
        \left\lfloor\frac{\mathcal{B}}{2}\right\rfloor-1,\,
        \left\lfloor\frac{\mathcal{B}}{2}\right\rfloor
    \right\},
\end{equation}
which is used for all model-specific configurations reported in
Table~\ref{tab:app_postcache_configurations}.

\paragraph{Comparison with DPCache.}
\label{app:comparison_with_dpcache}

PVR globally models cross-timestep dependencies when evaluating the
information value of an anchor. DPCache~\citep{cui2026dpcache} is therefore
the closest conceptual comparison, as it also formulates timestep selection
as a global optimization problem, but minimizes a path-additive cost through
dynamic programming.

To isolate the scheduling objective, we adapt DPCache to the same PoSTCache
online pipeline. \textbf{DPCache-Pre} follows the original pre-correction
principle and constructs its path cost from nearest-reuse proposal errors
before postdictive smoothing. \textbf{DPCache-Retro} is our correction-aware
extension, which retains the same dynamic-programming formulation but
computes the interval cost after applying postdictive smoothing. All
selectors use FLUX.1-dev with 50 sampling steps, a budget of $B=16$, ten
calibration trajectories, and the same nearest-reuse and single-anchor
smoothing pipeline. Prediction-space evaluation uses ten disjoint held-out
trajectories, while reconstruction metrics are evaluated over 200 prompts.

\begin{table}[t]
    \centering
    \caption{
        Comparison with path-based scheduling on FLUX.1-dev.
        DPCache-Retro is our correction-aware extension of DPCache.
        All selectors use the same online pipeline and computation budget.
    }
    \label{tab:app_path_selector_comparison}
    \setlength{\tabcolsep}{5pt}
    \begin{tabular}{lcccc}
        \toprule
        \textbf{Selector}
        & $\mathbf{E_{\mathrm{post}}}\downarrow$
        & \textbf{LPIPS $\downarrow$}
        & \textbf{SSIM $\uparrow$}
        & \textbf{PSNR $\uparrow$} \\
        \midrule
        DPCache-Pre
        & 0.2562
        & 0.1076
        & 0.8827
        & 25.16 \\
        DPCache-Retro
        & \textbf{0.1991}
        & 0.1059
        & 0.8853
        & 25.55 \\
        PVR
        & 0.2024
        & \textbf{0.0968}
        & \textbf{0.8944}
        & \textbf{26.28} \\
        \bottomrule
    \end{tabular}
\end{table}

As shown in Table~\ref{tab:app_path_selector_comparison},
DPCache-Retro consistently improves over DPCache-Pre, confirming that
anchor scheduling should account for the subsequent postdictive correction.
DPCache-Retro obtains a slightly lower held-out $E_{\mathrm{post}}$ than
PVR, which is expected because its path cost is directly constructed from
the post-smoothing prediction error. Nevertheless, PVR achieves the best
final reconstruction quality across all three metrics.

This discrepancy further reflects the offline--online gap: reducing
prediction error on stored full-computation trajectories does not fully
capture how reintegration changes the latent trajectory and recursively
affects later model predictions. It also highlights the distinction between
the two global objectives: DPCache minimizes fixed path-additive interval
costs, whereas PVR evaluates the uncertainty reduction of each anchor
conditioned on the complete selected set, thereby modeling redundancy and
complementarity among anchors.

\subsection{Anchor-State Consistency and Compute Allocation}
\label{app:anchor_consistency}

\paragraph{Local anchor-state consistency.}
At a future anchor $b$, the full prediction used to construct the anchor
innovation is evaluated at the state $\widetilde{x}_b$ reached by the
forward proposal trajectory.
Postdictive smoothing subsequently revises the preceding skipped predictions
and reintegrates the interval from anchor $a$, yielding a corrected anchor
state $\widehat{x}_b$.
Because the default implementation retains the already-computed anchor
prediction, it is not re-evaluated at $\widehat{x}_b$.
To quantify the resulting local state--prediction mismatch, we compare
\begin{equation}
    \bar{u}_b
    =
    f_\theta(\widetilde{x}_b,\tau_b),
    \qquad
    u_b^{\mathrm{fresh}}
    =
    f_\theta(\widehat{x}_b,\tau_b),
    \label{app:eq:anchor_consistency_predictions}
\end{equation}
where $u_b^{\mathrm{fresh}}$ is used only as a diagnostic probe and does not
alter the default sampling trajectory.

We measure the relative state displacement and prediction discrepancy as
\begin{equation}
    D_x(b)
    =
    \frac{
        \|\widehat{x}_b-\widetilde{x}_b\|_2
    }{
        \|\widetilde{x}_b\|_2
    },
    \qquad
    E_{\mathrm{anchor}}(b)
    =
    \frac{
        \|u_b^{\mathrm{fresh}}-\bar{u}_b\|_2
    }{
        \|u_b^{\mathrm{fresh}}\|_2
    }.
    \label{app:eq:anchor_consistency_metrics}
\end{equation}
We additionally report the directional discrepancy
\begin{equation}
    C_{\mathrm{anchor}}(b)
    =
    1-
    \cos
    \left(
        u_b^{\mathrm{fresh}},
        \bar{u}_b
    \right).
    \label{app:eq:anchor_cosine_discrepancy}
\end{equation}

To assess whether this prediction discrepancy materially affects the
subsequent numerical update, we further define
\begin{equation}
    E_{\mathrm{step}}(b)
    =
    \frac{
        \left\|
        \mathcal{S}_b
        \left(
            \widehat{x}_b,
            \bar{u}_b,
            \tau_b
        \right)
        -
        \mathcal{S}_b
        \left(
            \widehat{x}_b,
            u_b^{\mathrm{fresh}},
            \tau_b
        \right)
        \right\|_2
    }{
        \left\|
        \mathcal{S}_b
        \left(
            \widehat{x}_b,
            u_b^{\mathrm{fresh}},
            \tau_b
        \right)
        \right\|_2
    }.
    \label{app:eq:anchor_solver_discrepancy}
\end{equation}

We conduct this diagnostic analysis on Qwen-Image-2512 and Wan~2.1 using
their Fast and Slow configurations in
Table~\ref{tab:app_postcache_configurations}.
We use eight evaluation prompts with a fixed random seed and identical
prompt--seed pairs across all compared variants.
All consistency statistics are averaged over the cache intervals encountered
during inference.

\begin{table}[t]
    \centering
    \caption{
        Anchor-state consistency under the paper configurations.
        $D_x$ measures reintegration-induced state displacement,
        $E_{\mathrm{anchor}}$ the resulting prediction discrepancy,
        $C_{\mathrm{anchor}}$ the corresponding directional discrepancy,
        and $E_{\mathrm{step}}$ the normalized downstream state discrepancy
        after one solver update.
    }
    \label{tab:app_anchor_consistency}
    \setlength{\tabcolsep}{4pt}
    \begin{tabular}{llcccc}
        \toprule
        \textbf{Model}
        & \textbf{Regime}
        & $D_x$
        & $E_{\mathrm{anchor}}$
        & $C_{\mathrm{anchor}}$
        & $E_{\mathrm{step}}$ \\
        \midrule

        Qwen-Image-2512
        & Slow
        & 0.0216
        & 0.0531
        & 0.0033
        & 0.0035 \\

        Qwen-Image-2512
        & Fast
        & 0.0222
        & 0.0543
        & 0.0038
        & 0.0043 \\

        \midrule

        Wan~2.1
        & Slow
        & 0.0090
        & 0.0540
        & 0.0024
        & 0.0048 \\

        Wan~2.1
        & Fast
        & 0.0148
        & 0.0817
        & 0.0063
        & 0.0082 \\

        \bottomrule
    \end{tabular}
\end{table}

As shown in Table~\ref{tab:app_anchor_consistency}, reintegration changes the
future-anchor state only modestly, with $D_x<2.3\%$ across all evaluated
settings.
The resulting prediction discrepancy is measurable, whereas the normalized
downstream state discrepancy after one solver update is approximately one
order of magnitude smaller.
Qwen-Image exhibits nearly unchanged consistency statistics between its Fast
and Slow configurations, while Wan shows a larger discrepancy under the
sparser Fast schedule.
Furthermore, $D_x$ and $E_{\mathrm{anchor}}$ are strongly correlated
(Spearman $\rho=0.845$--$0.925$), as are
$E_{\mathrm{anchor}}$ and $E_{\mathrm{step}}$
($\rho=0.892$--$0.917$).
These results support the expected propagation from reintegration-induced
state displacement to prediction staleness and subsequently to the numerical
solver update, while showing that the immediate downstream perturbation
remains comparatively small.

\paragraph{Fresh-anchor oracle and compute allocation.}
A small discrepancy after a single numerical update does not necessarily
imply negligible end-to-end impact, since diffusion sampling recursively
feeds each updated latent state into subsequent model evaluations.
We therefore construct a \emph{Fresh-Anchor Oracle} that recomputes the model
prediction at the reintegrated anchor state and uses this fresh prediction
for subsequent sampling.
Importantly, the preceding cache interval is not smoothed or reintegrated a
second time, so this variant isolates the effect of restoring anchor-state
consistency rather than introducing an additional correction iteration.

\begin{table}[t]
    \centering
    \caption{
        Anchor refresh versus broader temporal anchor coverage.
        NFE denotes the actual number of neural-network evaluations recorded
        at runtime.
        The two $21$-NFE Qwen-Image variants provide a compute-matched
        comparison between refreshing existing anchors and allocating the same
        evaluations to additional standard PoSTCache anchors.
    }
    \label{tab:app_anchor_compute_allocation}
    \setlength{\tabcolsep}{4.2pt}
    \begin{tabular}{llcccc}
        \toprule
        \textbf{Model}
        & \textbf{Variant}
        & \textbf{NFE}
        & \textbf{LPIPS $\downarrow$}
        & \textbf{SSIM $\uparrow$}
        & \textbf{PSNR $\uparrow$} \\
        \midrule

        \multirow{4}{*}{Qwen}
        & Fast ($\mathcal{B}=14$)
        & 14
        & 0.0749
        & 0.9466
        & 28.90 \\

        & Slow ($\mathcal{B}=16$)
        & 16
        & 0.0646
        & 0.9547
        & 29.61 \\

        & Fast + Fresh
        & 21
        & 0.0553
        & 0.9622
        & 30.97 \\

        & $\mathcal{B}=21$
        & 21
        & \textbf{0.0239}
        & \textbf{0.9801}
        & \textbf{34.04} \\

        \midrule

        \multirow{3}{*}{Wan}
        & Fast ($\mathcal{B}=16$)
        & 16
        & 0.0914
        & 0.8647
        & 26.60 \\

        & Slow ($\mathcal{B}=20$)
        & 20
        & \textbf{0.0717}
        & \textbf{0.8843}
        & \textbf{28.17} \\

        & Fast + Fresh
        & 24
        & 0.0767
        & 0.8779
        & 27.75 \\

        \bottomrule
    \end{tabular}
\end{table}

Fresh-anchor recomputation consistently improves reconstruction relative to
the corresponding Fast configuration, reducing LPIPS from $0.0749$ to
$0.0553$ on Qwen-Image and from $0.0914$ to $0.0767$ on Wan.
This confirms that the local state--prediction mismatch, although small at
each individual solver update, can accumulate into a measurable end-to-end
reconstruction gap.

Restoring exact consistency, however, requires additional neural-network
evaluations.
To determine whether anchor refresh is an effective use of this extra
computation, we perform an NFE-matched comparison on Qwen-Image.
Its Fast configuration uses $\mathcal{B}=14$ and a dense-prefix length
$L_{\mathrm{prefix}}=7$.
The Fresh-Anchor Oracle retains these seven dense-prefix evaluations and
refreshes all seven post-prefix anchors, including the mandatory final anchor,
resulting in $14+7=21$ total NFEs.

For the compute-matched baseline, we instead set
$\mathcal{B}=21$ and $L_{\mathrm{prefix}}=10$ and rerun the same PVR anchor
selection procedure.
Thus, both variants use exactly $21$ model evaluations, but spend the
additional computation differently:
the Fresh-Anchor Oracle re-evaluates already-selected anchors, whereas the
standard $\mathcal{B}=21$ configuration allocates the evaluations to broader
temporal anchor coverage.

At the identical budget of $21$ NFEs, broader anchor coverage reduces LPIPS
from $0.0553$ to $0.0239$, while improving SSIM from $0.9622$ to $0.9801$
and PSNR from $30.97$ to $34.04$.
Hence, although explicit anchor refresh reveals genuine consistency
headroom, the same additional model evaluations are substantially more
effective when allocated to informative new anchors in this controlled
Qwen-Image comparison.

The Wan operating points provide complementary evidence.
Fresh-anchor recomputation improves the Fast configuration from LPIPS
$0.0914$ to $0.0767$, but requires $24$ NFEs and still remains slightly
behind the $20$-NFE Slow configuration with LPIPS $0.0717$.
Together, these results show that unconditional anchor refresh is not an
obviously preferable use of additional computation.
In the controlled Qwen-Image comparison, allocating the same NFE budget to
broader anchor coverage is substantially more effective, while Wan exhibits
the same qualitative tendency despite using different operating points.
This supports our default design of prioritizing informative temporal anchors
without introducing mandatory refresh evaluations.

\subsection{Efficiency and Runtime Analysis}
\label{app:efficiency_analysis}
\paragraph{Offline calibration cost.}
PoSTCache uses complete denoising trajectories to estimate the cross-timestep covariance kernel, but the raw high-dimensional predictions do not need to be persistently stored. For each calibration trajectory $Z^{(n)}\in\mathbb{R}^{T\times D}$, we accumulate the sufficient statistics
\begin{equation}
    S=\sum_{n=1}^{N} Z^{(n)},
    \qquad
    Q=\sum_{n=1}^{N} Z^{(n)}Z^{(n)\top},
\end{equation}
from which the kernel used by the PVR selector is recovered exactly as
\begin{equation}
    K
    =
    \frac{Q}{ND}
    -
    \frac{SS^{\top}}{N^{2}D}.
    \label{app:eq:streaming_covariance_kernel}
\end{equation}
Therefore, each trajectory can be discarded immediately after its contribution has been accumulated. This procedure reduces offline data-loading and repeated-processing overhead and substantially decreases the storage footprint, while exactly preserving all information required by the current PVR selector.

\begin{table*}[t]
    \centering
    \caption{Offline calibration and PVR selector cost. Collection time excludes model loading and VAE decoding. Kernel construction excludes model forward.
    Selector latency measures the time required to construct one complete anchor
    schedule after the covariance kernel and selector configuration have been
    loaded into memory; it is reported as the median / 95th percentile over 1,000 CPU runs. Raw storage is the theoretical size of all solver-consumed predictions, whereas kernel storage size is the actual serialized FP64 selector
    state.}
    \label{tab:app_offline_calibration_cost}
    \setlength{\tabcolsep}{4.5pt}
    \resizebox{\textwidth}{!}{
    \begin{tabular}{lcccccc}
        \toprule
        \textbf{Model}
        & \textbf{Collect / traj.}
        & \textbf{Total collect}
        & \textbf{Kernel build}
        & \textbf{Raw storage}
        & \textbf{Kernel storage}
        & \textbf{Selector med. / p95} \\
        & \textbf{(s)}
        & \textbf{(min)}
        & \textbf{(s)}
        & \textbf{(GiB)}
        & \textbf{(KiB)}
        & \textbf{(ms)} \\
        \midrule
        Qwen-Image-2512
        & $30.27\pm0.09$
        & 5.05
        & 2.96
        & 0.411
        & 22.23
        & 3.74 / 3.77 \\
        Wan~2.1
        & $66.69\pm0.16$
        & 11.12
        & 14.43
        & 1.953
        & 22.18
        & 3.47 / 3.50 \\
        \bottomrule
    \end{tabular}}
\end{table*}

As shown in Table~\ref{tab:app_offline_calibration_cost}, collecting ten calibration trajectories takes 5.05 minutes on Qwen-Image-2512 and 11.12 minutes on Wan~2.1. Streaming kernel construction adds only 2.96 seconds and 14.43 seconds, corresponding to approximately 1.0\% and 2.2\% of the respective collection time. The dominant offline cost therefore remains the N complete denoising runs rather than covariance accumulation. This cost is paid only once per model and sampling configuration, after which the resulting kernel and anchor schedule are shared across all subsequent inference runs.

After applying the sufficient-statistics compression, the persistent selector state is reduced from 0.411--1.953 GiB of raw predictions to approximately 22 KiB, as its dominant component scales with $T^2$ rather than $NTD$. The serialized artifact additionally contains temporal weights and configuration metadata, much of which can be readily reconstructed from the sampling configuration at runtime; retaining only the covariance kernel would further reduce the storage footprint. Once the kernel and selector configuration are available in memory, constructing a complete anchor schedule takes only 3.47--3.74 ms in median. This schedule-construction cost is incurred only once for each model and sampling configuration, after which the resulting schedule is stored and reused during inference, and therefore does not contribute to per-sample online latency.

\paragraph{Online inference overhead.}
Postdictive smoothing requires only lightweight vector operations and scheduler reintegration, without introducing additional neural-network evaluations at skipped timesteps. Table~\ref{tab:app_runtime_overhead} reports the runtime breakdown on Wan~2.1 with a full-computation budget of 20 steps and a dense prefix of 10 steps. We separately measure the costs of neural-network evaluations, postdictive correction and orchestration, scheduler reintegration, VAE decoding, and pipeline preparation. The correction and reintegration together account for only 0.21\% of the total latency, indicating that the computational overhead introduced by PoSTCache is negligible.

\begin{table}[t]
      \centering
      \caption{Runtime breakdown of PoSTCache on Wan.}
      \label{tab:app_runtime_overhead}
      \setlength{\tabcolsep}{5pt}
      \begin{tabular}{lcc}
          \toprule
          \textbf{Component} & \textbf{Latency} & \textbf{Fraction} \\
          \midrule
          Neural-network evaluations
          & 26.869 s & 93.82\% \\

          Postdictive correction / orchestration
          & 0.010 s & 0.03\% \\

          Scheduler reintegration
          & 0.051 s & 0.18\% \\

          VAE decoding
          & 1.657 s & 5.79\% \\

          Pipeline preparation
          & 0.051 s & 0.18\% \\
          \bottomrule
      \end{tabular}
  \end{table}

\begin{figure}[t]
    \centering

    \begin{subfigure}{0.96\linewidth}
        \centering
        \includegraphics[width=\linewidth]{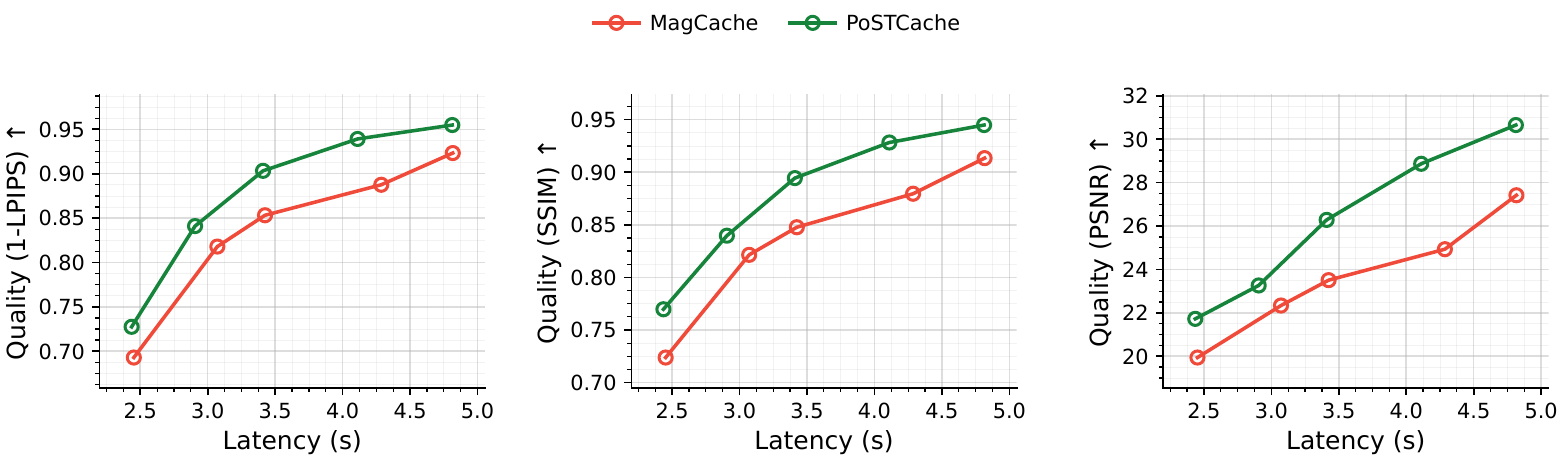}
        \caption{FLUX.1-dev.}
        \label{fig:tradeoff_flux}
    \end{subfigure}

    \vspace{0.5em}

    \begin{subfigure}{0.96\linewidth}
        \centering
        \includegraphics[width=\linewidth]{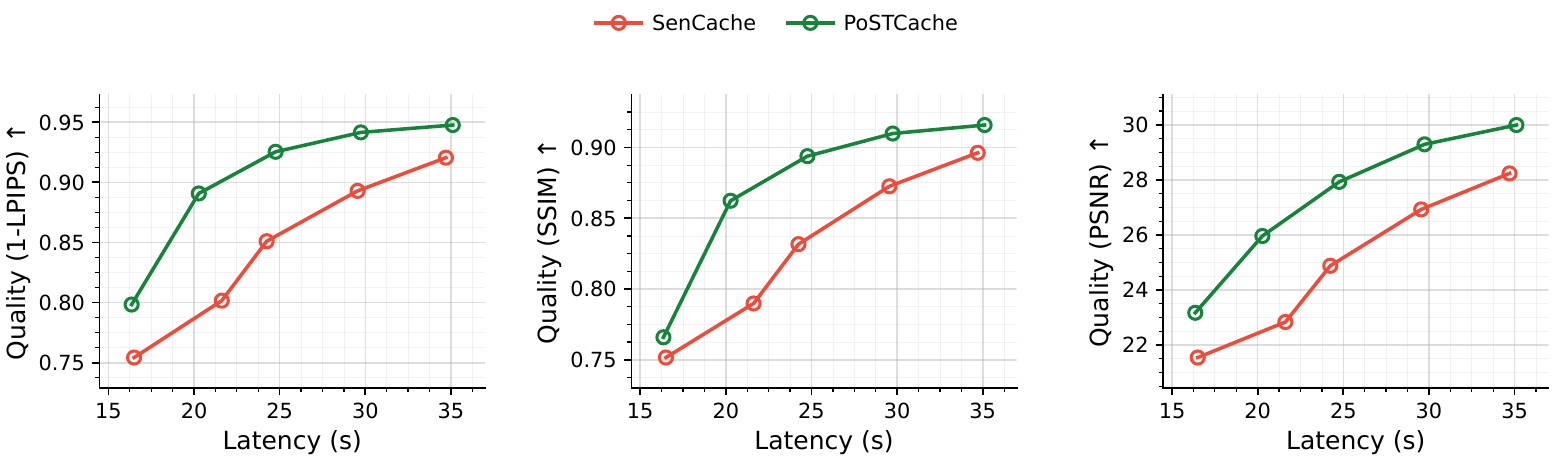}
        \caption{Wan 2.1.}
        \label{fig:tradeoff_wan}
    \end{subfigure}

    \caption{
        Quality--latency trade-off across reconstruction metrics.
        We report $1-\mathrm{LPIPS}$, SSIM, and PSNR under different
        computation budgets on (a) FLUX.1-dev and (b) Wan~2.1.
        PoSTCache consistently shifts the quality--latency frontier
        over MagCache on FLUX.1-dev and SenCache on Wan~2.1 across
        all three metrics.
    }
    \label{fig:additional_tradeoff}
\end{figure}

\paragraph{Efficiency--quality trade-off.}
Figure~\ref{fig:additional_tradeoff} further reports the complete
quality--latency trade-off using all three reconstruction metrics.
Across 1-LPIPS, SSIM, and PSNR, PoSTCache consistently
dominates the corresponding caching baseline over a broad range of
latency budgets.

\subsection{Additional Qualitative Results}
\label{app:additional_qualitative}

We provide additional qualitative comparisons to complement the image results
in the main paper. The examples cover diverse prompts, model architectures, and
acceleration regimes. Across these settings, PoSTCache more closely follows
full inference in terms of global composition, local appearance, and object
structure. For video generation, it additionally reduces structural and
appearance deviations across the sampled frames, while achieving acceleration
ratios comparable to or higher than those of the competing methods.

\begin{figure}[t]
    \centering
    \includegraphics[
        width=1.0\linewidth,
        trim={3.3cm 0.7cm 3.75cm 0.7cm},
        clip
    ]{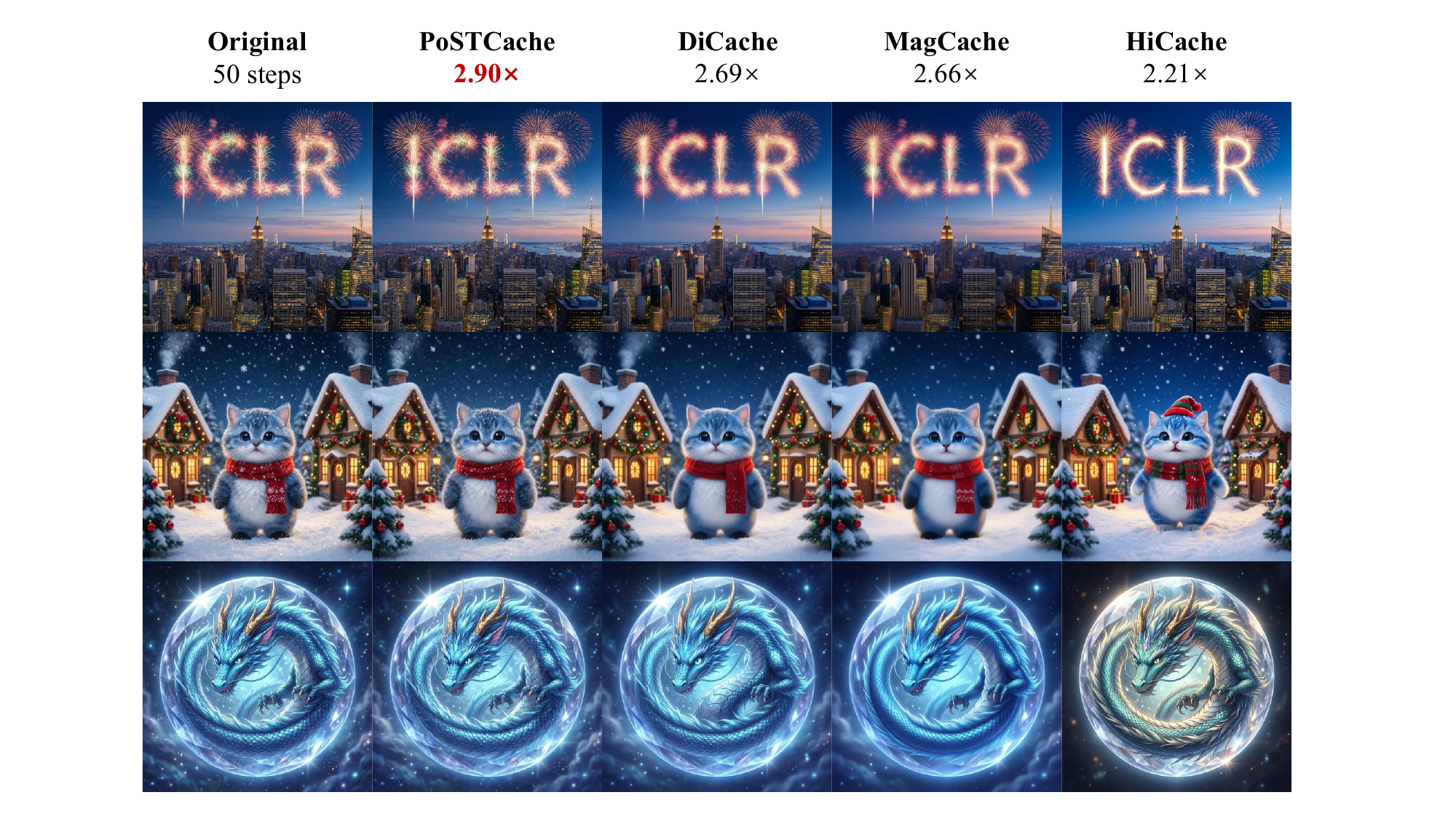}
    \caption{
        Qualitative comparison with previous acceleration methods on
        Qwen-Image-2512.
    }
    \label{fig:qwen_image_compare}
\end{figure}

\paragraph{Qwen-Image-2512.}
Figure~\ref{fig:qwen_image_compare} presents additional text-to-image results
on Qwen-Image-2512. At a $2.90\times$ speedup, PoSTCache closely preserves both
the global layout and fine-grained structures of full inference. In the
fireworks example, it retains the shapes of the rendered letters and the local
burst patterns. In the moonlit cat and industrial landscape examples, it
better preserves small foreground elements, cloud boundaries, vegetation, and
reflection details. In comparison, competing methods tend to smooth away these
local structures or introduce more noticeable changes in color, illumination,
and scene layout. These results indicate that postdictive correction mitigates
not only semantic degradation but also the accumulation of fine-grained
appearance errors.

\begin{figure*}[t]
    \centering
    \includegraphics[
        width=\linewidth,
        trim={0.6cm 7.7cm 1.4cm 3.0cm},
        clip
    ]{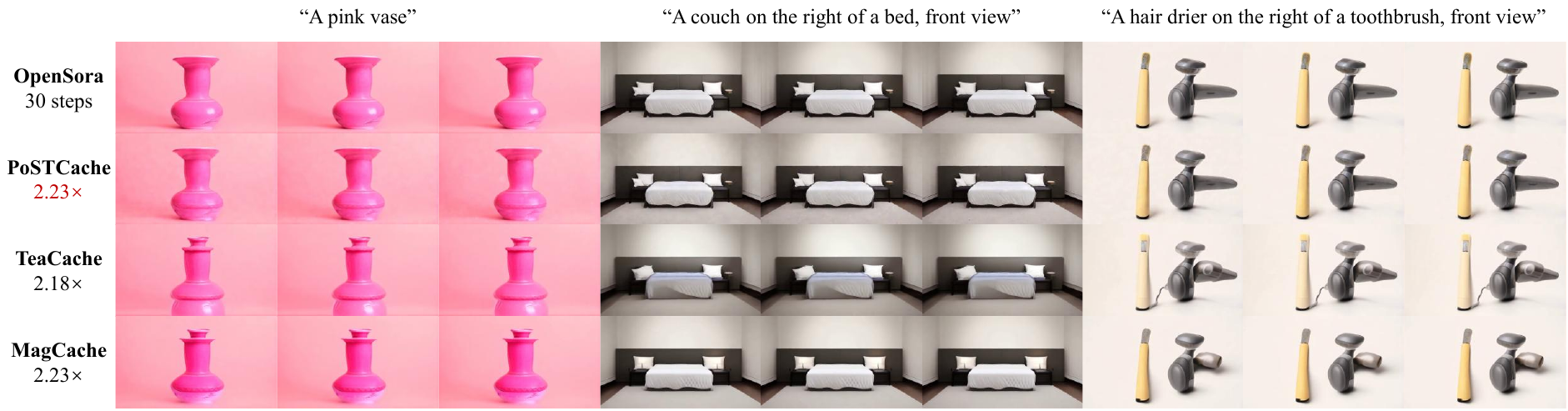}
    \caption{
        Qualitative comparison with previous acceleration methods on
        OpenSora~1.2.
    }
    \label{fig:video_compare_opensora}
\end{figure*}

\paragraph{OpenSora~1.2.}
Figure~\ref{fig:video_compare_opensora} compares PoSTCache with representative
caching methods on OpenSora~1.2. PoSTCache more faithfully preserves object
identity, geometry, and prompt-specified spatial relationships across the
sampled frames. For example, it maintains the contour and proportions of the
pink vase, the relative arrangement of the couch and bed, and the left--right
relationship between the toothbrush and hair drier. TeaCache and MagCache,
by contrast, produce more visible changes in object shape, orientation, or
relative placement. The improvements are obtained at a comparable acceleration
ratio, suggesting that retrospective correction is effective for preventing
structural deviations from propagating along the video trajectory.

\begin{figure*}[t]
    \centering
    \includegraphics[
        width=\linewidth,
        trim={0.6cm 6.0cm 1.4cm 1.0cm},
        clip
    ]{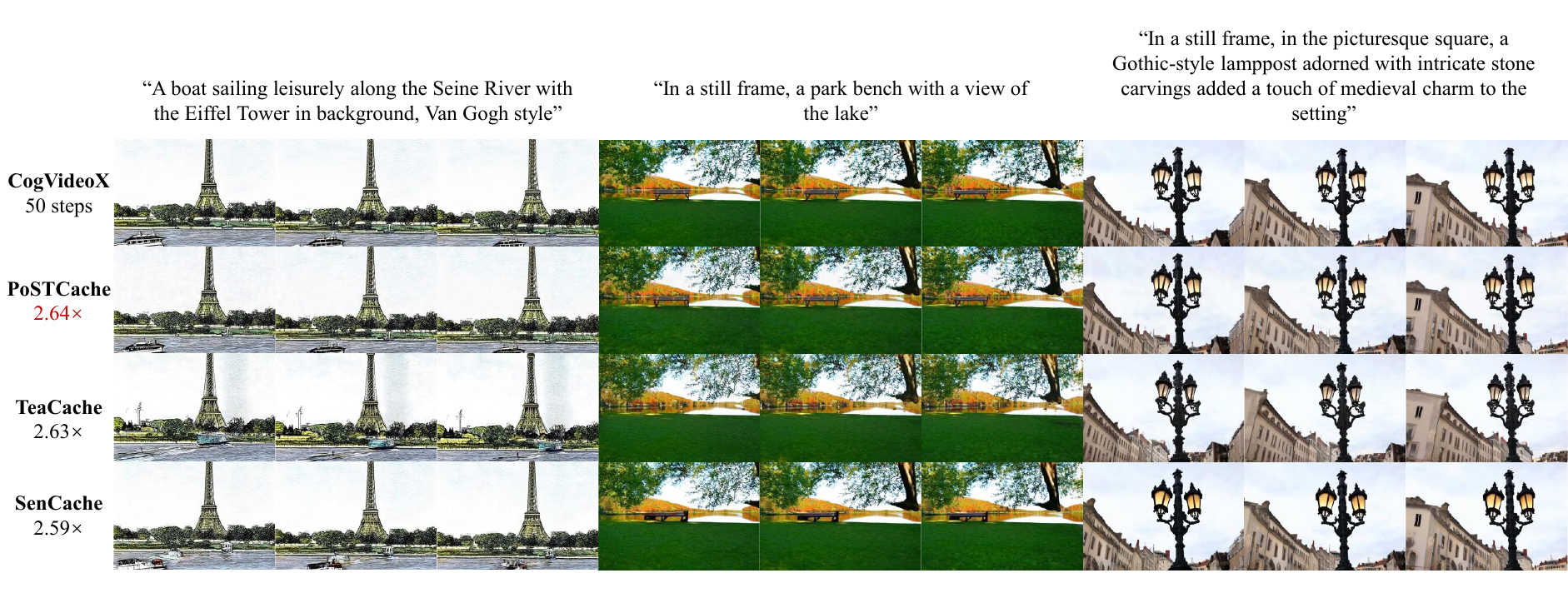}
    \caption{
        Qualitative comparison with previous acceleration methods on
        CogVideoX.
    }
    \label{fig:video_compare_cog}
\end{figure*}

\paragraph{CogVideoX.}
Figure~\ref{fig:video_compare_cog} further evaluates PoSTCache on CogVideoX
using stylized landscapes, natural scenes, and architectural objects.
PoSTCache closely reproduces the object scale and scene composition of full
inference, including the position of the Eiffel Tower and boat, the appearance
of the lakeside bench, and the silhouette of the Gothic-style lamppost.
Moreover, these structures remain comparatively stable across the displayed
frames. TeaCache and SenCache exhibit larger variations in object proportions,
background layout, or foreground appearance. This comparison demonstrates that
PoSTCache remains effective when the underlying model architecture and visual
content differ substantially from those of OpenSora.

\begin{figure*}[t]
    \centering
    \includegraphics[
        width=\linewidth,
        trim={0.6cm 7.7cm 1.4cm 3.0cm},
        clip
    ]{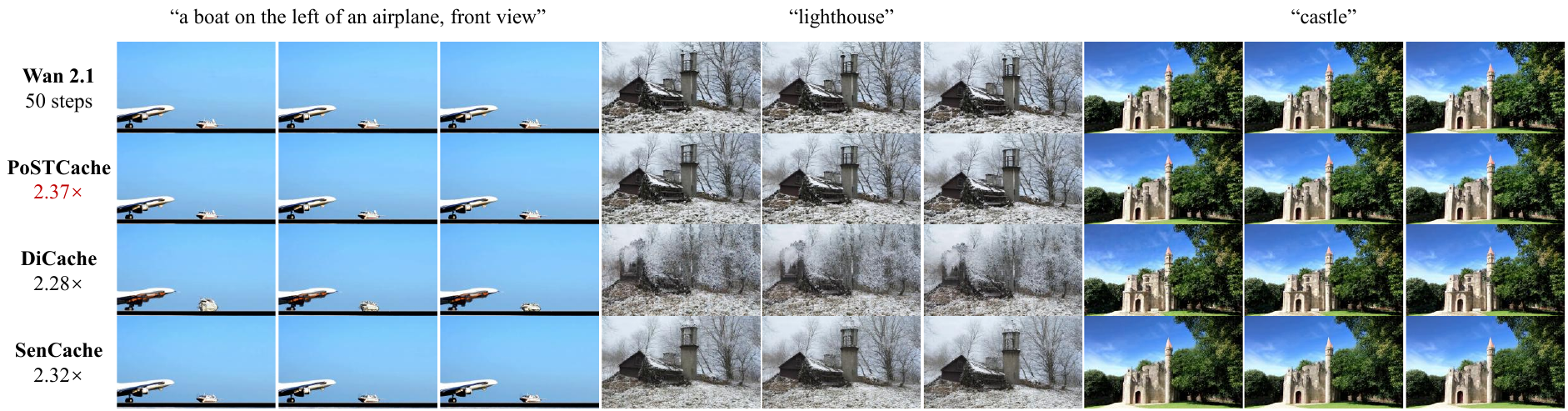}
    \caption{
        Qualitative comparison with previous acceleration methods on
        Wan~2.1.
    }
    \label{fig:video_compare_wan}
\end{figure*}

\paragraph{Wan~2.1.}
Figure~\ref{fig:video_compare_wan} shows the corresponding comparisons on
Wan~2.1. The selected prompts test both compositional relationships and the
preservation of distinctive scene structures. For the boat--airplane example,
PoSTCache maintains the prompt-specified relative positions and closely follows
the full-inference object silhouettes. For the lighthouse and castle examples,
it preserves the principal architectural structures, surrounding vegetation,
and overall scene appearance across frames. DiCache introduces substantial
structural degradation in the lighthouse scene, while SenCache produces more
noticeable changes in object geometry and background appearance. PoSTCache
therefore provides a closer approximation to full inference while attaining
the highest acceleration ratio among the compared methods in this setting.

Overall, the qualitative results are consistent with the reconstruction
metrics reported in the main paper. By revising skipped predictions using
future-anchor evidence, PoSTCache reduces accumulated errors in local texture,
object geometry, spatial composition, and temporal appearance across different
image and video generation architectures.

\subsection{Limitations and Boundary Conditions}
\label{app:limitations}

PoSTCache relies on offline trajectory statistics to construct a shared
anchor schedule. The resulting schedule transfers well across the evaluated
prompt distributions, guidance scales, and spatial configurations, but this
does not imply that the underlying dependency profile is invariant or that a
single schedule is optimal for every sample. Target-specific reprofiling can
improve the offline PVR objective or the held-out prediction-space error
without improving the final reconstruction. This discrepancy reflects an
offline--online gap: the offline profile is estimated on stored
full-computation trajectories, whereas online correction changes the latent
states and recursively affects subsequent model predictions. In addition,
the current greedy selector is path dependent; small finite-sample
perturbations of otherwise similar covariance profiles can alter early anchor
choices and produce substantially different schedules. Combining the shared
offline prior with inexpensive online trajectory observations is therefore a
promising direction for more stable and sample-adaptive anchor selection.

Retrospective reintegration can also introduce a local state--prediction
mismatch at future anchors.
Appendix~\ref{app:anchor_consistency} shows that unconditional anchor refresh
reduces this mismatch but is less compute-efficient than broader anchor
coverage in our matched-NFE comparison.
Our analysis nevertheless compares only two coarse allocation strategies and
does not exclude more selective policies.
An adaptive scheme that refreshes only high-risk anchors, or dynamically
allocates each additional model evaluation between anchor refresh and new
anchor placement, may yield a more favorable consistency--coverage trade-off.

Postdictive correction further relies on the temporal correlation between
skipped prediction errors and innovations revealed by future anchors. Our
implementation approximates the full covariance-dependent smoothing gain
with an isotropic linear gain determined by sampling progress. Although this
approximation is effective in the evaluated regimes, it cannot fully capture
nonlinear, feature-dependent, or sample-specific error evolution. Under very
aggressive acceleration or unusually long cache intervals, accumulated drift
may weaken the relationship between a future innovation and earlier skipped
predictions, limiting the recoverable error.

\end{document}